\documentclass[10pt]{article} 

\usepackage[accepted]{tmlr}

\usepackage{amsmath,amsfonts,bm}

\def\eqref#1{equation~\ref{#1}}

\def\1{\bm{1}}

\DeclareMathAlphabet{\mathsfit}{\encodingdefault}{\sfdefault}{m}{sl}
\SetMathAlphabet{\mathsfit}{bold}{\encodingdefault}{\sfdefault}{bx}{n}

\usepackage{hyperref}
\usepackage{url}

\usepackage{graphicx}
\usepackage{amsmath}
\usepackage{amssymb}
\usepackage{booktabs}
\usepackage[capitalize]{cleveref}
\crefname{section}{Sec.}{Secs.}
\crefname{table}{Tab.}{Tabs.}
\usepackage{subcaption}
\usepackage{tabularx}
\usepackage{multirow}
\usepackage{nicefrac}

\usepackage{tikz}
\usepackage{pgfplots}
\usetikzlibrary{positioning}
\usetikzlibrary{shapes.geometric, shapes, arrows, arrows.meta}
\usepackage{siunitx}
\usepackage{pgfplotstable}

\usepackage{bbding} 
\usepackage{fontawesome}
\usepackage{stmaryrd}

\newcolumntype{Y}{>{\centering\arraybackslash}X} 

\usepackage{xcolor}
\usepackage{colortbl}

\newcommand{\bn}[1]{\textbf{#1}}
\newcommand\gy[1]{\textcolor{gray}{#1}}
\newcommand\rebut[1]{{#1}}
\newcommand\rebuttwo[1]{{#1}}

\newcommand\cng[1]{#1} 
\newcommand\cngtwo[1]{#1}
\newcommand\cngsr[1]{#1} 

\usepackage{acronym}
\newacro{miou}[mIoU]{mean Intersection over Union}
\newacro{acc}[Acc]{Accuracy}
\newacro{pca}[PCA]{Principal Component Analysis}
\newacro{crf}[CRF]{Conditional Random Field}
\newacro{svd}[SVD]{Singular Value Decomposition}
\newacro{vit}[ViT]{Vision Transformer}
\newacro{em}[EM]{Expectation Maximization}
\newacro{ssl}[SSL]{self-supervised learning}
\newacro{ema}[EMA]{exponential moving average}

\newcommand*{\tran}{^{\mkern-1.5mu\top}}

\usepackage{xspace}
\newcommand{\eg}{\emph{e.\thinspace{}g.}\@\xspace}
\newcommand{\ie}{\emph{i.\thinspace{}e.}\@\xspace}

\newcommand{\wrt}{\emph{w.\thinspace{}r.\thinspace{}t.}\@\xspace}
\newcommand{\cf}{\emph{cf.}\@\xspace}

\newcommand{\ourmethod}{PriMaPs\xspace}
\newcommand{\ourmethodframework}{PriMaPs-EM\xspace}
\newcommand{\ourframework}{moving average stochastic \ac{em}\xspace}

\definecolor{color1}{RGB}{255, 176, 0} 
\definecolor{color2}{RGB}{227, 75, 228} 
\definecolor{color3}{RGB}{26, 133, 255} 
\definecolor{color4}{RGB}{111, 60, 117} 
\definecolor{city}{RGB}{100,143,255}
\definecolor{coco}{RGB}{254, 97, 0}
\definecolor{pots}{RGB}{220, 38, 127}

\definecolor{mylightgray}{RGB}{235, 235, 235}
\definecolor{mygray}{RGB}{150, 150, 150}

\newcommand*{\inparagraph}[1]{\smallskip\noindent\textbf{#1}\hspace{0.5em}}

\title{Boosting Unsupervised Semantic Segmentation\\ with Principal Mask Proposals}

\newcommand{\myspacing}{\hspace{1.5em}}
\author{Oliver Hahn\textsuperscript{\normalfont{}1}
        \myspacing
        Nikita Araslanov\textsuperscript{\normalfont{}2,3}
        \myspacing
        Simone Schaub-Meyer\textsuperscript{\normalfont{}1,4}
        \myspacing
        Stefan Roth\textsuperscript{\normalfont{}1,4}\\
        \addr
        $^{1}$ Department of Computer Science, TU Darmstadt
        \myspacing
        $^{2}$ Department of Computer Science, TU Munich\\
        $^{3}$ Munich Center for Machine Learning (MCML)
        \myspacing
        $^{4}$hessian.AI}

\def\month{09}  
\def\year{2024} 
\def\openreview{\url{https://openreview.net/forum?id=UawaTQzfwy}} 

\begin{document}

\setlength{\abovedisplayskip}{5pt}
\setlength{\belowdisplayskip}{5pt}

\maketitle

\vspace{-0.5em}
\begin{abstract}
Unsupervised semantic segmentation aims to automatically partition images into semantically meaningful regions by identifying global \cngsr{semantic} categories within an image corpus without any form of annotation. 
Building upon recent advances in self-supervised representation learning, we focus on how to leverage these large pre-trained models for the downstream task of unsupervised segmentation. 
We \cng{present} \ourmethod\ -- Principal Mask Proposals -- decomposing images into semantically meaningful masks based on their feature representation. 
\cng{This allows us to realize unsupervised semantic segmentation by fitting class prototypes to \ourmethod with a stochastic expectation-maximization algorithm, \ourmethodframework.}
Despite its conceptual simplicity, \cng{\ourmethodframework} leads to competitive results across various pre-trained backbone models, including DINO and DINOv2, and across \cngsr{different} datasets, such as Cityscapes, COCO-Stuff, and Potsdam-3. 
\cng{Importantly}, \ourmethodframework is able to boost results when applied orthogonally to current state-of-the-art unsupervised semantic segmentation pipelines. 
Code \cngsr{is} available at \url{https://github.com/visinf/primaps}.

\end{abstract}

\section{Introduction}
\label{sec:introduction}

Semantic image segmentation is a dense prediction task that classifies image pixels into categories from a pre-defined semantic taxonomy. 
Owing to its fundamental nature, semantic segmentation has a broad range of applications, such as image editing, medical imaging, robotics, or autonomous driving \cite[see][for an overview]{Minaee:2022:ISU}. 
Addressing this problem via supervised learning requires ground-truth labels for every pixel \citep{Long:2015:FCN, Ronneberger:2015:UNC, Chen:2018:DLS}.
Such manual annotation is extremely time and resource intensive. 
For instance, a trained human annotator requires an average of 90\:minutes to label up to 30 classes in a single 2\,MP image \citep{Cordts:2016:CDS}. 
While committing significant resources to large-scale annotation efforts achieves excellent results \citep{Kirillov:2023:SAM}, there is natural interest in a more economical approach.
Alternative lines of research aim to solve the problem using cheaper -- so-called ``weaker'' -- variants of annotation. 
For example, image-level supervision describing the semantic categories present in the image, or bounding-box annotations, can reach impressive levels of segmentation accuracy \citep{Dai:2015:BEB, Araslanov:2020:SSS, Oh:2021:BAP,Xu:2022:MCT,Ru:2023:TCW}. 

As an extreme problem scenario toward reducing the annotation effort, unsupervised semantic segmentation aims to consistently discover and categorize image regions in a given data domain without any labels, knowing only how many classes to discover.
Unsupervised semantic segmentation is highly ambiguous as class boundaries and the level of categorical granularity are task-dependent.\footnote{While assigning actual semantic labels to regions without annotation is generally infeasible, the \cngtwo{assumption} is that the categories of the discovered segments will strongly correlate with human notions of semantic meaning.}
However, we can leverage the fact that typical image datasets have a homogeneous underlying taxonomy and exhibit invariant domain characteristics.
Therefore, it is still feasible to decompose images in such datasets in a semantically meaningful and consistent manner without annotations.

Despite the challenges of unsupervised semantic segmentation, we have witnessed remarkable progress on this task in the past years \citep{Ji:20219:IIC, Cho:2021:PUS, Gansbeke:2021:USS, Gansbeke:2022:DOM, Ke:2022:UHS, Yin:2022:TTA, Hamilton:2022:USS, Karlsson:2022:VID, Li:2023DCN, Seong:2023:LHP, Seitzer:2023:BTG}. 
Deep representations obtained with \ac{ssl}, such as DINO~\citep{Caron:2021:EPS}, have played a critical role in this advance. 
However, it remains unclear whether previous work leverages the intrinsic properties of the original \ac{ssl} representations, or merely uses them for ``bootstrapping'' and learns a new representation on top. 
Exploiting the inherent properties of \ac{ssl} features is preferable for two reasons. 
First, training \ac{ssl} models incurs a substantial computational effort, justifiable only if the learned feature extractor is sufficiently versatile. 
\cngtwo{In other words, one} can amortize the high computational cost over many downstream tasks, provided that task specialization is computationally negligible. 
Second, studying \ac{ssl} representations with lightweight tools, such as linear models, leads to a more interpretable empirical analysis than with the use of more complex models, as evidenced by the widespread use of linear probing in \ac{ssl} evaluation. 
Such interpretability advances research on \ac{ssl} models toward improved cross-task generalization.

\begin{figure}[!t]
\centering
\input{figures/method/primaps_example}
\vspace{-1.0em}
\caption{\cng{\textbf{\ourmethod pseudo-label example.} Principal mask proposals (\ourmethod) are iteratively extracted from an image (dashed arrows). Each mask is assigned a semantic class, resulting in a pseudo label. The examples are taken from the Cityscapes \textit{(top)}, COCO-Stuff \textit{(middle)}, and Potsdam-3 \textit{(bottom)} datasets.  \label{fig:primaps_example}}}
\vspace{-1.0em}
\end{figure}

Equipped with essential tools of linear modeling, \ie \ac{pca}, we generate \textbf{Pri}ncipal \textbf{Ma}sk \textbf{P}roposal\textbf{s}, or \ourmethod, directly from the \ac{ssl} representation.
Complementing previous findings on object-centric images \citep{Tumanyan:2022:SVT,Amir:2021:DVT}, we show that principal components of \ac{ssl} features tend to identify visual patterns with high semantic correlation also in scene-centric imagery.
Leveraging \ourmethod and minimalist post-processing, we construct semantic pseudo labels for each image \cng{as illustrated in \cref{fig:primaps_example}.}
Finally, instead of learning a new embedding on top of the \ac{ssl} representation \citep{Hamilton:2022:USS, Seong:2023:LHP, Seitzer:2023:BTG, Zadaianchuk:2023:USS}, we employ a moving average implementation of stochastic \ac{em} \citep{Chen:2018:SEM} to assign a consistent category to each segment in the pseudo labels and directly optimize class prototypes in the feature space. Our experiments show that this straightforward approach not only boosts the segmentation accuracy of the DINO baseline, but also that of more advanced state-of-the-art approaches tailored for semantic segmentation, such as STEGO~\citep{Hamilton:2022:USS} and HP~\citep{Seong:2023:LHP}.

We make the following contributions:
\textit{(i)} We derive lightweight mask proposals, leveraging intrinsic properties of the embedding space, \eg, the covariance \cngsr{matrix}, provided by an off-the-shelf \ac{ssl} approach. 
\textit{(ii)} Based on the mask proposals, we construct pseudo labels and employ \ourframework{} to assign a consistent semantic class to each proposal.
\textit{(iii)} We demonstrate improved segmentation accuracy across a wide range of \ac{ssl} embeddings and datasets.

\section{Related Work}
\label{sec:related_work}

Our work builds upon recent advances in self-supervised representation learning, and takes inspiration from previous unsupervised semantic and instance segmentation methods.

The goal of \textbf{self-supervised representation learning (SSL)} is to provide generic, task-agnostic feature extractors \citep{He:2020:MCU, Chen:2020:AFC, Grill:2020:BYL}.
A pivotal role in defining the behavior of self-supervised features on future downstream tasks is taken by the self-supervised objective, the so-called \emph{pretext} task.
Examples of such tasks include predicting the context of a patch \citep{Doersch:2015:UVR} or its rotation \citep{Gidaris:2018:URL}, image inpainting \citep{Pathak:2016:CEF}, and ``solving'' jigsaw puzzles \citep{Noroozi:2016:ULV}.
Another family of self-supervised techniques is based on contrastive learning \citep{Chen:2020:AFC,Caron:2020:ULV}.
More recently, Transformer networks \citep{Dosovitskiy:2021:AIW} revived some older pretext tasks, such as context prediction \citep{Caron:2021:EPS,He:2022:MAE}, in a more data-scalable fashion.
 
While the standard evaluation practice in SSL (\eg, linear probing, transfer learning) offers some glimpse into the feature\cngsr{s'} properties, understanding the embedding space produced by SSL remains an active terrain for research \citep{Ericsson:2021:HWD, Naseer:2021:IPV}.
In particular, DINO features \citep{Caron:2021:EPS,Oquab:2023:DLR} are known to encode accurate object-specific information, such as object parts \citep{Amir:2021:DVT, Tumanyan:2022:SVT}. 
However, it remains unclear to what extent DINO embeddings allow for semantic representation of the more ubiquitous multi-object scenes.
Here, following previous work \citep[\eg,][]{Hamilton:2022:USS, Seong:2023:LHP}, we provide further insights.

Early techniques for \textbf{unsupervised semantic segmentation} using deep networks \citep{Cho:2021:PUS,Gansbeke:2021:USS} approach the problem in the spirit of transfer learning and, under certain nomenclatures, may not be considered fully unsupervised. 
Specifically, starting with \emph{supervised} ImageNet pre-training \citep{Russakovsky:2015:ILS}, a network obtains a fine-tuning signal from segmentation-oriented training objectives. Such supervised ``bootstrapping'' appears to be crucial in the ill-posed unsupervised formulation. 
Unsupervised training of a deep model for segmentation from scratch is possible, albeit sacrificing accuracy \citep{Ji:20219:IIC, Ke:2022:UHS}. 
However, training a new deep model for each downstream task contradicts the spirit of SSL of amortizing the high SSL training costs over many computationally cheap specializations of the learned features \citep{Bommasani:2021:OOR}.

Relying on \emph{self-supervised} DINO pre-training, recent work \citep{Hamilton:2022:USS, Li:2023DCN, Seong:2023:LHP} has demonstrated the potential of such amortization with more lightweight fine-tuning for semantic segmentation. 
Nevertheless, most of this work has treated the SSL representation as an inductive prior by learning a new embedding space over the SSL features \citep[\eg,][]{Hamilton:2022:USS,Seong:2023:LHP}. 
In contrast, following SSL principles, we use the SSL representation in a more direct and lightweight fashion -- by extracting mask proposals using linear models (\ac{pca}) with minimal post-processing and learning a direct mapping from feature to prediction space.

Mask proposals have an established role in computer vision \citep{Arbelaez:2011:CDH,Uijlings:2013:SSO}, and remain highly relevant in deep learning \citep{Hwang:2019:SSS, Gansbeke:2021:USS,Yin:2022:TTA}. 
Different from previous work, we directly derive the mask proposals from SSL representations. 
Our approach is inspired by the recent use of classical algorithms, such as normalized cuts \citep[Ncut][]{Shi:2000:NCI}, in the context of self-supervised segmentation \citep{Wang:2023:CAL, wang:2023:TSO}. However previous approaches \citep{Gansbeke:2021:USS, Gansbeke:2022:DOM, Wang:2023:CAL, wang:2023:TSO} mainly proposed foreground object masks on object-centric data, utilized in a multi-step self-training. In contrast, we develop a straightforward method for extracting dense pseudo labels for learning unsupervised semantic segmentation of scene-centric data and show consistent benefits in improving the segmentation accuracy across a variety of baselines and state-of-the-art methods \citep{Hamilton:2022:USS,Seong:2023:LHP}.

\section{\ourmethod: Principal Mask Proposals}
\label{sec:method}
\cngsr{In this paper, we leverage} recent advances in self-supervised representation learning \citep{Caron:2021:EPS, Oquab:2023:DLR} for the specific downstream task of unsupervised semantic segmentation.
Our approach is based on the observation that such pre-trained features already exhibit intrinsic spatial similarities, capturing semantic correlations\cngtwo{, \cngsr{thus} providing guidance to fit global pseudo-class representations.}

\inparagraph{A simple baseline.}
Consider a simple baseline that applies $K$-means clustering to DINO ViT features \citep{Caron:2021:EPS}. Surprisingly, this already leads to reasonably good unsupervised semantic segmentation results, \eg, around $15\:\%$ mean IoU to segment 27 classes on Cityscapes~\citep{Cordts:2016:CDS}, see \cref{tab:main_cs}. However, supervised linear probing between the same feature space and the ground-truth labels -- the theoretical upper bound -- leads to clearly superior results of almost  $36\:\%$ \cngsr{mean IoU}. 
Given this gap and the simplicity of the approach, we conclude that there is \emph{valuable potential} in directly obtaining semantic segmentations without enhancing the original feature representation, \cngtwo{unlike} in previous work \citep{Hamilton:2022:USS, Seong:2023:LHP}.

\inparagraph{From $K$-means to \ourmethodframework.}
When examining the $K$-means baseline as well as state-of-the-art methods \citep{Hamilton:2022:USS, Seong:2023:LHP}, see \cref{fig:qualitative}, it can be qualitatively observed that more local consistency within the respective predictions would already lead to less mis-classification. We take inspiration from \citep{Drineas:2004:CLG, Ding:2004:KMC}, who showed that the \ac{pca} subspace, spanned by principal components, is a relaxed solution to $K$-means clustering. We observe that principal components have high semantic correlation for object- as well as scene-centric image features (\cf \cref{fig:primaps_example}).
We utilize this by iteratively partitioning images based on dominant feature patterns, identified by means of the cosine similarity of the image features to the respective first principal component.
\cng{We name the resulting class-agnostic image decomposition \emph{\ourmethod} -- Principal Mask Proposals.
\ourmethod stem directly from SSL representations and guide the process of unsupervised semantic segmentation.
Shown in \cref{fig:method}, our optimization-based approach, 
\ourmethodframework, operates over an SSL feature representation computed from a frozen deep neural network backbone.
The optimization realizes stochastic \ac{em} of a clustering objective guided by \ourmethod.
Specifically, \ourmethodframework fits class prototypes to the proposals in a globally consistent manner by optimizing over two identically sized vector sets, with one of them being an \ac{ema} of the other.}
We show that \ourmethodframework enables accurate unsupervised partitioning of images into semantically meaningful regions while being comparatively lightweight and orthogonal to most previous approaches \cngsr{to} unsupervised semantic segmentation.

\subsection{Deriving \ourmethod} 
\label{sec:prima-pseudo}
We start with a frozen pre-trained self-supervised backbone model $\mathcal{F}: \mathbb{R}^{3\times h \times w} \rightarrow \mathbb{R}^{C\times H\times W}$, which embeds an image $I\in \mathbb{R}^{3\times h \times w}$ into a dense feature representation $f\in\mathbb{R}^{C\times H\times W}$ as
\begin{equation}
    \quad f = \mathcal{F}(I)\, .
\end{equation}
Here, $C$ refers to the channel dimension of the dense features, and $H = \nicefrac{h}{p}, W = \nicefrac{w}{p}$ with $p$ corresponding to the output stride of the backbone. Based on this image representation, the next step is to decompose the image into semantically meaningful masks to provide a local grouping prior for \cng{fitting} global class prototypes.



\inparagraph{Initial principal mask proposal.}
To identify the initial principal mask proposal in an image $I$, we analyze the spatial statistical correlations of its features \rebut{by means of PCA}.
Specifically, we consider the empirical feature covariance matrix
\begin{equation}
    \label{eq:covariance}
   \Sigma = \frac{1}{H W} \sum_{i=1}^H \sum_{j=1}^W \bigl(f_{:,i,j}-\bar{f}\bigr)\bigl(f_{:,i,j}-\bar{f}\bigr)\tran\,,
\end{equation}
where $f_{:,i,j}\in\mathbb{R}^C$ are the features at \cng{position} $(i,j)$ and $\bar{f}\in\mathbb{R}^C$ is the mean feature. To identify the feature direction that captures the largest variance in the feature distribution, we seek the first principal component of $\Sigma$ by solving 
\begin{equation}
    \label{eq:pca}
    \Sigma v = \lambda v\,.
\end{equation}
We obtain the first principal component as the eigenvector $v_1$ to the largest eigenvalue $\lambda_1$, which can be computed efficiently with \ac{svd} \cngtwo{using the flattened features $f$.}

To identify a candidate region, our next goal is to compute a spatial feature similarity map to the dominant feature direction.
We observe that doing so directly with the principal direction does not always lead to sufficient\cngsr{ly good} localization, \ie, high similarities arise across multiple visual concepts in an image, elaborated in more detail in \cngtwo{\cref{sec:details-nn}}. This can be circumvented by first anchoring the dominant feature vector in the feature map.
To that end, we obtain the nearest neighbor feature $\Tilde{f}\in\mathbb{R}^C$ of the first principal component $v_1$ by considering the cosine distance in the normalized feature space $\hat{f}$ as
\begin{equation}
    \label{eq:nn}
    \Tilde{f} = \hat{f}_{:,\cngsr{m,n}}\,,\quad \text{where}\quad (\cngsr{m,n}) = \operatorname*{arg\,max}_{i,j} \: \bigl(v_1\tran \hat{f}\bigr)\,.
\end{equation}
Given this, we compute the cosine-similarity map $M^{1}\in\mathbb{R}^{H\times W}$ of the dominant feature \wrt all features as
\begin{equation}
    \label{eq:similarity-map}
    M^1 = (M_{i,j})_{i,j}\,, \quad \text{where}\quad M_{i,j} = \bigl(\Tilde{f}\bigr)\tran \hat{f}_{:,i,j}\,.
\end{equation}
Next, a threshold $\psi \in (0,1)$ is applied to the similarity map in order to suppress noise and further localize the initial mask. Accordingly, elements of a binary similarity map $P^1\in\{0,1\}^{H\times W}$ are set to $1$ when \cngsr{the similarity is} larger than a fraction $\psi$ of the maximal similarity, and $0$ otherwise, \ie,
\begin{equation}
    \label{eq:principal-mask}
    P^1 = \Bigl[ M^{1}_{i,j} >  \psi \cdot \max_{m,n} M^{1}_{m,n} \Bigr]_{i,j}\,,
\end{equation}
where $[\cdot]$ denotes the Iverson bracket. This binary \emph{principal mask} $P^1$ gives rise to the first principal mask proposal in image $I$.

\inparagraph{Further principal mask proposals.}
\cngtwo{Subsequent mask proposals result from iteratively repeating the described procedure.}
To that end, it is necessary to suppress features that have already been assigned to a pseudo label.
Specifically in iteration $z$, given the mask proposals $P^s,\: s=1,\ldots,z-1$, extracted in previous iterations, we mask out the features that have already been considered as
\begin{equation}
    \label{eq:updated-features}
    f_{:,i,j}^z = f_{:,i,j} \left[\sum\nolimits_{s=1}^{z-1} P^s_{i,j}=0\right]\,.
\end{equation}
Applying \crefrange{eq:covariance}{eq:principal-mask} on top of the masked features $f^z$ yields the \cngsr{cosine-similarity map $M^{z}$ and the }principal mask proposal $P^z$, and so on.
We repeat this procedure until the majority of features (\eg, 95\%) have been assigned to a mask. 
In a final step, the remaining features, in case there are any, are assigned to an ``ignore'' mask
\begin{equation}
    P_{i,j}^0 = 1 - \sum_{z=1}^{Z-1} P_{i,j}^z\,.
\end{equation}
This produces a tensor $P \in  \{0,1\}^{Z\times H\times W}$ of $Z$ spatial similarity masks, decomposing a single image into $Z$ non-overlapping regions.

\inparagraph{Proposal post-processing.}
To further improve the alignment of the masks with edges and color-correlated regions in the image, a fully connected \ac{crf} with Gaussian edge potentials \citep{Kraehenbuehl:2011:EIF} is applied to the initial mask proposals $P$ (after bilinear upsampling to the image resolution) for 10 inference iterations. \rebuttwo{The process for obtaining \ourmethod is visualized in \cref{fig:primapsprocess}.}

\begin{figure}[!t]
\centering
\input{figures/method/primaps_exp/primaps_exp}
\vspace{-1.5em}
\caption{\rebut{\textbf{\ourmethod process.} Given the dense feature embeddings $f$ of an image $I$, we compute the cosine-similarity map $M^1$ of all features $f$ to their first principal component's nearest neighbor feature. The first PriMaP $P^1$ is obtained by thresholding $M^1$. To obtain $P^2$, the features assigned to $P^1$ are masked out, and the process is repeated \rebuttwo{with the remaining features $f^2$. We repeat the PriMaPs process until the majority of features have been assigned to masks.} Finally, all masks $P$ are upsampled and refined using a CRF.} \label{fig:primapsprocess}}
\end{figure}

\inparagraph{\ourmethod pseudo-label generation.} \rebut{In order to form a pseudo label for semantic segmentation out of the $Z$ class-agnostic mask proposals, each mask has to be assigned one out of $K$ pseudo-class labels. This is accomplished using a segmentation prediction of our optimization process, called \ourmethodframework, detailed in \cref{sec:prima}. Given a segmentation prediction $y$, we assign the pseudo-class ID that is most frequently predicted within each proposal, yielding the final pseudo-label map $P^{\ast}\in\{0,1\}^{K \times h\times w}$, a one-hot encoding of a pseudo-class ID. The entire \ourmethod pseudo-label generation process is illustrated in \Cref{fig:maskprop}.}

\begin{figure}[tb]
    \begin{subfigure}[c]{0.47\linewidth}
        \input{figures/method/architecture/architecture_fig_double}
        \vspace{-2.0em}
        \caption{\ourmethodframework architecture\label{fig:architecture}}
    \end{subfigure}
\hspace{0.5em}
    \begin{subfigure}[c]{0.47\linewidth}
        \input{figures/method/maskprop/maskprop_double}
        \vspace{-0.5em}
        \caption{\ourmethod pseudo-label generation\label{fig:maskprop}}
    \end{subfigure}
\caption{\rebut{\textbf{(a) \ourmethodframework architecture.} An image $I$ and its augmented version $I'$ are embedded by the frozen self-supervised backbone $\mathcal{F}$, resulting in the dense features $f$ and $f'$. The segmentation prediction $y$ by the momentum class prototypes $\theta_M$ arises via the dot product with $f$. Likewise, $y'$ arises from the dot product of the running class prototypes $\theta_R$ with $f'$. Pseudo labels $P^\ast$ are constructed using \ourmethod, $I$, and $y$. We use the pseudo labels to optimize $\theta_R$, applying a focal loss. $\theta_R$ is gradually transferred to $\theta_M$ by means of an \ac{ema}. \textbf{(b) \ourmethod pseudo-label generation.} Masks $P$ are proposed by iterative binary partitioning based on the cosine similarity of the features of any unassigned pixel to their first principal component’s nearest neighbor feature. Gray indicates these iterative steps. Next, the masks $P$ are aligned to the image $I$ using a CRF. Finally, a per-mask pseudo-class ID is assigned using majority voting based on the segmentation prediction $y$, resulting in the pseudo label $P^\ast$. \label{fig:method}}}

\end{figure}

\subsection{\cng{\ourmethodframework}\label{sec:prima}}
Shown in \cref{fig:method}, \ourmethodframework is an iterative optimization technique \rebuttwo{for fitting class prototypes for semantic segmentation across a dataset utilizing pseudo labels $P^\ast$ based on \ourmethod} (\cf \cref{sec:prima-pseudo}). \ourmethod provide guidance through local \rebut{per-image} consistency for fitting the global class prototypes \rebut{across a dataset}. \ourmethodframework leverages a frozen pre-trained self-supervised backbone model $\mathcal{F}$ and optimizes two identically sized vector sets, \rebuttwo{the running class prototypes $\theta_R$ and their moving average, the momentum class prototypes $\theta_M$.}
The class prototypes $\theta_R$ and $\theta_M$ are the $K$ pseudo-class representations in the feature space, projecting the $C$-dimensional features linearly to $K$ semantic pseudo classes, \rebut{which equates to segmenting.} \rebut{More precisely, we use the segmentation prediction of the momentum class prototypes $\theta_M$ to assign consistent semantic pseudo-class IDs to the class-agnostic \ourmethod. We optimize the running class prototypes $\theta_R$ using the pseudo labels $P^\ast$ and gradually transfer $\theta_R$ to $\theta_M$ using an \ac{ema}.}

\ourmethodframework \cngsr{performs the optimization} in two stages, since in our case, a meaningful initialization of the class prototypes is vital to provide a reasonable optimization signal. This can be traced back to the pseudo-label generation, which utilizes a segmentation prediction to assign globally consistent classes to the masks. Initializing the class prototypes randomly leads to a highly unstable and noisy signal.

\inparagraph{Initialization.}
We initialize the class prototypes $\theta_M$ with the first $K$ principal components \rebut{using vanilla PCA computed over the feature embeddings of a large number of images across the respective dataset}. Next, a cosine-distance batch-wise $K$-means~\citep{MacQueen:1967:SMC} loss 
\begin{equation}
    \label{eq:kmeans}
    \mathcal{L}_{K\text{-means}}(\theta_M) = - 
    \sum_{i,j}\max \bigl(\theta_M^\top f_{:,i,j}\bigr)
\end{equation}
is minimized with respect to $\theta_M$ for a fixed number of epochs. This minimizes the cumulative cosine distances of the image features $f_{:,i,j}$ to their respective closest class prototype. $\theta_R$ is initialized with the same prototypes.


\inparagraph{Moving average stochastic EM.}
\rebuttwo{Visualized in \cref{fig:architecture}, every optimization iteration starts with computing the}
\rebut{segmentation prediction $y$ by the dot product of the dense image features $f$ of the image $I$ with the momentum class prototypes $\theta_M$. \rebuttwo{Applying the $\arg\max$ results in the dominant prototype for each feature location and, accordingly, in prototype-based binary masks $y\in\{0,1\}^{K \times h\times w}$.} Note that we bilinearly upsample the features to the image resolution before the dot product. 
Using $y$, $f$, and $I$, we get \ourmethod pseudo labels $P^\ast$ as described in \cref{sec:prima-pseudo}. We obtain the semantic prediction $y'$ by the dot product of the running class prototypes $\theta_R$ with the dense image features $f'$ of the augmented image $I'$.}
$\theta_R$ is optimized by applying a batch-wise focal loss \citep{Lin:2020:FLD} with respect to the pseudo labels $P^\ast$. The focal loss $\mathcal{L}_\text{focal}$ is a weighted version of the cross-entropy loss, increasing the loss contribution of less confident classes, \ie,
\begin{gather}
    \label{eq:focal}
    \mathcal{L}_\text{focal}(\theta_R; y') = -\!\sum_{k,i,j}(1 - \chi_k)^2 P^{\ast}_{k,i,j} \log(y'_{k,i,j}) \,,\raisetag{6pt}
\end{gather}
where $y'_{k,i,j}=\operatorname*{softmax}(\theta_R^\top f'_{:,i,j})$ are the predictions and $\chi_k$ is the class-wise confidence value approximated by averaging $y'_{k,:,:}$ spatially. The running class prototypes $\theta_R$ are optimized with an augmented input image $I'$. We employ photometric augmentations (Gaussian blur, grayscaling, and color jitter), introducing a controlled noise, thereby strengthening the robustness of our class representation.
\cng{The momentum class prototypes $\theta_M$ are the \ac{ema} of the running class prototypes $\theta_R$.} This is utilized in order to stabilize the optimization, accounting for the noisy nature of the unsupervised signal used for optimization. \rebut{We update $\theta_M$ every $\kappa$ iterations with a decay $\gamma$ as
\begin{equation}
 \label{eq:ema}
 \theta_M^{t+\kappa} = \gamma \theta_M^t + (1 - \gamma) \theta_R^{t+\kappa}\,,
\end{equation}}%
where $t$ is the iteration index of the previous update.
This optimization approach resembles \ourframework{}. 
Hereby, the E-step amounts to finding pseudo labels using \ourmethod and the momentum class prototypes. 
The M-step optimizes the running class prototypes \cng{with respect to their focal loss $ \mathcal{L}_\text{focal}$.}
Stochasticity arises from performing \ac{em} in mini-batches \rebuttwo{of images.}

\inparagraph{Inference.}
At inference time, we obtain the segmentation prediction \rebuttwo{$y$ \cngsr{via} the momentum class prototypes $\theta_M$}, refined using a fully connected \ac{crf} with Gaussian edge potentials \citep{Kraehenbuehl:2011:EIF} following previous approaches \citep{Gansbeke:2021:USS, Hamilton:2022:USS, Seong:2023:LHP}.
This is the identical \ac{crf} as already used for refining the masks in the \ourmethod pseudo-label generation using the same CRF parameters as previous work \citep{Gansbeke:2021:USS, Hamilton:2022:USS, Seong:2023:LHP}.        

\section{Experiments}
\label{sec:experiments}

To assess the efficacy of our approach, we compare it to the current state-of-the-art in unsupervised semantic segmentation. For a fair comparison, we closely follow the overall setup used by numerous previous works \citep{Ji:20219:IIC,Cho:2021:PUS,Hamilton:2022:USS,Seong:2023:LHP}.

\subsection{Experimental Setup}
\label{sec:experimental_setup}
\paragraph{Datasets.}
Following the practice of previous work, we conduct experiments on Cityscapes \citep{Cordts:2016:CDS}, COCO-Stuff \citep{Caesar:2018:CST}, and Potsdam-3 \citep{PotsdamDataset}. Cityscapes and COCO-Stuff are evaluated using 27 classes, while Potsdam is evaluated on the 3-class variant. Adopting the established evaluation protocol \citep{Ji:20219:IIC,Cho:2021:PUS,Hamilton:2022:USS,Seong:2023:LHP}, we resize images to $320$ pixels along the smaller axis and crop the center $320 \times 320$ pixels. This is adjusted to $322$ pixels for DINOv2. Different from previous work, we apply this simple scheme throughout this work, thus dispensing with elaborate multi-crop approaches of previous methods \citep{Hamilton:2022:USS, Yin:2022:TTA, Seong:2023:LHP}.

\inparagraph{Self-supervised backbone.}
Experiments are conducted across a collection of pre-trained self-supervised feature embeddings: DINO \citep{Caron:2021:EPS} based on ViT-Small and ViT-Base using $8\times8$ patches; and DINOv2 \citep{Oquab:2023:DLR} based on ViT-Small and ViT-Base using $14\times14$ patches.
In the spirit of SSL principles, we keep the backbone parameters frozen throughout the experiments.
We use the output from the last network layer as our SSL feature embeddings.
Since \ourmethodframework is agnostic to the used embedding space, we can also apply it on top of current state-of-the-art unsupervised segmentation pipelines. Here, we consider STEGO \citep{Hamilton:2022:USS} and HP \citep{Seong:2023:LHP}, which also use DINO features \cngtwo{but learn a target domain-specific subspace.}

\inparagraph{Baseline.}
Following \citep{Hamilton:2022:USS, Seong:2023:LHP}, we train a single linear layer as a baseline with the same structure as \cng{$\theta_R$ and $\theta_M$} by minimizing the cosine-distance batch-wise $K$-Means loss from \cref{eq:kmeans}. Hereby, parameters, such as the number of epochs and the learning rate, are identical to those used when employing \ourmethodframework.

\inparagraph{\cng{\ourmethodframework.}}
As discussed in \cref{sec:prima}, \cng{the momentum class prototypes $\theta_M$} are initialized using the first $K$ principal components; 
we use $2975$ images for PCA, as this is the largest number of training images shared by all datasets. 
Next, $\theta_M$ is pre-trained by minimizing \cref{eq:kmeans} using Adam \citep{Kingma:2015:AMS}.
We use a learning rate of 0.005 for 2 epochs on all datasets and backbones. The weights are then copied to $\theta_R$.
\cng{For fitting the running class prototypes using \ac{em}, $\theta_R$ is optimized} by minimizing the focal loss from \cref{eq:focal} with Adam \citep{Kingma:2015:AMS} using a learning rate of 0.005.
The momentum class prototypes $\theta_M$ are updated using an \ac{ema} according to \cref{eq:ema} every $\gamma_s=\text{10}$ steps with decay $\gamma_\psi=\text{0.98}$. We set the \ourmethod mask-proposal threshold to $\psi=\text{0.4}$ \rebut{and provide detailed ablation experiments in \cref{sec:thresh_ablation}.} We use a batch size of 32 for 50 epochs on Cityscapes and Potsdam-3, and use 5 epochs on COCO-Stuff due to its larger size. Importantly, the same hyperparameters are used across \emph{all} datasets and backbones. \cng{Moreover, note that fitting class prototypes with \ourmethodframework is quite practical, \eg, about 2 hours on Cityscapes.} Experiments are conducted on a single NVIDIA A6000 GPU.

\begin{table}[!t]
\small
\centering
\renewcommand{\arraystretch}{0.95}
    \caption{\textbf{Cityscapes -- \ourmethodframework \textit{(Ours)} comparison to existing unsupervised semantic segmentation methods}, using Accuracy and mean IoU (in \%) for unsupervised and supervised probing. Double citations refer to a method's origin and the work conducting the experiment.} \label{tab:main_cs}
    \begin{tabularx}{\linewidth}{lYcc>{\color{mygray}}c>{\color{mygray}}c}
        \toprule
        \multirow{2}{*}{\vspace{-0.55em}\textbf{Method}} & \multirow{2}{*}{\vspace{-0.55em}\textbf{Backbone}}  & \multicolumn{2}{c}{\textit{Unsupervised}} & \multicolumn{2}{c}{\textit{\gy{Supervised}}} \\
        \cmidrule(l{0.5em}r{0.5em}){3-4} \cmidrule(l{0.5em}r{0.5em}){5-6}
        & & \textbf{Acc} & \textbf{mIoU} & \textbf{Acc} & \textbf{mIoU} \\
        \midrule
        IIC \citep{Ji:20219:IIC, Cho:2021:PUS}               &        & 47.9     &  \phantom{0}6.4   & --     & -- \\ 
        MDC \citep{Caron:2018:DCU, Cho:2021:PUS}             &        & 40.7     &  \phantom{0}7.1   & --     & -- \\ 
        PiCIE \citep{Cho:2021:PUS}                           &        & 65.5 & 12.3  & --     & -- \\ 
        VICE \citep{Karlsson:2022:VID}                       & \multirow{-4}{*}{\shortstack{ResNet18\\[0pt]+FPN}}       & 31.9 & 12.8 & 86.3 & 31.6 \\ 
        \midrule 
        Baseline \citep{Caron:2021:EPS}                      &        & 61.4      & 15.8 & 91.0 & 35.4 \\
       +\,TransFGU \citep{Yin:2022:TTA}                      &        & 77.9      & 16.8 & -- & -- \\
       +\,HP \citep{Seong:2023:LHP}                          &        & 80.1      & 18.4 & 91.2 & 30.6 \\
        \rowcolor{mylightgray}            
       +\,\ourmethodframework                               &      &\bn{81.2}  & \bn{19.4} & 91.0 & 35.4\\
        \rowcolor{mylightgray}  
       +\,HP \citep{Seong:2023:LHP}\,+\,\ourmethodframework  & \multirow{-5}{*}{\shortstack{DINO\\[0pt]ViT-S/8}}      & 76.6  & 19.2 & 91.2 & 30.6\\
        \midrule 

        Baseline \citep{Caron:2021:EPS}                 &        & 49.2      & 15.5 & 91.6 & 35.9\\
       +\,STEGO \citep{Hamilton:2022:USS, Koenig:2023:UIW}     &        & 73.2      & 21.0 & 89.6 & 28.0 \\
       +\,HP \citep{Seong:2023:LHP}                            &        & \bn{79.5}      & 18.4 & 90.9 & 33.0 \\
        \rowcolor{mylightgray}
       +\,\ourmethodframework                          &      & 59.6  & 17.6 & 91.6 & 35.9 \\
        \rowcolor{mylightgray}           
        +\,STEGO \citep{Hamilton:2022:USS}\,+\,\ourmethodframework         & \multirow{-5}{*}{\shortstack{DINO\\[0pt]ViT-B/8}}         & 78.6     &\bn{21.6}  & 89.6 & 28.0 \\
        \midrule 
        Baseline \citep{Oquab:2023:DLR}               &      & 49.5      & 15.3  & 90.8 & 41.9 \\
        \rowcolor{mylightgray}
        +\,\ourmethodframework                          & \multirow{-2}{*}{\shortstack{DINOv2\\[0pt]ViT-S/14}}      &\bn{71.5}  &\bn{19.0}  & 90.8 & 41.9 \\
        \midrule 
        Baseline \citep{Oquab:2023:DLR}               &       & 36.1      & 14.9 & 91.0 & 44.8 \\
        \rowcolor{mylightgray}
        +\,\ourmethodframework                         & \multirow{-2}{*}{\shortstack{DINOv2\\[0pt]ViT-B/14}}      &\bn{82.9}  &\bn{21.3} & 91.0 & 44.8 \\
        \bottomrule
    \end{tabularx}
    \vspace{-0.5em}
\end{table}

\inparagraph{Supervised upper bounds.}
To assess the potential of the SSL features used, we report supervised upper bounds. Specifically, we train a linear layer using cross entropy and Adam with a learning rate of 0.005. Since \ourmethodframework uses frozen SSL features, its supervised bound is the same as that of the underlying features.
This is not the case, however, for prior work \citep{Hamilton:2022:USS, Seong:2023:LHP}, which project the feature representation affecting the upper bound. 

\inparagraph{Evaluation.}
For inference, we use the prediction from the momentum class prototypes \cng{$\theta_M$}.
\ac{crf} refinement uses 10 inference iterations and standard parameters $a\!=\!4, b\!=\!3, \theta_\alpha\!=\!67, \theta_\beta\!=\!3, \theta_\gamma\!=\!1$ from prior work \citep{Gansbeke:2021:USS, Hamilton:2022:USS, Seong:2023:LHP}. 
We evaluate common metrics in unsupervised semantic segmentation, specifically the \ac{miou} and \ac{acc} over all classes after aligning the predicted class IDs with ground-truth labels by means of Hungarian matching \citep{Kuhn:1955:THM}. 

\inparagraph{\cng{SotA + \ourmethodframework.}}
To explore our method's potential, we additionally employ \ourmethodframework on top of STEGO~\citep{Hamilton:2022:USS} and HP~\citep{Seong:2023:LHP}. For each backbone-dataset combination, we apply it on top of the best previous method in terms of \ac{miou}. To that end, the training signal for learning the feature projection of \citep{Hamilton:2022:USS, Seong:2023:LHP} remains unchanged. We apply \cng{\ourmethodframework} fully orthogonally, using the DINO backbone features for pseudo-label generation and fit a direct connection between the feature space of the state-of-the-art method and the prediction space.

\begin{table}[!t]
\small
\renewcommand{\arraystretch}{0.95}
\centering
    \caption{\textbf{COCO-Stuff -- \ourmethodframework \textit{(Ours)} comparison to existing unsupervised semantic segmentation methods}, using Accuracy and mean IoU (in \%) for unsupervised and supervised probing. Double citations refer to a method's origin and the work conducting the experiment.
    \label{tab:main_coco}}
    \begin{tabularx}{\linewidth}{lYcc>{\color{mygray}}c>{\color{mygray}}c}
        \toprule
        \multirow{2}{*}{\vspace{-0.55em}\textbf{Method}} & \multirow{2}{*}{\vspace{-0.55em}\textbf{Backbone}}  & \multicolumn{2}{c}{\textit{Unsupervised}} & \multicolumn{2}{c}{\textit{\gy{Supervised}}} \\
        \cmidrule(l{0.5em}r{0.5em}){3-4} \cmidrule(l{0.5em}r{0.5em}){5-6}
        & & \textbf{Acc} & \textbf{mIoU} & \textbf{Acc} & \textbf{mIoU} \\
        \midrule
        IIC \citep{Ji:20219:IIC, Cho:2021:PUS}           &       & 21.8      &  \phantom{0}6.7      & 44.5      &  \phantom{0}8.4 \\ 
        MDC \citep{Caron:2018:DCU, Cho:2021:PUS}         &       & 32.2      &  \phantom{0}9.8      & 48.6      & 13.3 \\ 
        PiCIE \citep{Cho:2021:PUS}                       &       & 48.1      & 13.8      & 54.2      & 13.9 \\
        PiCIE+H \citep{Cho:2021:PUS}                     &       & 50.0  & 14.4  & 54.8      & 14.8 \\
        VICE \citep{Karlsson:2022:VID}                   & \multirow{-5}{*}{\shortstack{ResNet18\\[0pt]+FPN}}       & 28.9      & 11.4      & 62.8  & 25.5 \\ 
        \midrule 
        Baseline \citep{Caron:2021:EPS}             &       & 34.2      & \phantom{0}9.5       & 72.0      & 41.3 \\
       +\,TransFGU \citep{Yin:2022:TTA}                  &       & 52.7      & 17.5      & --         & -- \\
       +\,STEGO \citep{Hamilton:2022:USS}                &       & 48.3      & 24.5      & 74.4      & 38.3 \\
       +\,ACSeg \citep{Li:2023DCN}                       &       & --         & 16.4      & --         & -- \\

       +\,HP \citep{Seong:2023:LHP}                      &       & 57.2      & 24.6      & 75.6      & 42.7 \\
        \rowcolor{mylightgray}
       +\,\ourmethodframework                    &    & 46.5      & 16.4      & 72.0   & 41.3  \\
        \rowcolor{mylightgray}
       +\,HP \citep{Seong:2023:LHP}\,+\,\ourmethodframework    & \multirow{-7}{*}{\shortstack{DINO\\[0pt]ViT-S/8}}        & \bn{57.8} & \bn{25.1}  & 75.6    & 42.7 \\
        \midrule 
        Baseline \citep{Caron:2021:EPS}             &       & 38.8      & 15.7      & 74.0      & 44.6 \\
       +\,STEGO \citep{Hamilton:2022:USS}                &        & 56.9      & 28.2      & 76.1  & 41.0 \\
        \rowcolor{mylightgray}
       +\,\ourmethodframework                      &       & 48.5      & 21.9      & 74.0  & 44.6 \\
        \rowcolor{mylightgray}
       +\,STEGO \citep{Hamilton:2022:USS}\,+\,\ourmethodframework        & \multirow{-4}{*}{\shortstack{DINO\\[0pt]ViT-B/8}}       &\bn{57.9}  &\bn{29.7}  & 76.1  & 41.0  \\
        \midrule 
        Baseline \citep{Oquab:2023:DLR}           &      & 44.5      & 22.9      & 77.9   & 52.8 \\
        \rowcolor{mylightgray}
        +\,\ourmethodframework                      & \multirow{-2}{*}{\shortstack{DINOv2\\[0pt]ViT-S/14}}      & \bn{46.5}      & \bn{23.8}          & 77.9 & 52.8 \\
        \midrule 
        Baseline \citep{Oquab:2023:DLR}           &       & 35.0      & 17.9      & 77.3  & 53.7 \\
        \rowcolor{mylightgray}
        +\,\ourmethodframework                      & \multirow{-2}{*}{\shortstack{DINOv2\\[0pt]ViT-B/14}}      & \bn{52.8}      & \bn{23.6}      & 77.3 & 53.7 \\
        \bottomrule
    \end{tabularx}
    \vspace{-1.5em}
\end{table}

\subsection{Results}
We compare \ourmethodframework against prior work for unsupervised semantic segmentation \citep{Ji:20219:IIC, Cho:2021:PUS, Hamilton:2022:USS, Yin:2022:TTA, Li:2023DCN, Seong:2023:LHP}.
As in previous work, we use DINO~\citep{Caron:2021:EPS} as the main baseline.
Additionally, we also test \ourmethodframework on top of DINOv2~\citep{Oquab:2023:DLR}, STEGO~\citep{Hamilton:2022:USS}, and HP~\citep{Seong:2023:LHP}.
Overall, we observe that the DINO baseline already achieves strong results (\cf \cref{tab:main_cs,tab:main_coco,tab:main_pd}). 
\cng{DINOv2 features significantly raise the supervised upper bounds in terms of \ac{acc} and \ac{miou}, the improvement in the unsupervised case remains more modest. Nevertheless, \ourmethodframework further boosts the unsupervised segmentation performance.}

In \cref{tab:main_cs}, we compare to previous work on the Cityscapes dataset.
\ourmethodframework leads to a consistent improvement over all baselines in terms of unsupervised segmentation accuracy. 
For example, \ourmethodframework boosts DINO ViT-S/8 by $+3.6\%$ and $+19.8\%$ in terms of \ac{miou} and \ac{acc}, respectively, which leads to state-of-the-art performance. 
Notably, we find \ourmethodframework to be complementary to other state-of-the-art unsupervised segmentation methods like STEGO~\citep{Hamilton:2022:USS} and HP~\citep{Seong:2023:LHP} on the corresponding backbone model. 
This suggests that these methods use their SSL representation only to a limited extent and do not fully leverage the inherent properties of the underlying SSL embeddings. 
Similar observations can be drawn for the experiments on COCO-Stuff in \cref{tab:main_coco}. \ourmethodframework leads to a consistent improvement across all four SSL baselines, as well as an improvement over STEGO and HP. For instance, combining STEGO with \ourmethodframework leads to $+14.0\%$ and $+19.1\%$ improvement over the baseline in terms of \ac{miou} and \ac{acc} for DINO ViT-B/8. 
Experiments on the Potsdam-3 dataset follow the same pattern (\cf \cref{tab:main_pd}). \ourmethodframework leads to a consistent gain over the baseline, \eg $+17.6\%$ and $+14.4\%$ in terms of \ac{miou} and \ac{acc}, respectively, for DINO ViT-B/8. Moreover, it also boosts the accuracy of STEGO and HP. 
\cng{In some cases, the gain of \ourmethodframework is limited. For example, in \cref{tab:main_cs} for DINO ViT-B/8 + \ourmethodframework, the class prototype for ``sidewalk'' is poor while the classes ``road'' and ``vegetation'' superimpose smaller objects. For DINO ViT-S/8 + \ourmethodframework in \cref{tab:main_pd}, the class prototype ``road'' is poor. This limits the overall performance of our method while still outperforming the respective baseline in both cases.}

\begin{table}[!t]
\small
\renewcommand{\arraystretch}{0.95}
\centering
    \caption{\textbf{Potsdam-3 -- \ourmethodframework \textit{(Ours)} comparison to existing unsupervised semantic segmentation methods}, using Accuracy and mean IoU (in \%) for unsupervised and supervised probing. Double citations refer to a method's origin and the work conducting the experiment. \label{tab:main_pd}}
    \begin{tabularx}{\linewidth}{lYcc>{\color{mygray}}c>{\color{mygray}}c}
        \toprule
        \multirow{2}{*}{\vspace{-0.55em}\textbf{Method}} & \multirow{2}{*}{\vspace{-0.55em}\textbf{Backbone}}  & \multicolumn{2}{c}{\textit{Unsupervised}} & \multicolumn{2}{c}{\textit{\gy{Supervised}}} \\
        \cmidrule(l{0.5em}r{0.5em}){3-4} \cmidrule(l{0.5em}r{0.5em}){5-6}
        & & \textbf{Acc} & \textbf{mIoU} & \textbf{Acc} & \textbf{mIoU} \\
        \midrule
        RandomCNN \citep{Cho:2021:PUS}                            &      & 38.2       & -- & -- & -- \\ 
        K-Means \citep{Pedregosa:2011:SML, Cho:2021:PUS}          &      & 45.7       & -- & -- & -- \\ 
        SIFT \citep{Lowe:2004:DIF, Cho:2021:PUS}                  &      & 38.2       & -- & -- & -- \\  
        ContextPrediction \citep{Doersch:2015:UVR, Cho:2021:PUS}  &      & 49.6       & -- & -- & -- \\  
        CC \citep{Isola:2015:LVG, Cho:2021:PUS}                   &      & 63.9       & -- & -- & -- \\
        DeepCluster \citep{Caron:2018:DCU, Cho:2021:PUS}          &      & 41.7       & -- & -- & -- \\   
        IIC \citep{Ji:20219:IIC, Cho:2021:PUS}                    & \multirow{-7}{*}{\shortstack{VGG\\[0pt]11}}     & 65.1   & -- & -- & -- \\  
        \midrule 
        Baseline \citep{Caron:2021:EPS}                            &  & 56.6       & 33.6	    & 82.0       & 69.0 \\
        +\,STEGO \citep{Hamilton:2022:USS, Koenig:2023:UIW}        &  & 77.0       & 62.6      & 85.9  & 74.8 \\
        \rowcolor{mylightgray}
       +\,\ourmethodframework                                      &  & 62.5       & 38.9      & 82.0  & 69.0 \\
        \rowcolor{mylightgray}
        +\,STEGO \citep{Hamilton:2022:USS}\,+\,\ourmethodframework                                  & \multirow{-4}{*}{\shortstack{DINO\\[0pt]ViT-S/8}}    &\bn{78.4}   &\bn{64.2}  & 85.9  & 74.8 \\
        \midrule 
        Baseline \citep{Caron:2021:EPS}                      &    & 66.1       & 49.4      & 84.3     & 72.8 \\
        +\,HP \citep{Seong:2023:LHP}                               &    & 82.4  & 69.1 & 88.0 & 78.4 \\ 
        \rowcolor{mylightgray}
        +\,\ourmethodframework                               &   & 80.5   & 67.0  & 84.3 & 72.8 \\
        \rowcolor{mylightgray}
       +\,HP \citep{Seong:2023:LHP}+\,\ourmethodframework    & \multirow{-4}{*}{\shortstack{DINO\\[0pt]ViT-B/8}}   &\bn{83.3}   &\bn{71.0}  & 88.0 & 78.4 \\
        \midrule 
        Baseline \citep{Oquab:2023:DLR}                    &  & 75.9       & 61.0	& 86.6 & 76.2 \\
        \rowcolor{mylightgray}
        +\,\ourmethodframework                               & \multirow{-2}{*}{\shortstack{DINOv2\\[0pt]ViT-S/14}}  &\bn{78.5}   & \bn{64.3}  & 86.6 & 76.2 \\
        \midrule 
        Baseline \citep{Oquab:2023:DLR}                    &   & 82.4        & 69.9       & 87.9   & 78.3  \\
        \rowcolor{mylightgray}
        +\,\ourmethodframework                               & \multirow{-2}{*}{\shortstack{DINOv2\\[0pt]ViT-B/14}}   & \bn{83.2}   &\bn{71.1}  & 87.9	& 78.3 \\
        \bottomrule
    \end{tabularx}
\end{table}

Overall, \cng{\ourmethodframework provides modest but consistent} benefits over a wide range of baselines and datasets and reaches competitive segmentation \cngtwo{performance} \wrt the state-of-the-art \cng{using identical hyperparameters across \emph{all} backbones and datasets.}
Recalling the simplicity of the techniques behind \ourmethod, we believe that this is a significant result. 
The complementary effect of \ourmethodframework on other state-of-the-art methods (STEGO, HP) further suggests that they rely on DINO features for mere ``bootstrapping'' and learn feature representations with orthogonal properties to those of DINO. \cng{We conclude} that \ourmethodframework constitutes a straightforward, entirely orthogonal tool for boosting unsupervised semantic segmentation. 

\subsection{Ablation Study}
\cng{To untangle the factors behind \ourmethodframework, we examine the individual components in a variety of ablation experiments to access the contribution.}

\begin{table}[!t]
\vspace{0.5em}
\caption{\textbf{Ablation study} analyzing design choices and components in the \ourmethod pseudo-label generation \emph{(a)} and \ourmethodframework \emph{(b)} for COCO-Stuff using DINO ViT-B/8.} \label{tab:ablation}
\small
\renewcommand{\arraystretch}{0.95}
\centering
    \subcaptionbox{\ourmethod pseudo-label ablation \label{tab:ablation-a}}{
        \begin{tabularx}{0.48\linewidth}{@{}Xcc@{}}
            \toprule
            \textbf{Method} & \textbf{Acc} & \textbf{mIoU} \\
            \midrule
            Baseline \citep{Caron:2021:EPS}          & 38.8  & 15.7 \\
            Similarity Masks                         & 46.3  & 19.8 \\  
            +\,NN                                    & 44.9  & 20.0 \\            
            +\,P-CRF ($\equiv$ \ourmethodframework)  & 48.4  & 21.9 \\   
            \midrule
            \ourmethodframework \rebut{(non-iter.\ PC)}         & 47.9  & 21.7 \\ 
            \bottomrule
    \end{tabularx} 
    } \hfill
    \subcaptionbox{\ourmethodframework ablation \label{tab:ablation-b}}{%
        \begin{tabularx}{0.48\linewidth}{@{}Xcc@{}}
            \toprule
            \textbf{Method} & \textbf{Acc} & \textbf{mIoU} \\
            \midrule 
            Baseline \citep{Caron:2021:EPS}         & 38.8  & 15.7 \\ [1.25pt]
           +\,\ourmethod pseudo label               & 38.8  & 18.0 \\ [1.25pt] 
           +\,EMA                                   & 45.0  & 20.2 \\ [1.25pt] 
           +\,Augment                               & 46.0  & 20.4 \\ [1.25pt] 
           +\,CRF ($\equiv$ \ourmethodframework)    & 48.4  & 21.9 \\ 
            \bottomrule
        \end{tabularx}}
\end{table}

\inparagraph{\ourmethod pseudo-label ablations.} 
In \cref{tab:ablation-a}, we analyze the contribution of the individual sub-steps for \ourmethod pseudo-label generation by increasing the complexity \cng{of label generation.}
We provide the DINO baseline, which corresponds to $K$-means feature clustering, for reference.
In the most simplified case, we directly use the similarity mask, similar to \cref{eq:nn}. 
Next, we use the nearest neighbor (+NN in \cref{tab:ablation-a}) of the principal component to get the masks as in \cref{eq:similarity-map}, followed by the full approach with \ac{crf} refinement (+P-CRF). Except for the changes in the pseudo-label generation, the \cng{optimization} remains as described in \cref{sec:experimental_setup}. We observe that the similarity masks already provide a good staring point, yet we identify a gain from every single component step. 
This suggests that using the nearest neighbor improves the localization of the similarity mask.
Similarly, \ac{crf} refinement improves the alignment between the masks and the image content.
We also experiment with using the respective next principal \rebut{component (non-iter.\ PC) instead of iteratively extracting the first principal component from the masked features. This leads to slightly inferior results. Naively using the K leading eigenvectors and simply assigning the masks based on the $\arg \max$ of their cosine similarity to the features without any iterations would lead to significantly worse results with a pixel Accuracy of $43.1$ \% and a mIoU of $19.9$ \%. Note that nearest neighbor anchoring and thresholding are used in both experiments. Additionally, we ablate the nearest neighbor anchoring, the threshold $\psi$, and the stop criterion in \cref{sec:furtheranalysis}.}

\inparagraph{\ourmethodframework architecture ablations.}
In a similar vein, we analyze the contribution of the different architectural components of \ourmethodframework. \cng{Optimizing over a single set of class prototypes using the proposed \ourmethod pseudo labels already provides moderate improvement} (+\ourmethod pseudo label in \cref{tab:ablation-b}), despite the disadvantage of an unstable and noisy optimization. 
Adding the \ac{ema} (+EMA) leads to a more stable optimization and further improved segmentation. Augmenting the input (+Augment) results in a further gradual improvement. \rebut{We provide a more detailed breakdown of the augmentations in \cref{sec:ablation_extendedaugs}.} Similarly, refining the prediction with a \ac{crf} improves the results further (+CRF).

\begin{table}[!t]
\small
\renewcommand{\arraystretch}{0.95}
\centering
    \caption{\textbf{Oracle quality assessment of \ourmethod pseudo labels} for Cityscapes, COCO-Stuff, and Potsdam-3 by assigning oracle class IDs to the masks.
    ``Pseudo'' refers to evaluating only the pixels contained in the pseudo label, ``All'' to evaluating including the ``ignore'' assignments of the pseudo label. \label{tab:pseudo_oracle}}
    \begin{tabularx}{\linewidth}{@{}Xcccccc@{}}
        \toprule
        \multirow{2}{*}{\vspace{-0.55em}\textbf{Method}} & \multicolumn{2}{c}{\textit{Cityscapes}} & \multicolumn{2}{c}{\textit{COCO-Stuff}} & \multicolumn{2}{c}{\textit{Potsdam-3}} \\
        \cmidrule(l{0.5em}r{0.5em}){2-3} \cmidrule(l{0.5em}r{0.5em}){4-5} \cmidrule(l{0.5em}r{0.5em}){6-7}
        & \textbf{Acc} & \textbf{mIoU} & \textbf{Acc} & \textbf{mIoU}  & \textbf{Acc} & \textbf{mIoU}  \\
        \midrule 
        Pseudo                               & 92.4 & 54.0 & 93.4 & 82.4 & 95.2 & 90.9 \\
        All                                  & 73.2 & 32.4 & 74.1 & 55.9 & 67.4 & 48.9 \\
        \midrule
        DINO ViT-B/8 Baseline \citep{Caron:2021:EPS}  & 49.2 & 15.5 & 38.8 & 15.7 & 66.1 & 49.4 \\
        \bottomrule
    \end{tabularx}
    \vspace{-1.0em}
\end{table}

\inparagraph{Assessing \ourmethod pseudo labels. \label{sec:quanti_pseudolabels}} 
To estimate the quality of the pseudo labels, respectively the principal masks, we decouple those from the class ID assignment by providing the oracle ground-truth class for each mask in \cref{tab:pseudo_oracle}. 
To that end, we evaluate all pixels included in our pseudo labels (``Pseudo''), corresponding to the upper bound of our \cng{optimization signal}. Furthermore, we evaluate ``All'' by assigning the ``ignore'' pixels to a wrong class. The results indicate a high quality of the pseudo-label maps. \rebut{We show qualitative examples of the \ourmethod mask proposals and pseudo labels in \cref{sec:qual_pseudolabels}.}

\def\imgwidth{0.1035\linewidth}
\begin{figure}[!t]
    \small
    \smallskip
    \setlength\tabcolsep{1.0pt}
    \renewcommand{\arraystretch}{0.6666}
    \centering

    \begin{tabularx}{\linewidth}{@{}Xccccccccc@{}}
    & \multicolumn{3}{c}{Cityscapes} & \multicolumn{3}{c}{COCO-Stuff} & \multicolumn{3}{c}{Potsdam-3} \\
    \cmidrule(l{0.5em}r{0.5em}){2-4} \cmidrule(l{0.5em}r{0.5em}){5-7} \cmidrule(l{0.5em}r{0.5em}){8-10}
    
    \rotatebox[origin=lB]{90}{\scriptsize{\hspace{-31.7em}\shortstack{STEGO +\\ \ourmethodframework}\hspace{1.2em} \shortstack{\vspace{0.4em}STEGO \hspace{1.2em} \ourmethodframework \hspace{1.0em} Baseline \hspace{1.2em}Ground Truth \hspace{1.2em} Image}}}
    & \includegraphics[width=\imgwidth]
    {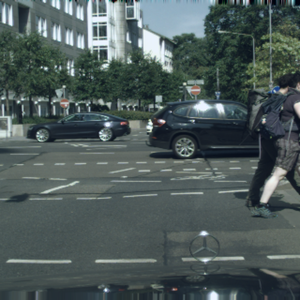} 
    & \includegraphics[width=\imgwidth]
    {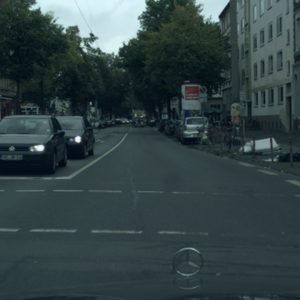} 
    & \includegraphics[width=\imgwidth]
    {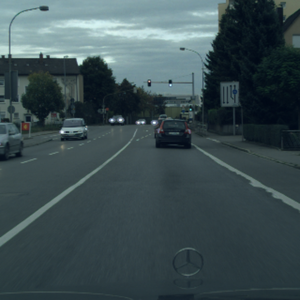} 
    & \includegraphics[width=\imgwidth]
    {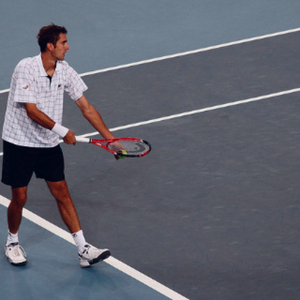} 
    & \includegraphics[width=\imgwidth]
    {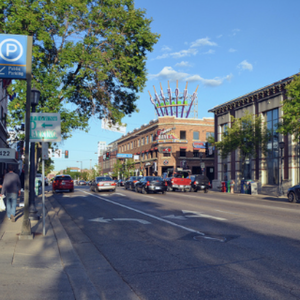} 
    & \includegraphics[width=\imgwidth]
    {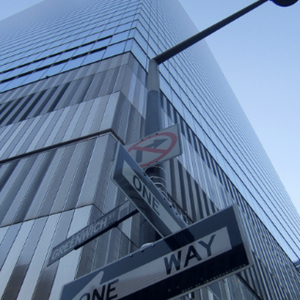} 
    & \includegraphics[width=\imgwidth]
    {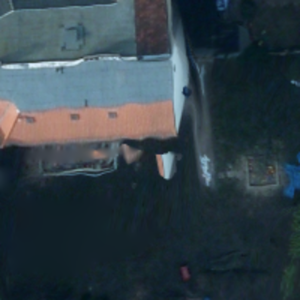}  
    & \includegraphics[width=\imgwidth]
    {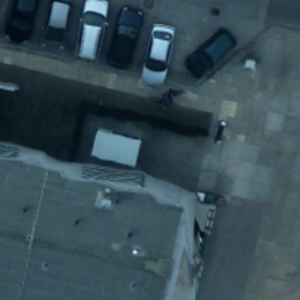} 
    & \includegraphics[width=\imgwidth]
    {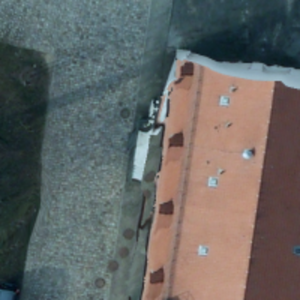} \\
    & \includegraphics[width=\imgwidth]
    {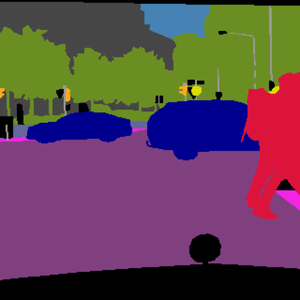} 
    & \includegraphics[width=\imgwidth]
    {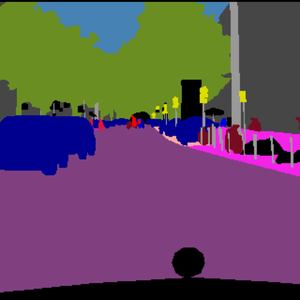}  
    & \includegraphics[width=\imgwidth]
    {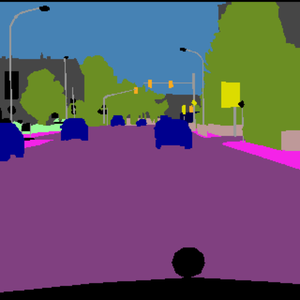}  
    & \includegraphics[width=\imgwidth]
    {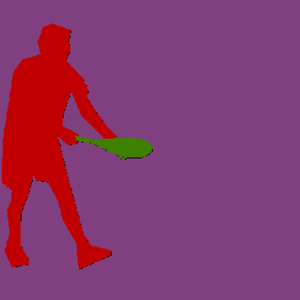}  
    & \includegraphics[width=\imgwidth]
    {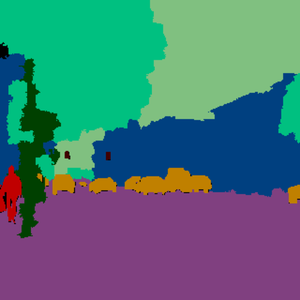}  
    & \includegraphics[width=\imgwidth]
    {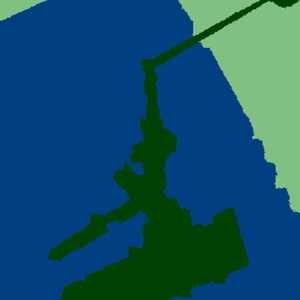}  
    & \includegraphics[width=\imgwidth]
    {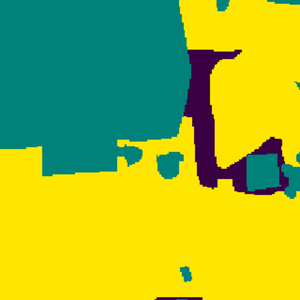}  
    & \includegraphics[width=\imgwidth]
    {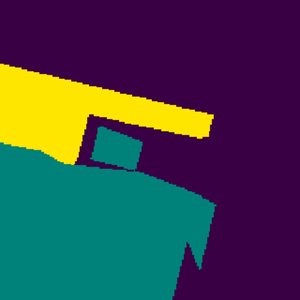}  
    & \includegraphics[width=\imgwidth]
    {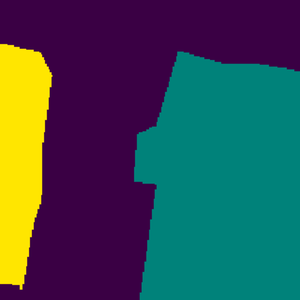}  \\
    & \includegraphics[width=\imgwidth]
    {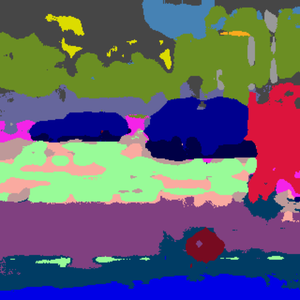} 
    & \includegraphics[width=\imgwidth]
    {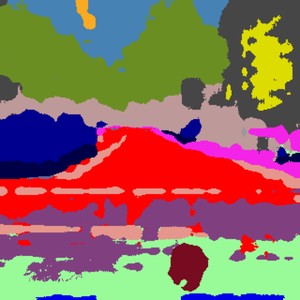} 
    & \includegraphics[width=\imgwidth]
    {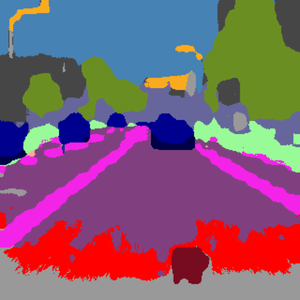} 
    & \includegraphics[width=\imgwidth]
    {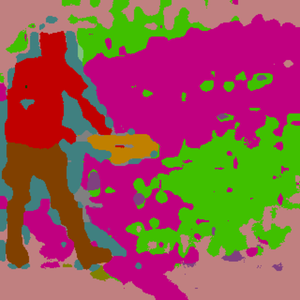} 
    & \includegraphics[width=\imgwidth]
    {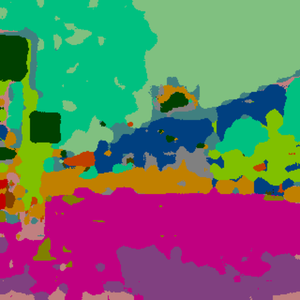} 
    & \includegraphics[width=\imgwidth]
    {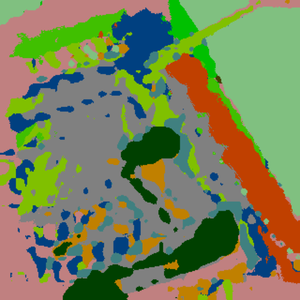} 
    & \includegraphics[width=\imgwidth]
    {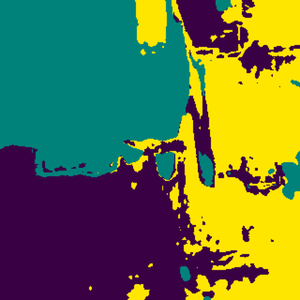} 
    & \includegraphics[width=\imgwidth]
    {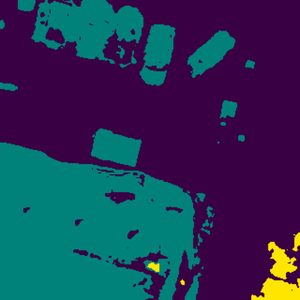} 
    & \includegraphics[width=\imgwidth]
    {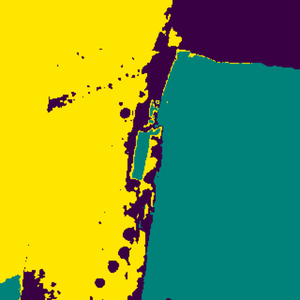} \\
    & \includegraphics[width=\imgwidth]
    {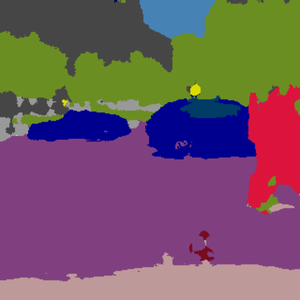} 
    & \includegraphics[width=\imgwidth]
    {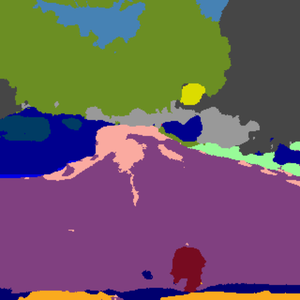} 
    & \includegraphics[width=\imgwidth]
    {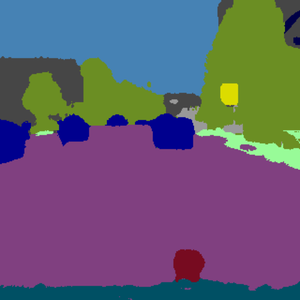} 
    & \includegraphics[width=\imgwidth]
    {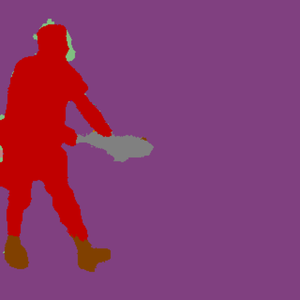} 
    & \includegraphics[width=\imgwidth]
    {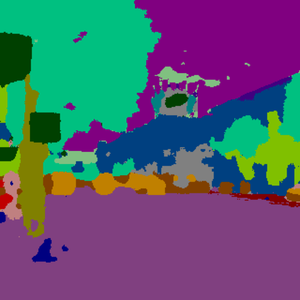} 
    & \includegraphics[width=\imgwidth]
    {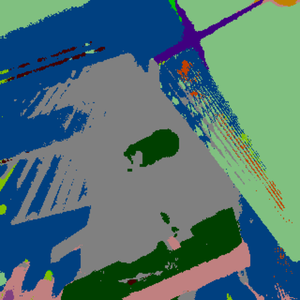} 
    & \includegraphics[width=\imgwidth]
    {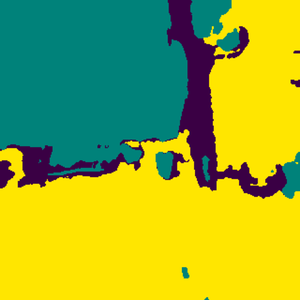} 
    & \includegraphics[width=\imgwidth]
    {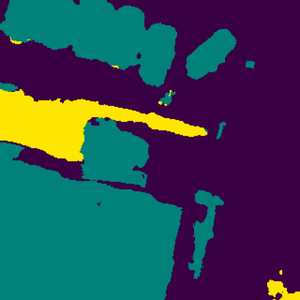} 
    & \includegraphics[width=\imgwidth]
    {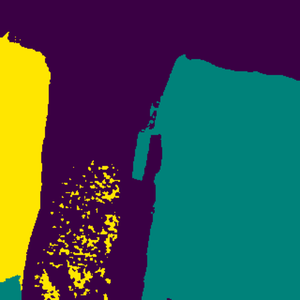} \\
    & \includegraphics[width=\imgwidth]
    {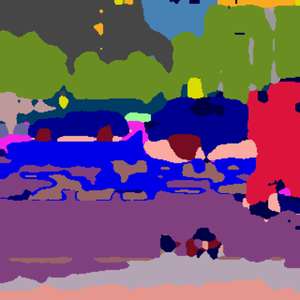} 
    & \includegraphics[width=\imgwidth]
    {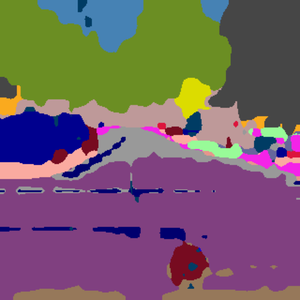} 
    & \includegraphics[width=\imgwidth]
    {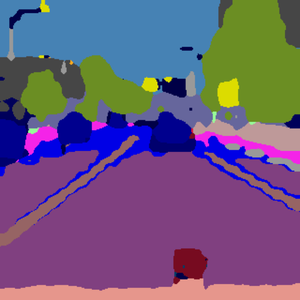} 
    & \includegraphics[width=\imgwidth]
    {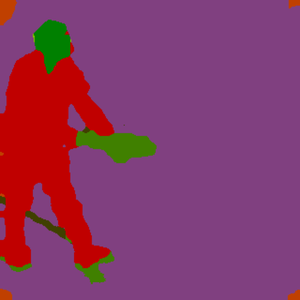} 
    & \includegraphics[width=\imgwidth]
    {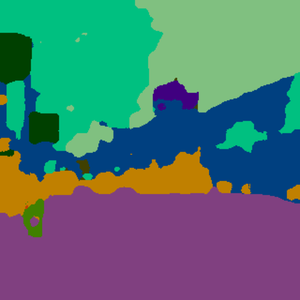} 
    & \includegraphics[width=\imgwidth]
    {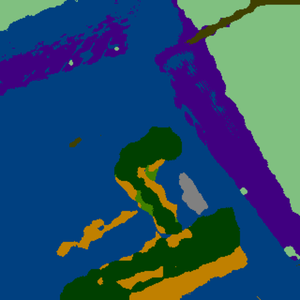} 
    & \includegraphics[width=\imgwidth]
    {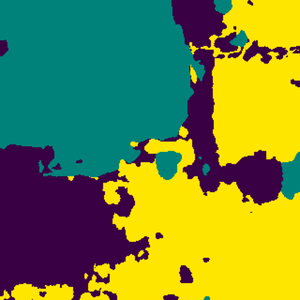} 
    & \includegraphics[width=\imgwidth]
    {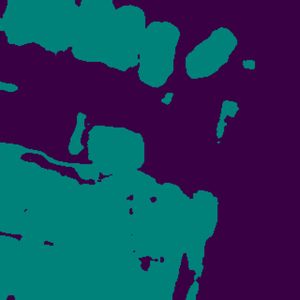} 
    & \includegraphics[width=\imgwidth]
    {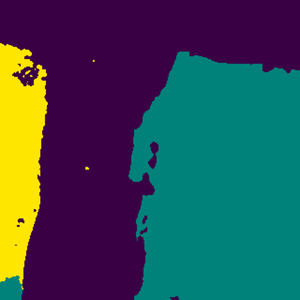} \\
    
    & \includegraphics[width=\imgwidth]
    {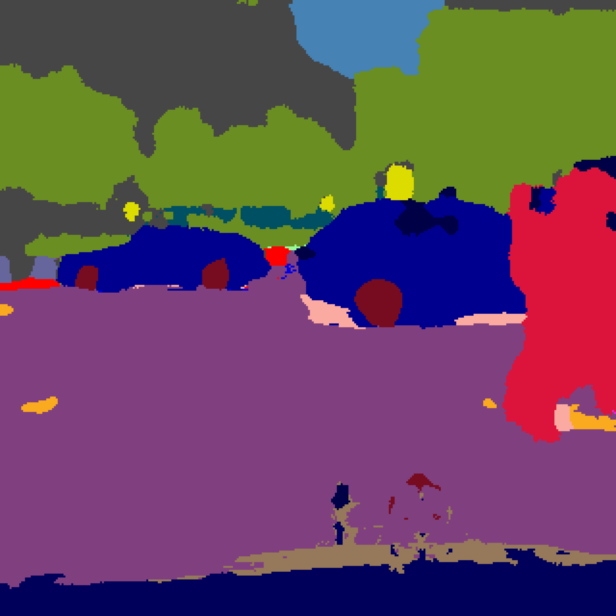} 
    & \includegraphics[width=\imgwidth]
    {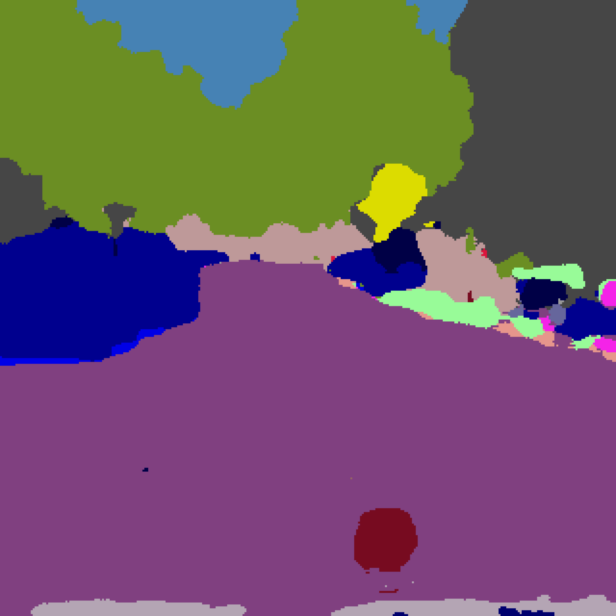} 
    & \includegraphics[width=\imgwidth]
    {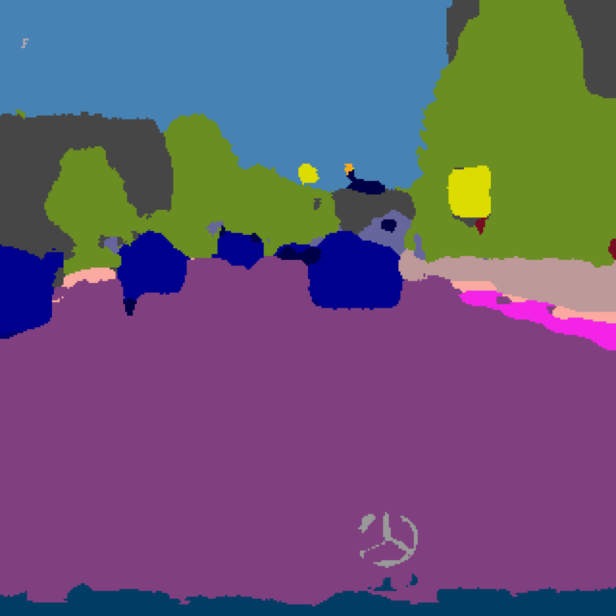} 
    & \includegraphics[width=\imgwidth]
    {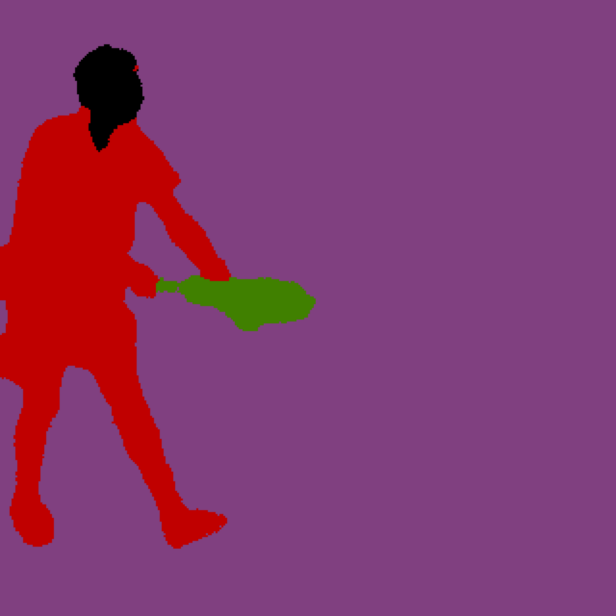} 
    & \includegraphics[width=\imgwidth]
    {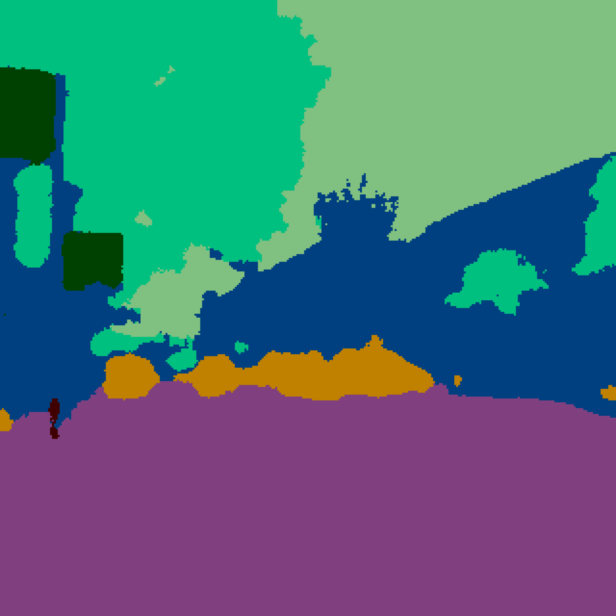} 
    & \includegraphics[width=\imgwidth]
    {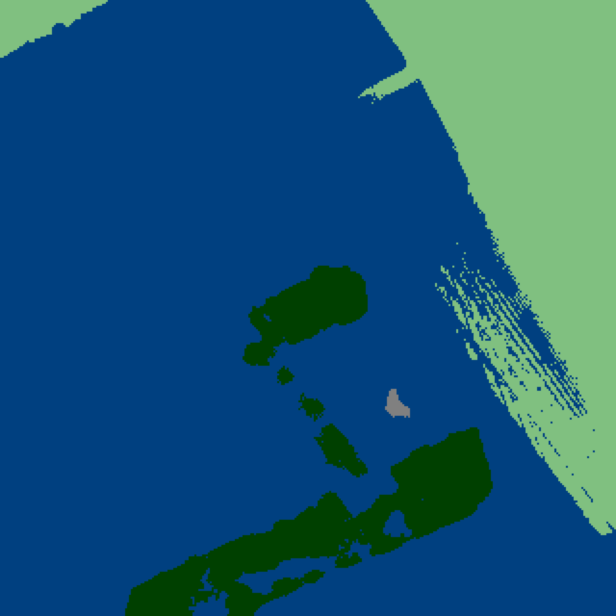} 
    & \includegraphics[width=\imgwidth]
    {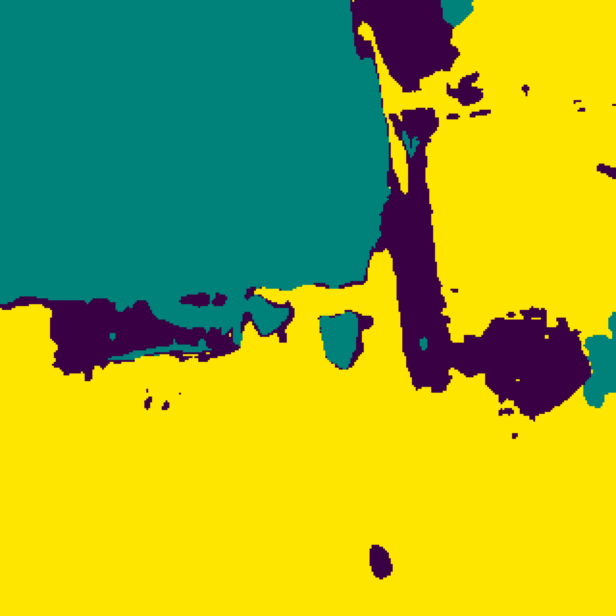}  
    & \includegraphics[width=\imgwidth]
    {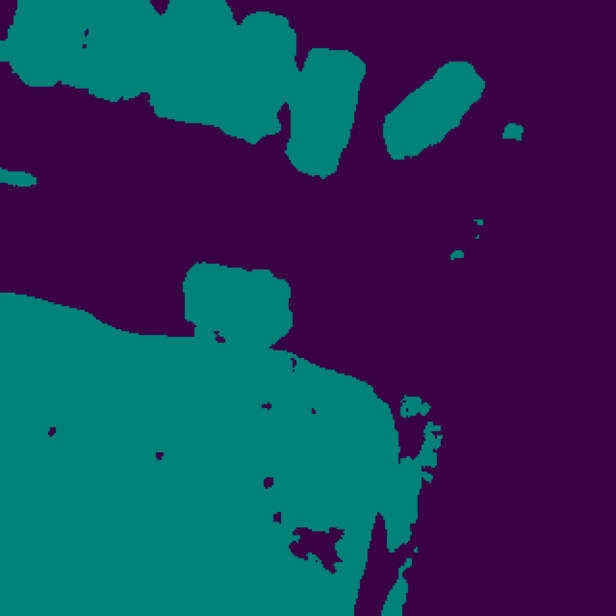} 
    & \includegraphics[width=\imgwidth]
    {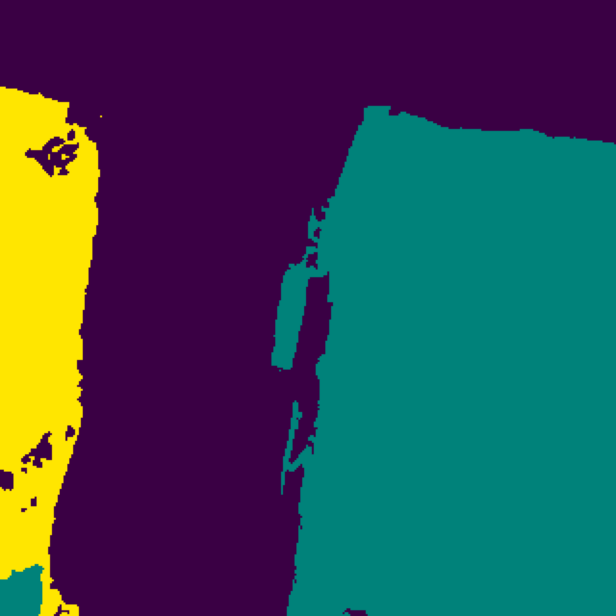} \\
    \end{tabularx}

    \caption{\textbf{Qualitative results} for the DINO ViT-B/8 baseline, \ourmethodframework \textit{(Ours)}, STEGO \citep{Hamilton:2022:USS}, and STEGO+\ourmethodframework \textit{(Ours)} for Cityscapes, COCO-Stuff, and Potsdam-3. Our method produces locally more consistent segmentation results reducing overall misclassification compared to the corresponding baseline. \label{fig:qualitative}}
\end{figure}

\inparagraph{Qualitative results.} 
We show qualitative results for Cityscapes, COCO-Stuff, and Potsdam-3 in \cref{fig:qualitative}. We observe that \ourmethodframework leads to less noisy results compared to the baseline, showcasing an improved local consistency of the segmentation and reduced mis-classification. The comparison with STEGO as a baseline exhibits a similar trend. For further examples and comparisons with HP, please refer to \cngtwo{\cref{sec:qual_comp_hp}}.

\inparagraph{Limitations.} 
One of the main challenges is to distinguish between classes that \cng{happen to share highly similar SSL feature representations. This is hardly avoidable if the feature representation is fixed, as was the case here and in previous work \citep{Hamilton:2022:USS,Seong:2023:LHP}.
Another limitation across existing unsupervised semantic segmentation approaches is the limited spatial resolution.
This limitation comes from the SSL training objectives \citep{Caron:2021:EPS,Oquab:2023:DLR}, which are image-level rather than pixel-level.
As a result, we can observe difficulties in segmenting very small, finely resolved structures.}

\vspace{\baselineskip}
\section{Conclusion}
\label{sec:conclusion}

We \cngtwo{present} \ourmethod, a novel dense pseudo-label generation approach for unsupervised semantic segmentation. We derive light\-weight mask proposals directly from off-the-shelf self-supervised learned features, leveraging the intrinsic properties of their embedding space. 
Our mask proposals can be used as pseudo labels to effectively fit global class prototypes using moving average stochastic \ac{em} with \ourmethodframework.
Despite the simplicity, \ourmethodframework leads to a consistent boost in unsupervised segmentation accuracy when applied to a variety of SSL features or orthogonally to current state-of-the-art unsupervised semantic segmentation pipelines, as shown by our results across multiple datasets.

\section*{Acknowledgments}
This project is partially funded by the European Research Council (ERC) under the European Union’s Horizon 2020 research and innovation programme (grant agreement No.\ 866008) as well as the State of Hesse (Germany) through the cluster projects  ``The Third Wave of Artificial Intelligence (3AI)'' and ``The Adaptive Mind (TAM)''.

\bibliography{visinf_bibtex/short, visinf_bibtex/new}
\bibliographystyle{tmlr}

\clearpage
\appendix
\section{Further Analysis \label{sec:furtheranalysis}}
In this appendix, we provide more detailed insights into \ourmethodframework beyond the scope of the main paper.  

\subsection{Nearest-Neighbor Anchoring}
\label{sec:details-nn}
As described in \cref{sec:method} of the main paper, \ourmethod are anchoring the first principal component in each iteration of the mask proposal generation. Corresponding to the quantitative findings (\cf \cref{tab:ablation-a}), here we additionally analyze this qualitatively. \cref{fig:nn_ablation} shows similarity maps of all features to the iteratively computed first principal component (Similarity 1st PC) as well as to the respective nearest-neighbor feature (Similarity 1st PC NN) for example images from Cityscapes \citep{Cordts:2016:CDS}, COCO-Stuff \citep{Caesar:2018:CST}, and Potsdam-3 \citep{PotsdamDataset}. We observe that the originally computed principal direction can have high similarities with multiple semantic concepts in an image. 
Hence, finding a suitable threshold that isolates a single main concept is difficult. 

However, using the nearest image feature to the principal component as an anchoring element helps to circumvent high similarity values to multiple visual concepts. For instance, in the first example for Cityscapes, the similarity map for the first principal component has high similarities to both ``building'' and ``vegetation''. In contrast, the similarity map for the nearest neighbor of the first principal component results in high similarity to the class ``vegetation'' only. Consequently, a suitable mask proposal can be obtained through thresholding. We observe this particularly for the Cityscapes dataset and some examples of COCO-Stuff. Further, in cases where the similarity map is already localized to one visual concept, anchoring merely leads to a change in the similarity values without changing the shape of the thresholded proposal.

\def\imgwidth{0.118\linewidth}
\begin{figure}[!t]
    \small
    \smallskip
    \setlength\tabcolsep{1.0pt}
    \renewcommand{\arraystretch}{0.66}
    \centering  
    \vspace{-2.0em}
    \begin{tabularx}{\linewidth}{lcccccccc}

    \rotatebox[origin=lB]{90}{\hspace{-35.0em}Potsdam-3\hspace{7.5em}COCO-Stuff\hspace{6.7em}Cityscapes} & & & & & & & & \\
    
    & Image  & \multicolumn{3}{c}{Similarity 1st PC} & \multicolumn{3}{c}{Similarity 1st PC NN} & Ground\,truth \\
    \cmidrule(l{0.5em}r{0.5em}){3-5} \cmidrule(l{0.5em}r{0.5em}){6-8}
    
    & \includegraphics[width=\imgwidth]
    {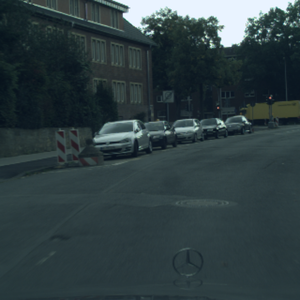}  
    & \includegraphics[width=\imgwidth]
    {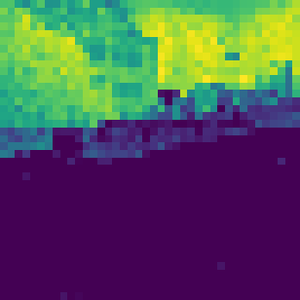} 
    & \includegraphics[width=\imgwidth]
    {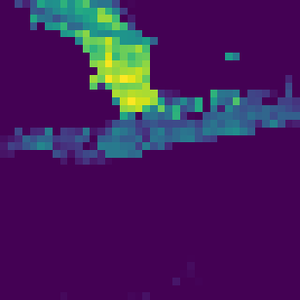} 
    & \includegraphics[width=\imgwidth]
    {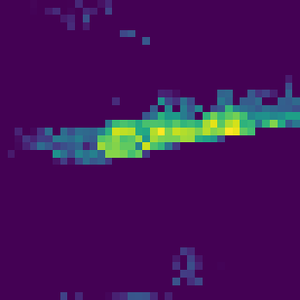} 
    & \includegraphics[width=\imgwidth]
    {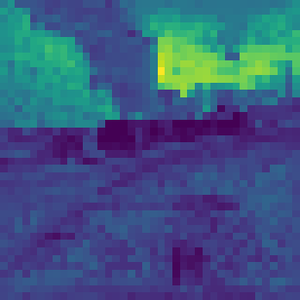} 
    & \includegraphics[width=\imgwidth]
    {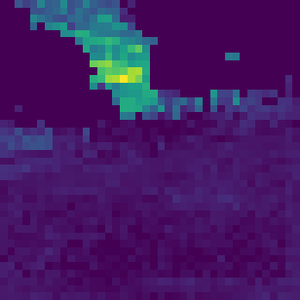} 
    & \includegraphics[width=\imgwidth]
    {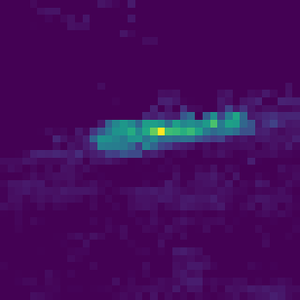} 
    & \includegraphics[width=\imgwidth]
    {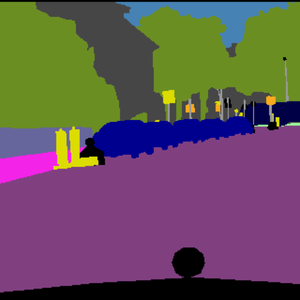} \\

    & \includegraphics[width=\imgwidth]
    {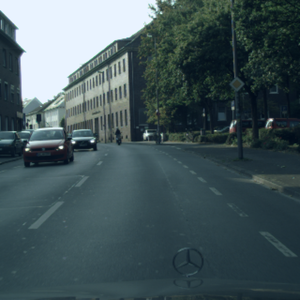}  
    & \includegraphics[width=\imgwidth]
    {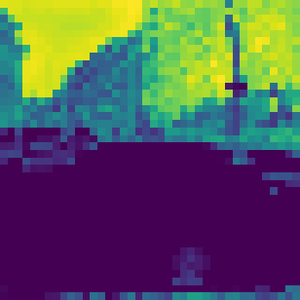} 
    & \includegraphics[width=\imgwidth]
    {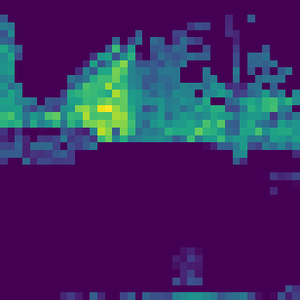} 
    & \includegraphics[width=\imgwidth]
    {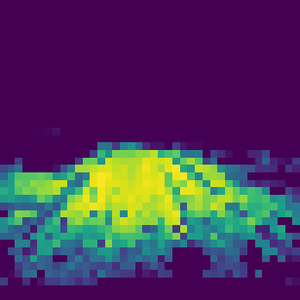}  
    & \includegraphics[width=\imgwidth]
    {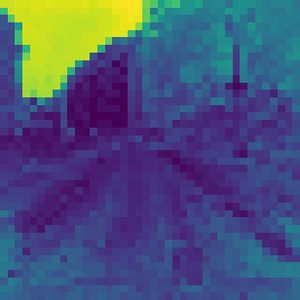} 
    & \includegraphics[width=\imgwidth]
    {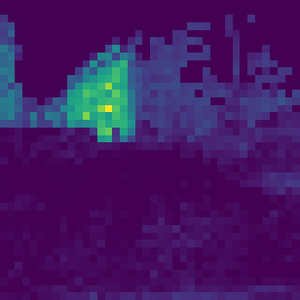} 
    & \includegraphics[width=\imgwidth]
    {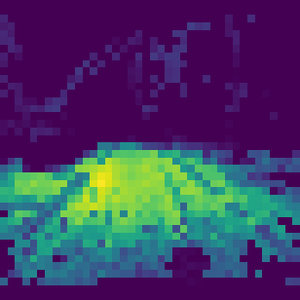}
    & \includegraphics[width=\imgwidth]
    {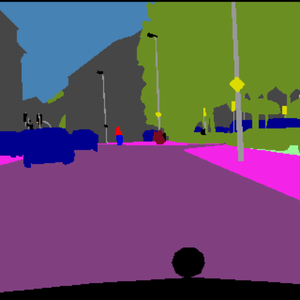} \\

    & \includegraphics[width=\imgwidth]
    {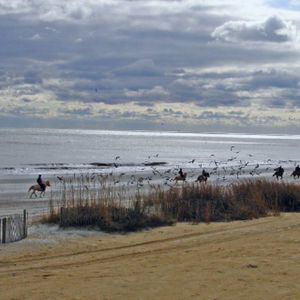}  
    & \includegraphics[width=\imgwidth]
    {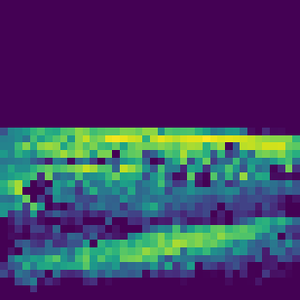}  
    & \includegraphics[width=\imgwidth]
    {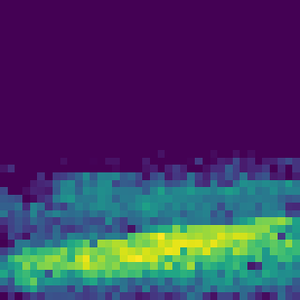}   
    & \includegraphics[width=\imgwidth]
    {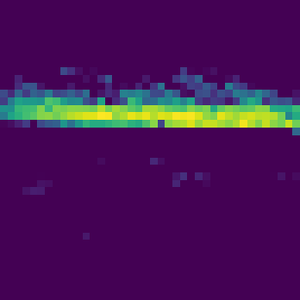}   
    & \includegraphics[width=\imgwidth]
    {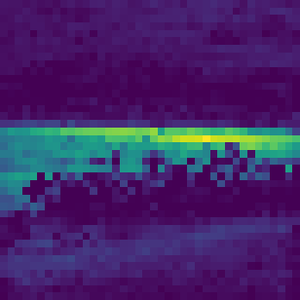}  
    & \includegraphics[width=\imgwidth]
    {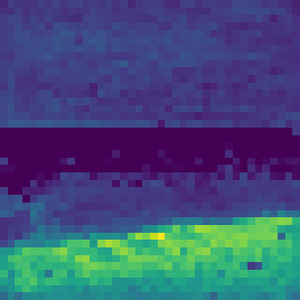}  
    & \includegraphics[width=\imgwidth]
    {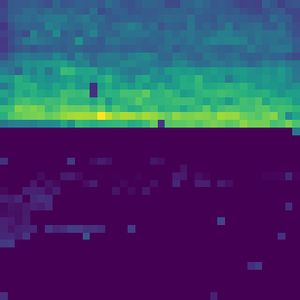} 
    & \includegraphics[width=\imgwidth]
    {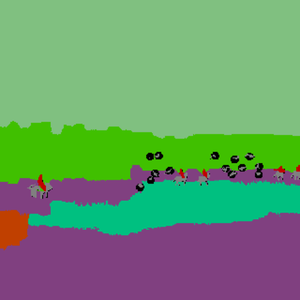} \\

    & \includegraphics[width=\imgwidth]
    {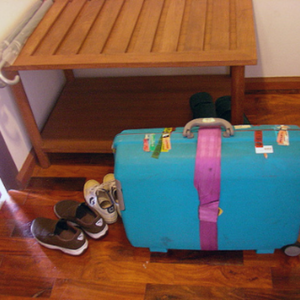}  
    & \includegraphics[width=\imgwidth]
    {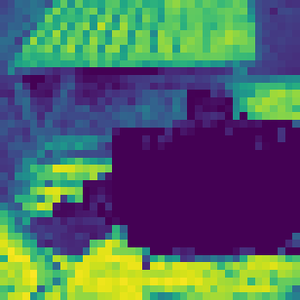}  
    & \includegraphics[width=\imgwidth]
    {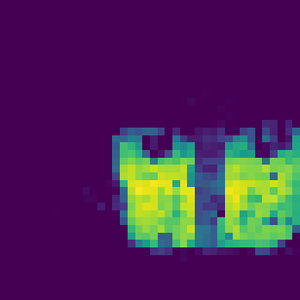}   
    & \includegraphics[width=\imgwidth]
    {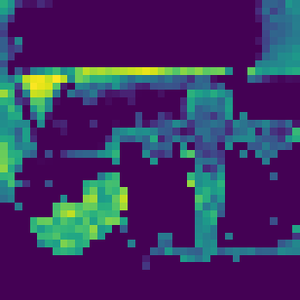}    
    & \includegraphics[width=\imgwidth]
    {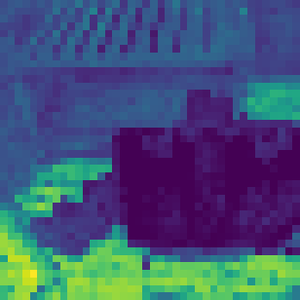}  
    & \includegraphics[width=\imgwidth]
    {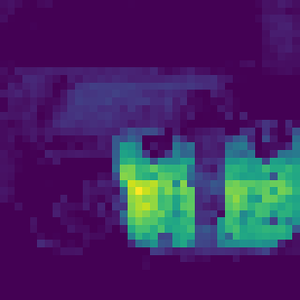}  
    & \includegraphics[width=\imgwidth]
    {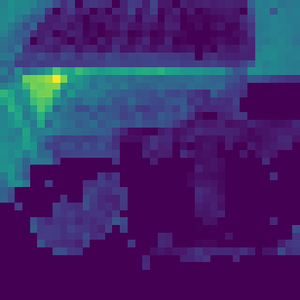}  
    & \includegraphics[width=\imgwidth]
    {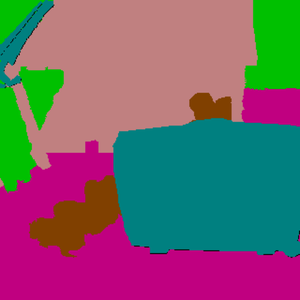} \\

    & \includegraphics[width=\imgwidth]
    {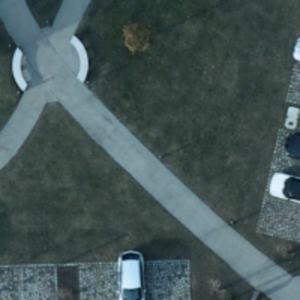}
    & \includegraphics[width=\imgwidth]
    {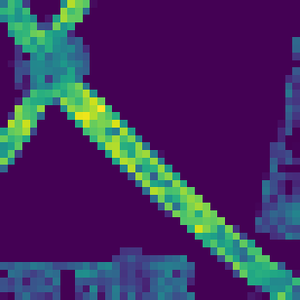}
    & \includegraphics[width=\imgwidth]
    {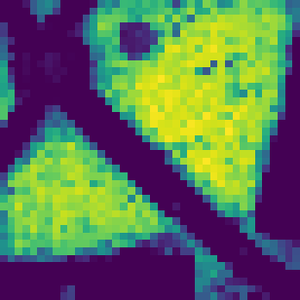}
    & \includegraphics[width=\imgwidth]
    {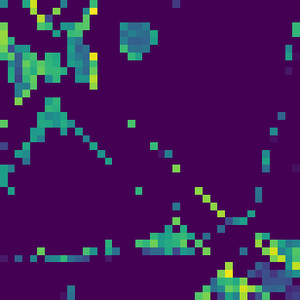}
    & \includegraphics[width=\imgwidth]
    {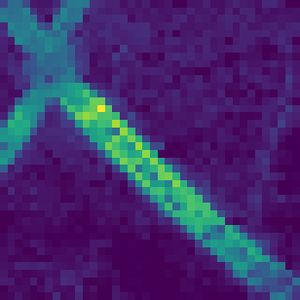}
    & \includegraphics[width=\imgwidth]
    {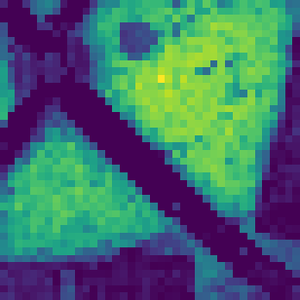}
    & \includegraphics[width=\imgwidth]
    {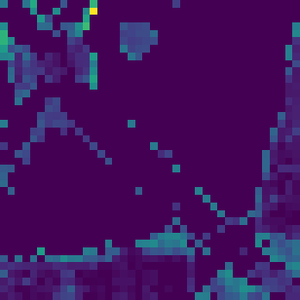}
    & \includegraphics[width=\imgwidth]
    {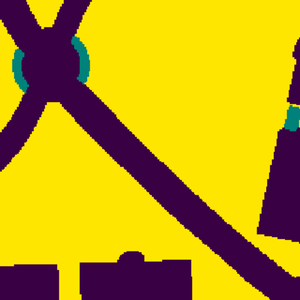} \\

    & \includegraphics[width=\imgwidth]
    {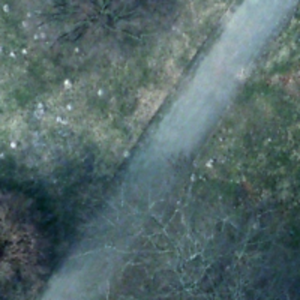}
    & \includegraphics[width=\imgwidth]
    {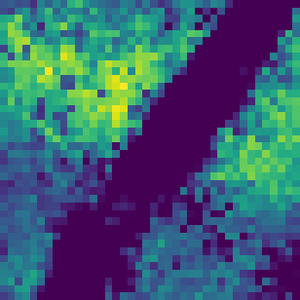}
    & \includegraphics[width=\imgwidth]
    {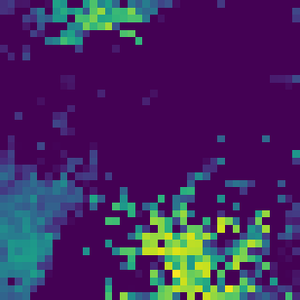}
    & \includegraphics[width=\imgwidth]
    {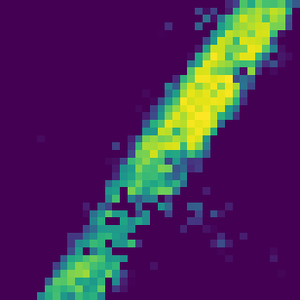}
    & \includegraphics[width=\imgwidth]
    {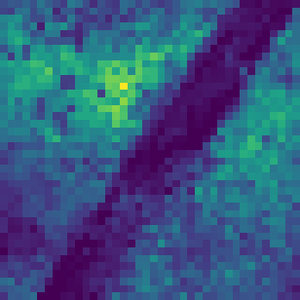}
    & \includegraphics[width=\imgwidth]
    {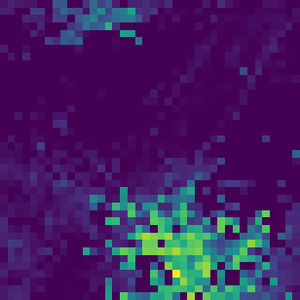}
    & \includegraphics[width=\imgwidth]
    {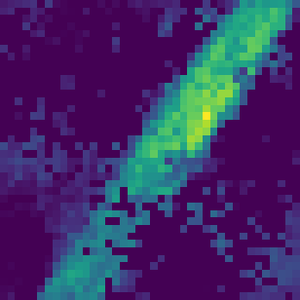}
    & \includegraphics[width=\imgwidth]
    {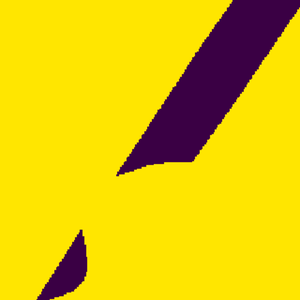} \\

    \end{tabularx}
    \caption{\textbf{Nearest neighbor anchoring of the principal direction in \ourmethod}. Image, ground-truth label, and the first three similarity maps with respect to the principal direction \emph{(left)} and their nearest neighbor \emph{(right)} for all three datasets using DINO ViT-B/8. Anchoring localizes the signal for principal directions with high similarities to multiple visual concepts. \label{fig:nn_ablation}}
\end{figure}

\def\imgwidth{0.10345\linewidth}
\begin{figure}[t]
    \small
    \smallskip
    \setlength\tabcolsep{1.0pt}
    \renewcommand{\arraystretch}{0.66}
    \centering
    \begin{tabularx}{\textwidth}{lccccccccc}
    & \multicolumn{3}{c}{Cityscapes} & \multicolumn{3}{c}{COCO-Stuff} & \multicolumn{3}{c}{Potsdam-3} \\
    \cmidrule(l{0.5em}r{0.5em}){2-4} \cmidrule(l{0.5em}r{0.5em}){5-7} \cmidrule(l{0.5em}r{0.5em}){8-10}
    \rotatebox[origin=lB]{90}{\scriptsize{\hspace{-17.8em}\shortstack{\\$\psi=0.3$} \hspace{2.0em} \shortstack{\ourmethod \\$\psi=0.4$} \hspace{2.0em} \shortstack{\\$\psi=0.5$} \hspace{3.0em} Image}}
    
    & \includegraphics[width=\imgwidth]
    {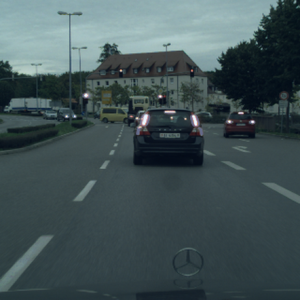} 
    & \includegraphics[width=\imgwidth]
    {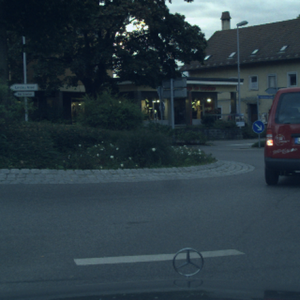} 
    & \includegraphics[width=\imgwidth]
    {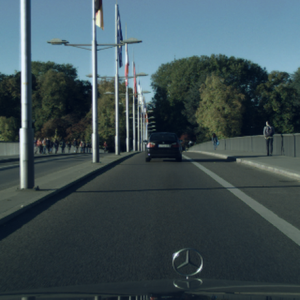}  
    & \includegraphics[width=\imgwidth]
    {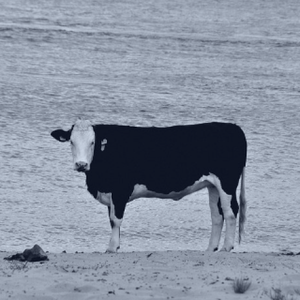} 
    & \includegraphics[width=\imgwidth]
    {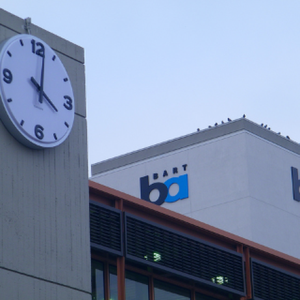} 
    & \includegraphics[width=\imgwidth]
    {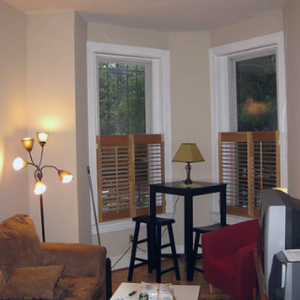}  
    & \includegraphics[width=\imgwidth]
    {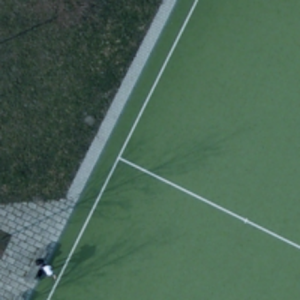}  
    & \includegraphics[width=\imgwidth]
    {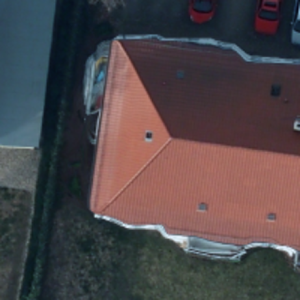} 
    & \includegraphics[width=\imgwidth]
    {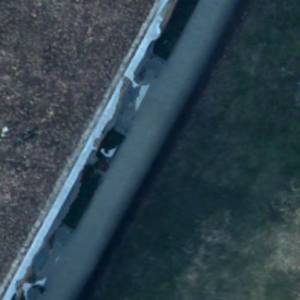} \\

    & \includegraphics[width=\imgwidth]
    {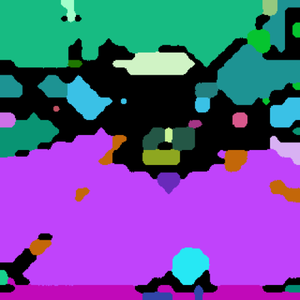}   
    & \includegraphics[width=\imgwidth]
    {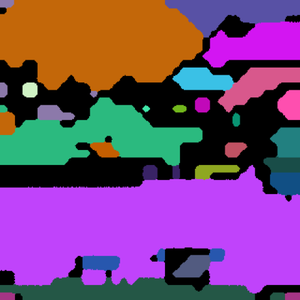}   
    & \includegraphics[width=\imgwidth]
    {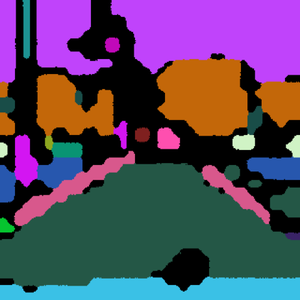}   
    & \includegraphics[width=\imgwidth]
    {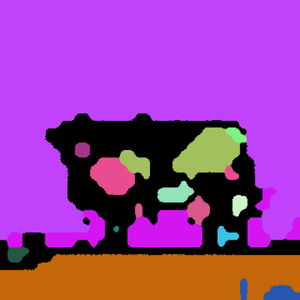}   
    & \includegraphics[width=\imgwidth]
    {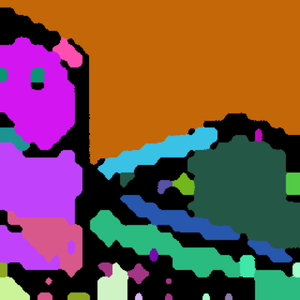}   
    & \includegraphics[width=\imgwidth]
    {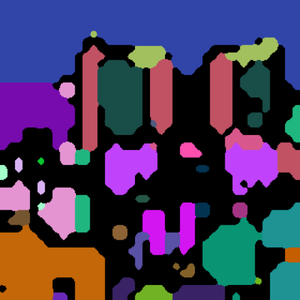}   
    & \includegraphics[width=\imgwidth]
    {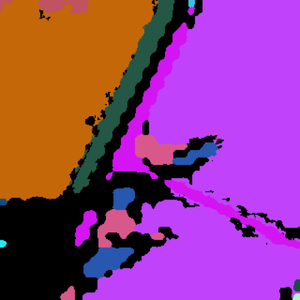}  
    & \includegraphics[width=\imgwidth]
    {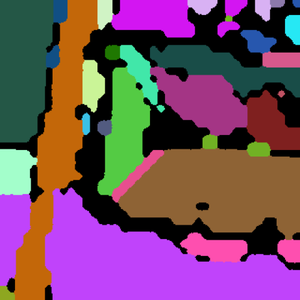} 
    & \includegraphics[width=\imgwidth]
    {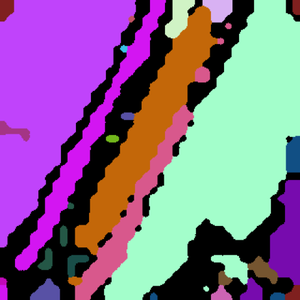}   \\

    & \includegraphics[width=\imgwidth]
    {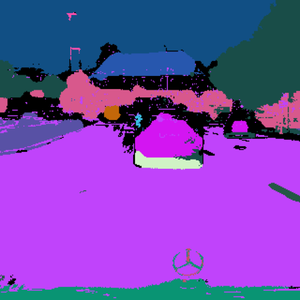} 
    & \includegraphics[width=\imgwidth]
    {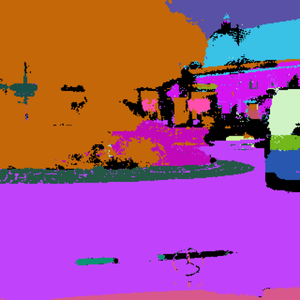}   
    & \includegraphics[width=\imgwidth]
    {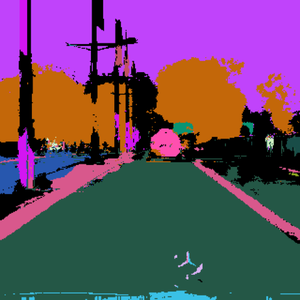}
    & \includegraphics[width=\imgwidth]
    {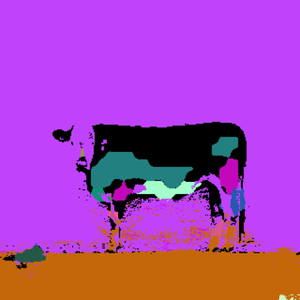} 
    & \includegraphics[width=\imgwidth]
    {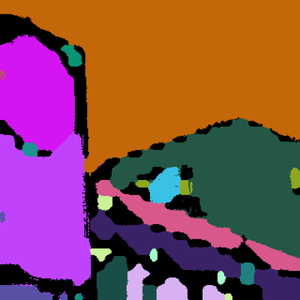} 
    & \includegraphics[width=\imgwidth]
    {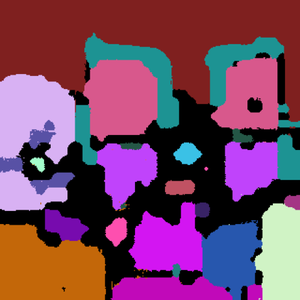} 
    & \includegraphics[width=\imgwidth]
    {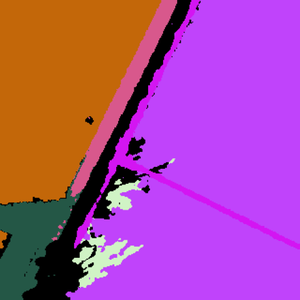} 
    & \includegraphics[width=\imgwidth]
    {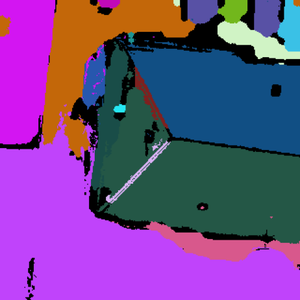} 
    & \includegraphics[width=\imgwidth]
    {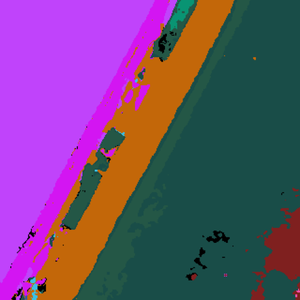} \\

    & \includegraphics[width=\imgwidth]
    {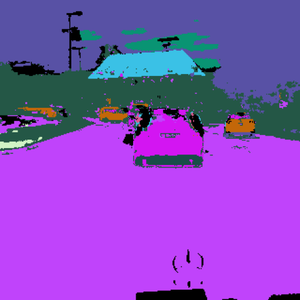} 
    & \includegraphics[width=\imgwidth]
    {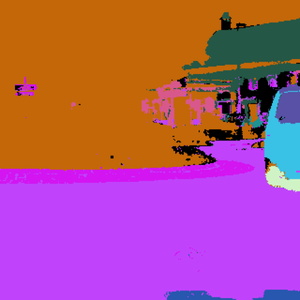}
    & \includegraphics[width=\imgwidth]
    {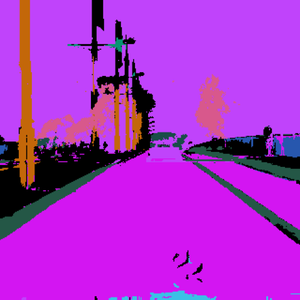} 
    & \includegraphics[width=\imgwidth]
    {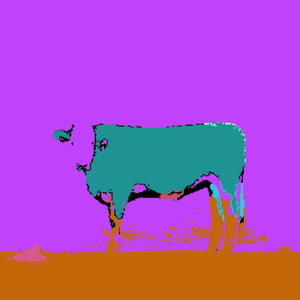} 
    & \includegraphics[width=\imgwidth]
    {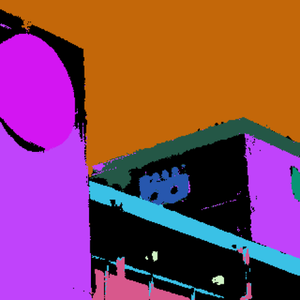} 
    & \includegraphics[width=\imgwidth]
    {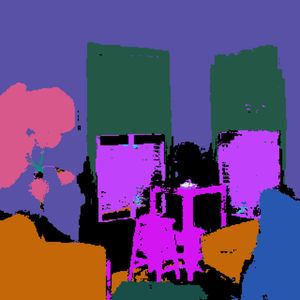} 
    & \includegraphics[width=\imgwidth]
    {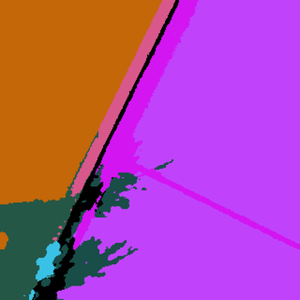} 
    & \includegraphics[width=\imgwidth]
    {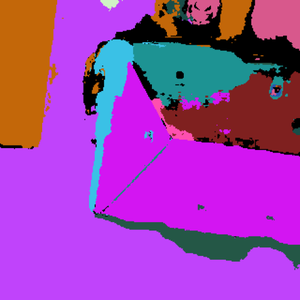} 
    & \includegraphics[width=\imgwidth]
    {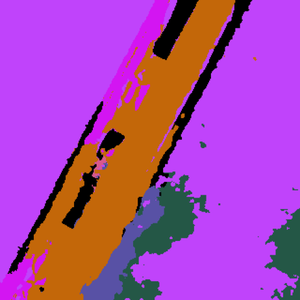} \\
       
    \end{tabularx}
    \caption{\textbf{Qualitative threshold $\psi$ ablation} with DINO ViT-B/8 using different $\psi$ values for \ourmethod mask-proposal generation on all three datasets. The hyperparameter exhibits favorable stability properties. \label{fig:thresh_ablation}}
    \vspace{-0.5em}
\end{figure}

\subsection{Ablation of the Threshold \label{sec:thresh_ablation}}
In the spirit of unsupervised learning, our method effectively has only a single additional hyperparameter -- the threshold $\psi$. Furthermore, this parameter can be set simply by examining the mask proposal as detailed next. In addition, it should be noted that we keep this parameter unchanged for all backbone models and datasets, which further emphasizes the generalizability of our method. As described in \cref{sec:method} of the main paper, the threshold $\psi$ is used to remove noise in the similarity masks and to localize the optimization signal. In \cref{fig:thresh_ablation}, qualitative examples of \ourmethod mask proposals are visualized for DINO ViT-B/8 on Cityscapes, COCO-Stuff, and Potsdam-3. We show mask proposals for $\psi=0.3$, $\psi=0.4$, and $\psi=0.5$. Visually, it can be observed that meaningful masks are produced for all thresholds. Especially those for threshold $0.3$ and $0.4$ align very well with the semantic content in the images. While $\psi=0.3$ seems to provide better mask proposals for COCO-Stuff and Potsdam-3, a problem arises with Cityscapes. Here, the mask proposals contain several semantic concepts and spatially small objects in a  large mask (\cf the second Cityscapes example, where the mask of the bush in the left half of the image covers both the traffic sign in the foreground, parts of the house in the background, as well as the sky). Since this would lead to a poor optimization signal and the masks of the other two datasets using $\psi=0.4$ seem visually appealing, the threshold is set to $0.4$ in all other experiments.

To further shed light on these qualitative observations, we apply \ourmethodframework for the scenario described above and vary the threshold from $\psi=0.2$ to $\psi=0.6$ with a step size of 0.1. We perform this with the DINO ViT-B/8 backbone for Cityscapes, COCO-Stuff, and Potsdam-3 and show the mIoU in \cref{fig:thresh_ablation_quantitative}. The quantitative results reflect our qualitative conclusion of the threshold well, and show the trade-off between better segmentation accuracy on COCO-Stuff and Potsdam-3 for a lower threshold and vice versa for Cityscapes. In numbers, a threshold of $\psi=0.5$ instead of $\psi=0.4$ for Cityscapes would lead to an additional gain of 1.4\:\% mIoU, but also to slight losses on COCO-Stuff of 0.8\:\%. Additionally, this results in more mask proposal iterations. We conclude that the determined threshold $\psi=0.4$ appears to be reasonable. Even if better results could be achieved for some backbone-dataset combinations with individually set thresholds, we consider a fixed threshold that generalizes well across all scenarios to be sensible. We would like to emphasize that this experiment was conducted solely for this single backbone and serves only to validate the qualitative judgement from above. Importantly, we did not determine the hyperparameters based on the evaluation sets.

\begin{figure}[ht]
    \begin{minipage}[b]{0.49\textwidth}
    \centering
    \begin{tikzpicture}[
scale=\textwidth/8.25cm, %
every node/.style={font=\small}]
\definecolor{city}{RGB}{100,143,255}
\definecolor{coco}{RGB}{254, 97, 0}
\definecolor{pots}{RGB}{220, 38, 127}
\begin{axis}[
    xlabel={\small{Threshold $\psi$}},
    ylabel={\small{mIoU (\%)}},
    legend style={at={(0.05,0.6)},anchor=west, scale=0.85, , row sep=0.5pt},
    grid=both,
    xmin=0.18,
    xmax=0.62,
    ymin=5,
    ymax=70,
    width=8.5cm,
    height=5cm,
    xtick={0.2, 0.3, 0.4, 0.5, 0.6},
    xticklabels={0.2, 0.3, 0.4, 0.5, 0.6},
    xlabel near ticks,
    ylabel near ticks,
    ylabel shift = -0.2cm,
    xlabel shift = -0.1cm,
]

\addplot[mark=*, color=city, very thick] coordinates {
    (0.2, 6.4)
    (0.3, 14.8)
    (0.4, 17.6)
    (0.5, 19.0)
    (0.6, 18.1)
};
\addlegendentry{\small{Cityscapes}}

\addplot[mark=*, color=coco, very thick] coordinates {
    (0.2, 22.3)
    (0.3, 22.0)
    (0.4, 21.9)
    (0.5, 21.1)
    (0.6, 20.5)
};
\addlegendentry{\small{COCO-Stuff}}

\addplot[mark=*, color=pots, very thick] coordinates {
    (0.2, 66.8)
    (0.3, 67.3)
    (0.4, 66.9)
    (0.5, 66.9)
    (0.6, 63.5)
};

\addlegendentry{\small{Potsdam-3}}

\end{axis}
\end{tikzpicture}
    \vspace{-0.5em}
    \caption{\textbf{Quantitative threshold $\psi$ ablation} with DINO ViT-B/8 using different $\psi$ values for \ourmethod mask-proposal generation on all three datasets. The single hyperparameter of our method exhibits favorable stability properties. \label{fig:thresh_ablation_quantitative}}
    \vspace{-0.5em}
\end{minipage}
\hfill
\begin{minipage}[b]{0.49\textwidth}
    \centering
    \begin{tikzpicture}[
scale=\textwidth/8.25cm, %
every node/.style={font=\small}]
\definecolor{city}{RGB}{100,143,255}
\definecolor{coco}{RGB}{254, 97, 0}
\definecolor{pots}{RGB}{220, 38, 127}
\begin{axis}[
    xlabel={\small{Stop Criterion (\%)}},
    ylabel={\small{mIoU (\%)}},
    legend style={at={(0.05,0.75)},anchor=west, scale=0.85, row sep=0.5pt},
    grid=both,
    xmin=0.84,
    xmax=1.0,
    ymin=10,
    ymax=70,
    width=8.5cm,
    height=5cm,
    xtick={0.85, 0.90, 0.95, 0.99},
    xticklabels={85, 90, 95, 99},
    xlabel near ticks,
    ylabel near ticks,
    ylabel shift = -0.2cm,
    xlabel shift = -0.1cm,
]

\addplot[mark=*, color=city, very thick] coordinates {
    (0.85, 16.1)
    (0.90, 16.8)
    (0.95, 17.6)
    (0.99, 18.5)
};
\addlegendentry{\small{Cityscapes}}

\addplot[mark=*, color=coco, very thick] coordinates {
    (0.85, 21.9)
    (0.90, 22.1)
    (0.95, 21.9)
    (0.99, 21.7)
};
\addlegendentry{\small{COCO-Stuff}}

\addplot[mark=*, color=pots, very thick] coordinates {
    (0.85, 27.8)
    (0.90, 30.0)
    (0.95, 67.0)
    (0.99, 59.5)
};

\addlegendentry{\small{Potsdam-3}}

\end{axis}
\end{tikzpicture}
    \vspace{-0.5em}
    \caption{\rebut{\textbf{Quantitative stopping-criterion ablation} with DINO ViT-B/8 using different stopping criteria for \ourmethod mask proposal generation on all three datasets.\vspace{\baselineskip}\label{fig:stopping_ablation_quantitative}}}
    \vspace{-0.5em}
\end{minipage}
\end{figure}


\subsection{\rebut{Ablation of the Stopping Criterion}}
\rebut{Our method iteratively decomposes images into mask proposals. Hence, a stopping criterion is essential to decide whether a sufficient number of masks has been generated for a particular image. This allows us to divide different images into a different number of masks. We use the percentage of features assigned to masks. Specifically, the iterative \ourmethod process stops when 95\% of image features are assigned to masks. While setting this value as high as possible to cover the entire image with masks might seem intuitive, not all features can be assigned reasonably due to the inherent noise in the self-supervised feature representation. Setting the value too high would result in masks with very few features in the final iterations. We chose a stopping criterion of 95\%, which we confirmed quantitatively to yield good results across all datasets as shown in \cref{fig:stopping_ablation_quantitative}. The relatively low mIoU for the Potsdam-3 dataset with smaller stopping criteria results from a few initial mask proposals covering large areas, while the proposals in the final iterations represent finer details.}

\subsection{\rebut{Ablation of the Image Augmentations} \label{sec:ablation_extendedaugs}}
\rebut{In addition to the \cref{tab:ablation-b}, we analyze the photometric augmentations used in \ourmethodframework in greater detail. As can be seen, the additional augmentation leads to slightly improved performance in terms of metrics. We used a combination of standard augmentations as well as the standard torchvision implementation and did not tune any parameters. \rebuttwo{\cref{tab:ablation_extendedaugs} provides} a contribution breakdown of the used augmentation techniques (grayscaling, Gaussian blur, color jitter) on COCO-Stuff using DINO ViT-B/8. Each of the three photometric augmentations contributes to the method's performance in terms of downstream metrics.}

\begin{table}[!t]
\vspace{0.5em}
\caption{\textbf{Ablation study} analyzing design choices and components of \ourmethodframework for COCO-Stuff using DINO ViT-B/8 including a breakdown of the image augmentations.} \label{tab:ablation_extendedaugs}
\small
\renewcommand{\arraystretch}{1.0}
\centering
        \begin{tabularx}{\linewidth}{@{}Xcc@{}}
            \toprule
            \textbf{Method} & \textbf{Acc} & \textbf{mIoU} \\
            \midrule 
            Baseline \citep{Caron:2021:EPS}                      & 38.8  & 15.7 \\ 
           +\,\ourmethod pseudo label                            & 38.8  & 18.0 \\  
           +\,EMA                                                & 45.0  & 20.2 \\  
           +\,Augment (grayscaling) 	                          & 45.3  & 20.4 \\
           +\,Augment (grayscaling, Gaussian blur)               & 45.8  & 20.2 \\
           +\,Augment (grayscaling, Gaussian blur, color jitter) & 46.0  & 20.4 \\
           +\,CRF ($\equiv$ \ourmethodframework)                 & 48.4  & 21.9 \\
            \bottomrule
        \end{tabularx}
\end{table}

\subsection{\rebut{Ablation on the Number of Pseudo Classes $K$}}

\begin{table}[!t]
\caption{\rebut{\textbf{Over clustering} with \ourmethodframework for COCO-Stuff using DINO ViT-B/8.}} \label{tab:overclustering}
\small
\renewcommand{\arraystretch}{1.0}
\centering
        \begin{tabularx}{\linewidth}{@{}Xcc@{}}
            \toprule
            \textbf{Method} & \textbf{Acc} & \textbf{mIoU} \\
            \midrule 
            \ourmethodframework (K=27; 100\%) 	& 48.5 & 21.9 \\
            \ourmethodframework (K=40; ~150\%) 	& 53.1 & 23.2 \\
            \bottomrule
        \end{tabularx}
\end{table}     

\rebuttwo{Following previous works \citep{Ji:20219:IIC,Cho:2021:PUS,Hamilton:2022:USS,Seong:2023:LHP} in unsupervised semantic segmentation, generally the number of pseudo classes $K$ is set to the number of annotated semantic classes in the dataset.} \rebut{This is done to evaluate the performance in terms of downstream metrics on every semantic class after matching the predicted pseudo-class IDs with the ground-truth classes using Hungarian matching. Using a $K$ smaller than the number of ground-truth classes is hard to evaluate and compare as it results in ground-truth classes that need to be ignored by the metric. \rebuttwo{However, it is possible to have more pseudo classes than ground-truth classes by assigning multiple pseudo classes to a single ground-truth class.} We conduct an experimented in this over-segmentation setup, \ie predicting $K$ categories, where $K$ is larger than the number of ground-truth classes. We realize the multi-to-one matching by applying Hungarian matching first and subsequently matching the remaining pseudo-class IDs based on their highest correspondence to the ground-truth classes. In \cref{tab:overclustering}, we report the unsupervised semantic segmentation results for COCO-Stuff with DINO ViT-B/8 once using $K=27$, which matches the number of ground-truth classes, and for $K=40$. \rebuttwo{Over clustering} leads to better performance in terms of metrics, as multiple pseudo classes represent a single ground-truth class. Overall, estimating the number of pseudo classes or semantic concepts in a dataset in an unsupervised manner represents an intriguing direction for future research.}


\section{Further Experiments}
This appendix provides insights beyond the experiments and ablations shown in \cref{sec:experiments}.

\begin{figure}
\caption{\textbf{COCO-Stuff -- Comparison of the segmentation performance for individual classes} in terms of IoU (in \%) \emph{(a)} and class confusion for the DINO ViT-B/8 Baseline \emph{(b)}, 
\ourmethodframework \emph{(c)}, STEGO \emph{(d)}, and STEGO + \ourmethodframework \emph{(e)}. Overall, \ourmethodframework preserves or boosts the individual class IoU across most classes and moderately reduces class confusion.\label{fig:coco_confusion}}
\begin{subfigure}{\linewidth}
    \caption{Class IoUs (in \%) for DINO ViT-B/8 Baseline, \ourmethodframework\textit{(Ours)}, STEGO, and STEGO+\ourmethodframework\textit{(Ours)}. \label{tab:cococlsiou}}
    \vspace{-1em}
    \tiny
\centering
\renewcommand{\arraystretch}{1.0}
\setlength\tabcolsep{-0.1pt}
\centering

\begin{tabularx}{\linewidth}{@{}XSSSSSSSSSSSSSSSSSSSSSSSSSSS@{}}
\toprule

&\rotatebox{90}{Electronic} 
&\rotatebox{90}{Appliance} 
&\rotatebox{90}{Food} 
&\rotatebox{90}{Furniture} 
&\rotatebox{90}{Indoor} 
&\rotatebox{90}{Kitchen} 
&\rotatebox{90}{Accessory} 
&\rotatebox{90}{Animal} 
&\rotatebox{90}{Outdoor} 
&\rotatebox{90}{Person} 
&\rotatebox{90}{Sports} 
&\rotatebox{90}{Vehicle} 
&\rotatebox{90}{Ceiling} 
&\rotatebox{90}{Floor} 
&\rotatebox{90}{Food} 
&\rotatebox{90}{Furniture} 
&\rotatebox{90}{Rawmaterial} 
&\rotatebox{90}{Textile} 
&\rotatebox{90}{Wall} 
&\rotatebox{90}{Window} 
&\rotatebox{90}{Building} 
&\rotatebox{90}{Ground} 
&\rotatebox{90}{Plant} 
&\rotatebox{90}{Sky} 
&\rotatebox{90}{Solid} 
&\rotatebox{90}{Structural} 
&\rotatebox{90}{Water} \\

\midrule
Baseline &\phantom{0}0.0 &\phantom{0}0.0 &\phantom{0}0.0 &\phantom{0}0.2 &\phantom{0}0.0 &\phantom{0}0.0 &\phantom{0}0.0 &15.3 &\phantom{0}0.1 &58.3 &\phantom{0}0.5 &\phantom{0}9.1 &23.7 &\phantom{0}3.1 &\phantom{0}2.2 &\phantom{0}8.3 &12.2 &\phantom{0}0.2 &30.7 &11.7 &31.9 &21.5 &33.8 &77.4 &15.0 &27.1 &45.1 \\

\textit{Ours} &0.0 &0.2 &0.0 &0.8 &0.0 &0.0 &0.0 &21.4 &4.6 &68.7 &0.5 &12.6 &36.1 &44.4 &0.0 &12.1 &14.0 &1.1 &45.7 &15.9 &38.1 &33.3 &42.9 &69.1 &29.1 &38.9 &61.4 \\
        \rowcolor{mylightgray}
STEGO &0.0 &0.3 &0.3 &13.7 &0.1 &0.0 &0.7 &74.1 &0.1 &61.7 &10.9 &40.7 &36.3 &30.2 &38.8 &22.5 &15.6 &11.0 &36.0 &2.7 &51.8 &44.1 &50.2 &82.9 &19.8 &37.8 &58.8 \\
        \rowcolor{mylightgray}
STEGO+\textit{Ours} &0.2 &0.0 &0.0 &14.1 &0.2 &0.0 &0.3 &76.8 &10.9 &64.3 &0.9 &53.8 &48.2 &31.9 &39.4 &25.4 &16.0 &19.7 &36.1 &0.8 &55.6 &44.9 &51.2 &83.8 &27.4 &36.2 &62.6 \\

\bottomrule
\end{tabularx}

\end{subfigure}

\vspace{1.5em}
\begin{subfigure}{0.47\linewidth}
    \caption{Confusion matrix for the Baseline\label{fig:coco_base_conf}}
    \vspace{-1.5em}
    \def\moduloop#1#2{\ifnum\numexpr(#1-(#1/#2)*(#2))\relax<0(#1-(#1/#2)*(#2)+#2)\else(#1-(#1/#2)*(#2))\fi}
\def\truncdiv#1#2{((#1-\moduloop{#1}{#2})/(#2))}

\pgfplotstableread[col sep=comma]{
0,0,0,0,0,0,0,40,0,0,0,12,0,0,5,8,14,0,1,13,1,0,1,0,0,0,0
0,0,0,0,0,0,0,48,0,0,0,12,0,0,4,16,2,0,2,10,0,0,0,0,0,0,0
0,0,0,0,0,0,0,55,0,0,0,11,0,0,18,6,1,0,0,0,0,0,3,0,0,0,0
0,0,0,0,0,0,0,44,0,0,0,13,0,0,9,22,0,1,0,1,0,0,2,0,0,0,0
0,0,0,0,0,0,0,40,0,1,0,13,0,0,8,4,25,1,0,1,0,0,0,0,0,0,0
0,0,0,0,0,0,0,23,0,0,0,48,0,0,7,11,4,0,0,1,0,1,0,0,0,0,0
0,0,0,0,0,0,0,31,0,7,0,18,0,0,8,5,21,2,0,1,0,0,0,0,0,0,0
0,0,0,0,0,0,0,63,0,1,0,10,0,0,9,2,0,8,0,0,0,0,0,0,0,0,0
0,0,0,0,0,0,0,44,0,0,0,13,0,0,7,7,17,1,0,2,0,0,0,0,0,0,0
0,0,0,0,0,0,0,4,0,45,0,9,0,0,7,2,1,27,0,0,0,0,0,0,0,0,0
0,0,0,1,0,0,0,38,0,2,2,21,0,0,5,2,18,2,0,0,0,0,0,0,0,1,0
0,0,0,0,0,0,0,48,0,0,0,27,0,1,4,4,4,1,0,2,1,0,0,0,0,1,0
0,0,0,0,0,0,0,4,0,0,0,1,0,0,3,20,2,0,57,3,4,0,0,1,0,1,0
0,0,0,1,0,0,0,2,0,0,0,2,0,38,8,23,0,0,0,0,0,18,0,0,0,0,1
0,0,0,0,0,0,0,37,0,0,2,14,0,0,18,16,0,0,0,0,0,2,2,0,0,0,1
0,0,0,0,0,0,0,20,0,0,0,7,0,1,6,24,2,0,10,18,2,1,0,0,0,1,1
0,0,0,0,0,0,0,21,0,0,0,14,0,0,6,13,19,0,2,13,1,0,0,0,0,1,0
0,0,0,0,0,0,0,15,0,2,0,10,0,2,7,20,9,1,4,17,0,2,0,0,0,0,2
0,0,0,0,0,0,0,7,0,0,0,3,0,0,5,12,2,0,41,10,8,1,0,1,1,1,0
0,0,0,0,0,0,0,7,0,0,0,4,0,1,4,8,1,0,4,48,6,0,4,3,2,2,0
0,0,0,0,0,0,0,15,0,0,0,7,0,0,5,8,3,0,3,9,38,0,1,0,1,2,0
0,0,0,2,0,0,0,2,6,0,1,1,0,25,3,13,0,0,0,0,0,23,0,0,1,0,14
0,0,0,0,0,0,0,2,22,0,0,2,0,0,5,8,0,0,0,0,1,3,41,1,6,1,0
0,0,0,0,0,0,0,0,0,0,0,0,0,0,2,3,0,0,0,0,0,0,0,86,1,1,0
0,0,0,0,0,0,0,6,6,0,1,3,0,1,4,12,0,0,0,1,1,8,4,2,40,0,2
0,0,0,0,0,0,0,6,4,0,0,4,0,2,6,7,0,0,1,6,4,1,5,0,2,44,0
0,0,0,0,0,0,0,2,0,0,17,1,0,0,2,11,0,0,0,0,0,1,0,10,1,0,48

}\mymatrix
\pgfplotstablegetrowsof{\mymatrix}%
\pgfmathtruncatemacro{\numrows}{\pgfplotsretval}%
\pgfplotstablegetcolsof{\mymatrix}%
\pgfmathtruncatemacro{\numcols}{\pgfplotsretval}%
\pgfmathtruncatemacro{\numnew}{\numrows*\numcols}%
\pgfmathdeclarefunction{myentry}{2}{%
\begingroup
\pgfkeys{/pgf/fpu,/pgf/fpu/output format=fixed}%
\pgfmathtruncatemacro{\myx}{#1}%
\pgfmathtruncatemacro{\myy}{#2}%
\pgfplotstablegetelem{\myy}{[index]\myx}\of\mymatrix%
\let\pgfmathresult\pgfplotsretval%
\pgfmathsmuggle\pgfmathresult
\endgroup}%

\pgfplotstablenew[
    create on use/y/.style={create col/expr={\the\numexpr\moduloop{\pgfplotstablerow}{\numrows}}},
    create on use/x/.style={create col/expr={\the\numexpr\truncdiv{\pgfplotstablerow}{\numcols}}},
    create on use/C/.style={create col/expr={myentry(\the\numexpr\truncdiv{\pgfplotstablerow}{\numcols},\the\numexpr\moduloop{\pgfplotstablerow}{\numrows})}},
    columns={x,y,C}
  ]
  {\numnew}%
  \newtable

\begin{tikzpicture}

\begin{axis}[axis equal,
             axis line style={draw=none},
             width=9cm,
             height=9cm,
             tick align=outside,
             /pgfplots/colormap={whiteblue}{rgb255(0cm)=(250,250,255);
                                           rgb255(1cm)=(54,125,189)},
             xtick=data,
             ticks=none,
             xtick pos=bottom,
             xticklabels={
             \footnotesize{class1},
             \footnotesize{class1},
             \footnotesize{class1}
             },
             ytick pos=left,
             ymin=-0.5,ymax=26.5,
             ytick=data,
             yticklabels={
             \footnotesize{class1},
             \footnotesize{class1},
             \footnotesize{class1}
             },
             ]
 \addplot [matrix plot,
           nodes={scale=0.4, color=black!70},
           nodes near coords,
           nodes near coords style={anchor=center},
           mesh/cols=\numcols,
           point meta=explicit] table [meta=C] {\newtable};

\end{axis}

\node (classes) [] at (3.75, -0.75) {\footnotesize{\textbf{Predicted labels}}};
\node (classes) [rotate=90] at (-0.75, 3.7) {\footnotesize{\textbf{Ground-truth labels}}};

\node (classes) [scale=0.563, align=left] at (0.0, 3.7) {
Electronic \\
Appliance \\
Food \\
Furniture \\
Indoor \\
Kitchen \\
Accessory \\
Animal \\
Outdoor \\
Person \\
Sports \\
Vehicle \\
Ceiling \\
Floor \\
Food \\
Furniture \\
Rawmaterial \\
Textile \\
Wall \\
Window \\
Building \\
Ground \\
Plant \\
Sky \\
Solid \\
Structural \\
Water
};

\node (classes) [scale=0.563, align=left, rotate=90] at (3.7, 0.0) {
Electronic \\
Appliance \\
Food \\
Furniture \\
Indoor \\
Kitchen \\
Accessory \\
Animal \\
Outdoor \\
Person \\
Sports \\
Vehicle \\
Ceiling \\
Floor \\
Food \\
Furniture \\
Rawmaterial \\
Textile \\
Wall \\
Window \\
Building \\
Ground \\
Plant \\
Sky \\
Solid \\
Structural \\
Water
};
\end{tikzpicture}
\end{subfigure}
\hspace{0.03\linewidth}
\begin{subfigure}{0.47\linewidth}
    \caption{Confusion matrix for \ourmethodframework\label{fig:coco_ours_conf}}
    \vspace{-1.5em}
    \def\moduloop#1#2{\ifnum\numexpr(#1-(#1/#2)*(#2))\relax<0(#1-(#1/#2)*(#2)+#2)\else(#1-(#1/#2)*(#2))\fi}
\def\truncdiv#1#2{((#1-\moduloop{#1}{#2})/(#2))}

\pgfplotstableread[col sep=comma]{
0,0,0,0,0,0,0,38,4,0,0,16,0,0,0,3,18,0,5,8,0,0,2,0,0,0,0
0,0,0,0,0,0,0,19,6,0,0,37,0,0,0,11,1,1,10,8,0,0,0,0,0,0,0
0,0,0,0,0,0,0,76,0,0,0,10,0,0,0,6,0,0,0,0,0,0,3,0,0,0,0
0,0,0,1,0,0,0,16,2,0,0,48,0,1,0,18,0,1,2,0,0,0,2,0,0,0,0
0,0,0,0,0,0,0,51,1,0,0,11,0,0,0,3,26,0,1,0,0,0,0,0,0,0,0
0,0,0,0,0,0,0,19,2,0,0,53,0,0,0,8,1,9,2,0,0,0,0,0,0,0,0
0,0,0,0,0,0,0,35,0,9,0,37,3,0,0,2,2,2,1,0,0,0,0,0,0,0,0
0,0,1,0,0,0,0,87,0,0,0,1,0,0,2,1,0,0,0,0,0,0,1,0,0,0,0
0,0,1,5,0,0,0,35,16,0,0,10,0,0,1,6,14,0,0,0,1,0,1,0,0,0,0
0,0,0,0,0,0,0,8,0,75,0,2,0,0,0,1,0,5,0,0,0,0,0,0,0,0,0
0,0,0,0,0,0,0,37,2,2,4,28,0,0,0,2,7,5,0,0,0,0,0,1,0,0,1
0,0,0,0,0,0,0,49,4,0,0,27,0,0,0,2,1,4,0,0,2,0,0,0,0,0,0
0,0,0,0,0,0,0,3,0,0,0,0,84,0,0,1,0,0,5,0,1,0,0,0,0,0,0
0,0,2,0,0,0,0,1,1,0,0,1,0,55,0,6,0,0,2,0,0,24,0,0,0,0,0
0,0,0,0,0,0,0,64,0,0,0,5,0,0,0,20,0,1,0,0,0,0,5,0,0,0,0
0,0,0,0,0,0,0,10,6,0,0,7,3,3,0,20,1,1,20,17,1,2,0,0,0,0,0
0,0,0,9,0,0,0,17,4,0,0,9,1,0,0,14,20,2,7,4,2,0,1,0,0,0,0
0,0,0,0,0,0,0,20,0,4,0,7,1,5,0,17,8,1,13,15,0,0,0,0,0,0,0
0,0,0,0,0,0,0,3,1,0,0,1,6,0,0,5,2,0,64,6,2,0,0,0,0,0,0
0,0,0,0,0,0,1,6,3,0,0,1,1,0,0,4,1,0,9,49,6,1,4,3,0,2,0
0,0,0,1,0,0,0,13,4,0,0,1,3,0,0,4,2,1,13,4,42,0,2,1,0,0,0
0,2,20,0,1,0,0,0,1,0,16,0,0,0,9,2,0,0,1,0,0,35,0,0,0,0,2
0,0,1,0,0,0,8,2,0,0,0,0,0,0,31,2,0,0,0,0,0,0,45,0,1,0,0
0,0,0,0,0,18,0,0,0,0,0,0,0,0,0,0,0,0,0,0,0,0,1,76,0,0,0
0,0,6,0,0,0,1,4,1,0,2,0,0,0,10,11,0,0,8,0,1,0,5,3,35,0,1
0,0,1,2,0,0,0,3,5,0,0,0,0,0,7,3,0,0,10,1,3,1,8,0,0,45,0
0,0,0,0,0,0,0,1,0,0,14,0,0,0,0,1,0,0,1,0,0,0,0,11,0,0,65

}\mymatrix
\pgfplotstablegetrowsof{\mymatrix}%
\pgfmathtruncatemacro{\numrows}{\pgfplotsretval}%
\pgfplotstablegetcolsof{\mymatrix}%
\pgfmathtruncatemacro{\numcols}{\pgfplotsretval}%
\pgfmathtruncatemacro{\numnew}{\numrows*\numcols}%
\pgfmathdeclarefunction{myentry}{2}{%
\begingroup
\pgfkeys{/pgf/fpu,/pgf/fpu/output format=fixed}%
\pgfmathtruncatemacro{\myx}{#1}%
\pgfmathtruncatemacro{\myy}{#2}%
\pgfplotstablegetelem{\myy}{[index]\myx}\of\mymatrix%
\let\pgfmathresult\pgfplotsretval%
\pgfmathsmuggle\pgfmathresult
\endgroup}%

\pgfplotstablenew[
    create on use/y/.style={create col/expr={\the\numexpr\moduloop{\pgfplotstablerow}{\numrows}}},
    create on use/x/.style={create col/expr={\the\numexpr\truncdiv{\pgfplotstablerow}{\numcols}}},
    create on use/C/.style={create col/expr={myentry(\the\numexpr\truncdiv{\pgfplotstablerow}{\numcols},\the\numexpr\moduloop{\pgfplotstablerow}{\numrows})}},
    columns={x,y,C}
  ]
  {\numnew}%
  \newtable

\begin{tikzpicture}

\begin{axis}[axis equal,
             axis line style={draw=none},
             width=9cm,
             height=9cm,
             tick align=outside,
             /pgfplots/colormap={whiteblue}{rgb255(0cm)=(250,250,255);
                                           rgb255(1cm)=(54,125,189)},
             xtick=data,
             ticks=none,
             xtick pos=bottom,
             xticklabels={
             \footnotesize{class1},
             \footnotesize{class1},
             \footnotesize{class1}
             },
             ytick pos=left,
             ymin=-0.5,ymax=26.5,
             ytick=data,
             yticklabels={
             \footnotesize{class1},
             \footnotesize{class1},
             \footnotesize{class1}
             },
             ]
 \addplot [matrix plot,
           nodes={scale=0.4, color=black!70},
           nodes near coords,
           nodes near coords style={anchor=center},
           mesh/cols=\numcols,
           point meta=explicit] table [meta=C] {\newtable};

\end{axis}

\node (classes) [] at (3.75, -0.75) {\footnotesize{\textbf{Predicted labels}}};
\node (classes) [rotate=90] at (-0.75, 3.7) {\footnotesize{\textbf{Ground-truth labels}}};

\node (classes) [scale=0.563, align=left] at (0.0, 3.7) {
Electronic \\
Appliance \\
Food \\
Furniture \\
Indoor \\
Kitchen \\
Accessory \\
Animal \\
Outdoor \\
Person \\
Sports \\
Vehicle \\
Ceiling \\
Floor \\
Food \\
Furniture \\
Rawmaterial \\
Textile \\
Wall \\
Window \\
Building \\
Ground \\
Plant \\
Sky \\
Solid \\
Structural \\
Water
};

\node (classes) [scale=0.563, align=left, rotate=90] at (3.7, 0.0) {
Electronic \\
Appliance \\
Food \\
Furniture \\
Indoor \\
Kitchen \\
Accessory \\
Animal \\
Outdoor \\
Person \\
Sports \\
Vehicle \\
Ceiling \\
Floor \\
Food \\
Furniture \\
Rawmaterial \\
Textile \\
Wall \\
Window \\
Building \\
Ground \\
Plant \\
Sky \\
Solid \\
Structural \\
Water
};
\end{tikzpicture}
\end{subfigure}

\vspace{1.5em}
\begin{subfigure}{0.47\linewidth}
    \caption{Confusion matrix for STEGO\label{fig:coco_stego_conf}}
    \vspace{-1.5em}
    \def\moduloop#1#2{\ifnum\numexpr(#1-(#1/#2)*(#2))\relax<0(#1-(#1/#2)*(#2)+#2)\else(#1-(#1/#2)*(#2))\fi}
\def\truncdiv#1#2{((#1-\moduloop{#1}{#2})/(#2))}

\pgfplotstableread[col sep=comma]{
0,0,0,46,0,0,2,0,0,0,0,2,0,2,1,7,34,0,0,0,0,0,0,0,0,0,0
0,0,0,63,0,0,1,1,0,0,1,1,0,2,0,21,0,0,4,0,0,0,0,0,0,0,0
0,0,0,1,0,0,0,0,0,0,1,0,0,1,92,0,0,0,0,0,0,0,1,0,0,0,0
0,0,0,53,0,0,0,0,0,0,4,2,0,3,3,2,0,22,0,1,0,0,1,0,0,1,0
0,0,0,22,0,0,0,3,0,1,5,3,0,2,21,0,33,0,0,0,2,0,0,0,0,0,0
0,0,0,62,0,0,0,0,0,0,3,0,0,18,10,0,0,0,0,0,0,0,0,0,0,0,0
0,0,0,9,0,0,3,1,0,10,19,7,13,0,0,0,14,12,0,0,1,0,1,0,1,0,0
0,0,0,0,0,0,0,88,3,0,3,0,0,0,0,0,0,0,0,0,0,0,1,0,0,0,0
0,0,0,2,0,0,0,0,1,0,34,18,0,0,0,0,15,0,0,0,2,1,1,0,1,12,0
0,0,10,0,0,0,0,0,0,69,8,2,0,0,0,0,0,0,0,2,0,0,0,0,0,0,0
4,0,0,2,0,0,0,1,0,3,63,5,0,0,1,0,9,0,0,0,0,0,0,0,3,0,0
0,0,0,0,0,0,0,0,0,0,16,72,0,0,0,0,0,0,0,0,2,1,0,0,1,0,0
0,0,0,0,0,0,0,0,0,0,0,0,84,0,0,0,0,0,8,0,2,0,0,0,0,0,0
0,0,0,4,0,0,0,0,1,0,0,0,0,56,0,1,0,0,0,0,0,29,0,0,0,1,0
0,0,0,3,0,0,0,0,0,0,0,0,0,8,86,0,0,0,0,0,0,0,1,0,0,0,0
0,0,0,19,1,0,2,0,0,0,1,0,2,17,1,32,1,1,9,1,2,0,0,0,0,1,0
0,0,0,14,0,0,2,0,0,0,8,9,1,7,2,4,24,2,0,2,10,1,1,0,0,2,0
0,0,0,8,0,0,0,1,0,4,6,1,1,12,1,22,9,14,2,2,2,0,0,0,0,4,0
0,5,0,5,2,0,1,0,1,0,1,1,3,2,0,11,3,0,38,2,9,1,0,0,0,5,0
0,0,0,2,0,0,1,0,0,0,0,1,4,0,0,43,1,0,1,4,24,1,5,0,0,2,0
0,1,0,0,0,0,1,0,0,0,0,7,5,0,0,1,2,0,0,2,68,0,1,0,1,2,0
18,0,0,0,0,8,0,0,12,0,2,1,0,0,0,0,0,0,0,0,0,47,0,0,1,1,2
0,0,0,0,0,26,0,0,6,0,0,0,0,0,1,0,0,0,0,0,1,1,53,0,2,1,0
0,0,0,0,0,0,2,0,0,0,0,0,1,0,0,0,0,0,0,0,1,0,2,89,1,0,0
4,9,0,1,0,6,0,0,21,0,1,0,0,1,1,0,0,0,0,0,1,1,8,3,29,1,1
0,0,0,0,0,2,0,0,4,1,2,2,0,0,0,1,0,0,0,1,5,1,7,0,1,65,0
18,0,0,0,1,0,0,0,1,0,0,0,0,0,0,0,0,0,0,0,0,0,0,10,1,0,62

}\mymatrix
\pgfplotstablegetrowsof{\mymatrix}%
\pgfmathtruncatemacro{\numrows}{\pgfplotsretval}%
\pgfplotstablegetcolsof{\mymatrix}%
\pgfmathtruncatemacro{\numcols}{\pgfplotsretval}%
\pgfmathtruncatemacro{\numnew}{\numrows*\numcols}%
\pgfmathdeclarefunction{myentry}{2}{%
\begingroup
\pgfkeys{/pgf/fpu,/pgf/fpu/output format=fixed}%
\pgfmathtruncatemacro{\myx}{#1}%
\pgfmathtruncatemacro{\myy}{#2}%
\pgfplotstablegetelem{\myy}{[index]\myx}\of\mymatrix%
\let\pgfmathresult\pgfplotsretval%
\pgfmathsmuggle\pgfmathresult
\endgroup}%

\pgfplotstablenew[
    create on use/y/.style={create col/expr={\the\numexpr\moduloop{\pgfplotstablerow}{\numrows}}},
    create on use/x/.style={create col/expr={\the\numexpr\truncdiv{\pgfplotstablerow}{\numcols}}},
    create on use/C/.style={create col/expr={myentry(\the\numexpr\truncdiv{\pgfplotstablerow}{\numcols},\the\numexpr\moduloop{\pgfplotstablerow}{\numrows})}},
    columns={x,y,C}
  ]
  {\numnew}%
  \newtable

\begin{tikzpicture}

\begin{axis}[axis equal,
             axis line style={draw=none},
             width=9cm,
             height=9cm,
             tick align=outside,
             /pgfplots/colormap={whiteblue}{rgb255(0cm)=(250,250,255);
                                           rgb255(1cm)=(54,125,189)},
             xtick=data,
             ticks=none,
             xtick pos=bottom,
             xticklabels={
             \footnotesize{class1},
             \footnotesize{class1},
             \footnotesize{class1}
             },
             ytick pos=left,
             ymin=-0.5,ymax=26.5,
             ytick=data,
             yticklabels={
             \footnotesize{class1},
             \footnotesize{class1},
             \footnotesize{class1}
             },
             ]
 \addplot [matrix plot,
           nodes={scale=0.4, color=black!70},
           nodes near coords,
           nodes near coords style={anchor=center},
           mesh/cols=\numcols,
           point meta=explicit] table [meta=C] {\newtable};

\end{axis}

\node (classes) [] at (3.75, -0.75) {\footnotesize{\textbf{Predicted labels}}};
\node (classes) [rotate=90] at (-0.75, 3.7) {\footnotesize{\textbf{Ground-truth labels}}};

\node (classes) [scale=0.563, align=left] at (0.0, 3.7) {
Electronic \\
Appliance \\
Food \\
Furniture \\
Indoor \\
Kitchen \\
Accessory \\
Animal \\
Outdoor \\
Person \\
Sports \\
Vehicle \\
Ceiling \\
Floor \\
Food \\
Furniture \\
Rawmaterial \\
Textile \\
Wall \\
Window \\
Building \\
Ground \\
Plant \\
Sky \\
Solid \\
Structural \\
Water
};

\node (classes) [scale=0.563, align=left, rotate=90] at (3.7, 0.0) {
Electronic \\
Appliance \\
Food \\
Furniture \\
Indoor \\
Kitchen \\
Accessory \\
Animal \\
Outdoor \\
Person \\
Sports \\
Vehicle \\
Ceiling \\
Floor \\
Food \\
Furniture \\
Rawmaterial \\
Textile \\
Wall \\
Window \\
Building \\
Ground \\
Plant \\
Sky \\
Solid \\
Structural \\
Water
};
\end{tikzpicture}
\end{subfigure}
\hspace{0.03\linewidth}
\begin{subfigure}{0.47\linewidth}
    \caption{Confusion matrix for STEGO\,+\,\ourmethodframework\label{fig:coco_stegoours_conf}}
    \vspace{-1.5em}

\def\moduloop#1#2{\ifnum\numexpr(#1-(#1/#2)*(#2))\relax<0(#1-(#1/#2)*(#2)+#2)\else(#1-(#1/#2)*(#2))\fi}
\def\truncdiv#1#2{((#1-\moduloop{#1}{#2})/(#2))}

\pgfplotstableread[col sep=comma]{
0,0,0,46,0,0,0,0,0,0,0,0,0,1,0,28,19,0,0,0,1,0,0,0,0,0,0
0,0,0,55,0,0,0,1,0,0,0,0,0,2,0,32,0,0,3,1,0,1,0,0,0,0,0
0,0,0,2,0,0,0,0,2,0,0,0,0,1,91,0,1,0,0,0,0,0,0,0,0,0,0
0,0,0,51,0,0,0,0,2,0,0,0,0,4,3,4,0,23,0,0,1,2,1,0,0,2,0
0,0,0,21,0,0,0,3,5,1,0,0,0,2,20,1,35,1,1,0,4,0,0,0,0,0,0
0,0,0,62,0,0,0,0,2,0,0,0,0,20,7,2,0,0,0,0,0,0,0,0,0,0,0
0,0,0,12,0,0,0,0,14,13,0,0,1,0,0,1,30,11,0,0,3,5,1,0,0,0,0
0,0,1,0,0,3,0,84,2,1,0,0,0,0,0,0,0,0,0,1,0,0,2,0,0,0,0
0,0,1,1,0,1,0,0,31,0,0,4,0,0,0,0,16,0,0,7,9,10,1,0,0,10,0
10,0,0,0,0,0,2,0,3,71,1,0,0,0,0,0,0,0,0,0,1,2,1,0,0,1,0
0,0,0,1,0,0,0,0,45,4,8,0,0,0,0,0,14,0,0,0,0,10,0,3,0,1,2
0,0,0,0,0,0,0,0,14,0,0,61,0,0,0,0,1,0,0,0,9,5,1,1,0,1,0
0,0,0,0,0,0,0,0,0,0,0,0,77,0,0,0,1,0,3,0,13,0,0,0,0,1,0
0,0,0,3,0,1,0,0,0,0,0,0,0,58,0,2,0,0,0,1,0,30,0,0,0,0,0
0,0,0,3,0,0,0,0,0,0,0,0,0,5,89,0,0,0,0,0,0,0,0,0,0,0,0
0,0,0,17,0,0,0,0,0,0,0,0,1,18,0,37,1,1,8,3,3,1,0,0,0,1,0
0,0,0,15,0,0,0,0,4,0,0,2,0,7,2,5,23,1,0,6,16,5,1,0,0,2,0
0,0,0,7,0,0,0,1,3,5,0,0,0,13,0,14,8,27,1,1,4,1,0,0,0,7,0
0,0,0,3,0,0,0,0,0,0,0,0,2,1,0,13,2,0,38,17,5,1,0,0,0,8,0
0,0,0,2,0,0,0,0,0,0,0,0,1,0,0,46,0,0,1,3,32,2,6,0,0,1,0
0,0,0,0,0,0,0,0,0,0,0,0,2,0,0,0,2,0,0,4,77,2,2,0,0,2,0
0,0,8,0,0,15,0,0,0,0,17,0,0,0,0,0,0,0,0,1,0,50,1,0,0,0,1
0,0,26,0,0,6,0,0,0,0,0,0,0,0,1,0,0,0,0,2,1,1,56,0,1,0,0
0,0,0,0,2,0,0,0,0,0,0,0,0,0,0,0,0,0,0,0,1,0,2,91,0,0,0
0,0,6,1,0,8,0,0,0,0,2,0,0,0,0,0,0,0,0,26,1,3,8,2,30,0,1
0,0,3,0,0,1,0,0,0,0,0,0,0,0,0,0,0,0,0,4,7,3,8,0,0,66,0
0,0,0,0,0,1,0,0,0,0,15,0,0,0,0,0,0,0,0,0,0,1,1,11,0,0,66

}\mymatrix
\pgfplotstablegetrowsof{\mymatrix}%
\pgfmathtruncatemacro{\numrows}{\pgfplotsretval}%
\pgfplotstablegetcolsof{\mymatrix}%
\pgfmathtruncatemacro{\numcols}{\pgfplotsretval}%
\pgfmathtruncatemacro{\numnew}{\numrows*\numcols}%
\pgfmathdeclarefunction{myentry}{2}{%
\begingroup
\pgfkeys{/pgf/fpu,/pgf/fpu/output format=fixed}%
\pgfmathtruncatemacro{\myx}{#1}%
\pgfmathtruncatemacro{\myy}{#2}%
\pgfplotstablegetelem{\myy}{[index]\myx}\of\mymatrix%
\let\pgfmathresult\pgfplotsretval%
\pgfmathsmuggle\pgfmathresult
\endgroup}%

\pgfplotstablenew[
    create on use/y/.style={create col/expr={\the\numexpr\moduloop{\pgfplotstablerow}{\numrows}}},
    create on use/x/.style={create col/expr={\the\numexpr\truncdiv{\pgfplotstablerow}{\numcols}}},
    create on use/C/.style={create col/expr={myentry(\the\numexpr\truncdiv{\pgfplotstablerow}{\numcols},\the\numexpr\moduloop{\pgfplotstablerow}{\numrows})}},
    columns={x,y,C}
  ]
  {\numnew}%
  \newtable


\begin{tikzpicture}

\begin{axis}[axis equal,
             axis line style={draw=none},
             width=9cm,
             height=9cm,
             tick align=outside,
             /pgfplots/colormap={whiteblue}{rgb255(0cm)=(250,250,255);
                                           rgb255(1cm)=(54,125,189)},
             xtick=data,
             ticks=none,
             xtick pos=bottom,
             xticklabels={
             \footnotesize{class1},
             \footnotesize{class1},
             \footnotesize{class1}
             },
             ytick pos=left,
             ymin=-0.5,ymax=26.5,
             ytick=data,
             yticklabels={
             \footnotesize{class1},
             \footnotesize{class1},
             \footnotesize{class1}
             },
             ]
 \addplot [matrix plot,
           nodes={scale=0.4, color=black!70},
           nodes near coords,
           nodes near coords style={anchor=center},
           mesh/cols=\numcols,
           point meta=explicit] table [meta=C] {\newtable};

\end{axis}

\node (classes) [] at (3.75, -0.75) {\footnotesize{\textbf{Predicted labels}}};
\node (classes) [rotate=90] at (-0.75, 3.7) {\footnotesize{\textbf{Ground-truth labels}}};

\node (classes) [scale=0.563, align=left] at (0.0, 3.7) {
Electronic \\
Appliance \\
Food \\
Furniture \\
Indoor \\
Kitchen \\
Accessory \\
Animal \\
Outdoor \\
Person \\
Sports \\
Vehicle \\
Ceiling \\
Floor \\
Food \\
Furniture \\
Rawmaterial \\
Textile \\
Wall \\
Window \\
Building \\
Ground \\
Plant \\
Sky \\
Solid \\
Structural \\
Water
};

\node (classes) [scale=0.563, align=left, rotate=90] at (3.7, 0.0) {
Electronic \\
Appliance \\
Food \\
Furniture \\
Indoor \\
Kitchen \\
Accessory \\
Animal \\
Outdoor \\
Person \\
Sports \\
Vehicle \\
Ceiling \\
Floor \\
Food \\
Furniture \\
Rawmaterial \\
Textile \\
Wall \\
Window \\
Building \\
Ground \\
Plant \\
Sky \\
Solid \\
Structural \\
Water
};
\end{tikzpicture}

\end{subfigure}
\end{figure}

\subsection{Class-Level Quantitative Analysis}
To gain a deeper understanding of \ourmethodframework, we assess the segmentation accuracy in terms of IoU for individual classes. Additionally, we present the confusion matrices among the semantic classes for the DINO ViT-B/8 Baseline, \ourmethodframework, STEGO, and STEGO+\ourmethodframework for the COCO-Stuff dataset in \cref{fig:coco_confusion}. Generally, we can observe that for both the DINO and STEGO baseline, for certain classes (\eg, ``Appliance'', ``Indoor'', ``Kitchen'') the discovered unsupervised class concept does not correlate with human-defined semantic classes. This suggests that the respective backbone feature representation may already be hard to separate. Furthermore, some of the 27 intermediate COCO-Stuff classes merge visually distinct concepts. For instance, the class ``indoor'' combines ``hairbrush'', ``toothbrush'', ``hair dryer'', ``teddy bear'', ``scissors'', ``vase'', ``clock'', and ``book''. We see this assumption partially confirmed by analyzing the class IoUs of linear probing of the DINO features. For the problematic classes, the linear probing IoUs are in the range of approx.\ 16\:\%--30\:\%, whereas for the other classes the IoU is 50\:\% and higher. We conclude that if it is already difficult to linearly distinguish the classes based on the backbone features, our method can hardly improve upon this. 

However, for classes meeting this requirement, our method clearly boosts class IoUs. In rare cases (\eg, STEGO comparison to STEGO+\ourmethodframework for classes ``Outdoor'' and ``Sports''), there is a decrease in one class IoU with \ourmethodframework while another class IoU increases. This can occur if the change in the prototype representation results in a change of the Hungarian matching for evaluation, though this is rarely observed.
In terms of class confusion, the model's predictions align with the ground-truth labels. Furthermore, the existing confusions are reasonable. For instance, when using STEGO, confusions of the two ``food'' mid-level classes emerge. Overall, \cref{tab:cococlsiou} indicates that our method either enhances or at least maintains the segmentation performance per class in terms of IoU regardless of the backbone model. Additionally, our method aids in reducing the confusion among classes.

\subsection{Qualitative Comparison to HP \label{sec:qual_comp_hp}}
Similar to the qualitative comparison in \cref{fig:qualitative} in the main paper, we aim to compare \ourmethodframework with HP \citep{Seong:2023:LHP}. We present qualitative examples for the baseline, \ourmethodframework, HP, and the combination of HP and \ourmethodframework in \cref{fig:qualitative_hp}. These qualitative examples align with the findings from the quantitative results in the main paper (\cf \cref{tab:main_cs,tab:main_coco,tab:main_pd}). It is evident that our method produces locally more consistent and less noisy results compared to both baselines.
Despite the already impressive qualitative results of HP on COCO-Stuff, our method excels in correcting misclassifications and achieving better segmentation of object boundaries. We observe one limitation, particularly with the DINO ViT-S/8 baseline, both independently and in conjunction with \ourmethodframework for the Potsdam-3 dataset. In this case, the semantic concept of ``street'' is not recognized and is consequently rarely predicted.

\def\imgwidth{0.1035\linewidth}
\begin{figure*}[!t]
    \small
    \smallskip
    \setlength\tabcolsep{1.0pt}
    \renewcommand{\arraystretch}{0.666}
    \centering
       
    \begin{tabularx}{\linewidth}{@{}Xccccccccc@{}}
    & \multicolumn{3}{c}{Cityscapes} & \multicolumn{3}{c}{COCO-Stuff} & \multicolumn{3}{c}{Potsdam-3} \\
    \cmidrule(l{0.5em}r{0.5em}){2-4} \cmidrule(l{0.5em}r{0.5em}){5-7} \cmidrule(l{0.5em}r{0.5em}){8-10}

    \rotatebox[origin=lB]{90}{\scriptsize \hspace{-31.8em}\shortstack{HP + \\\ourmethodframework} \hspace{1.9em} \shortstack{\vspace{0.4em} HP \hspace{2.0em} \ourmethodframework\hspace{1.0em} Baseline \hspace{1.5em}Ground truth \hspace{0.9em} Image}}
    & \includegraphics[width=\imgwidth]
    {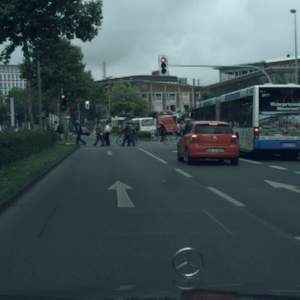} 
    & \includegraphics[width=\imgwidth]
    {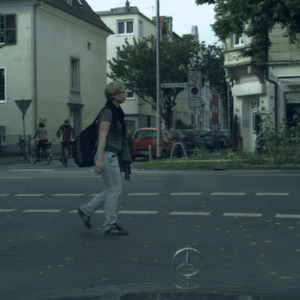} 
    & \includegraphics[width=\imgwidth]
    {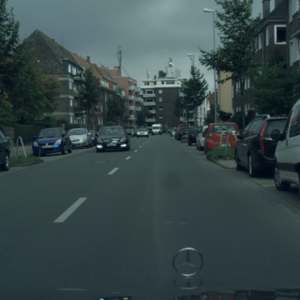} 
    & \includegraphics[width=\imgwidth]
    {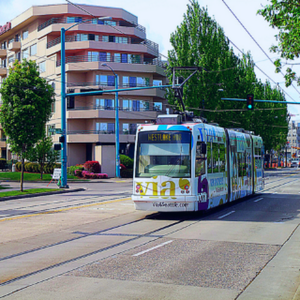} 
    & \includegraphics[width=\imgwidth]
    {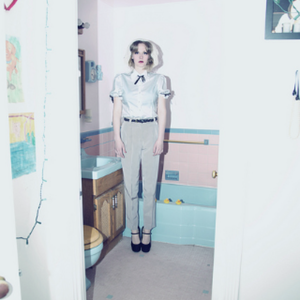} 
    & \includegraphics[width=\imgwidth]
    {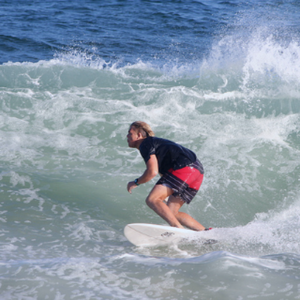}  
    & \includegraphics[width=\imgwidth]
    {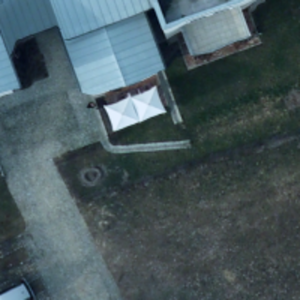}  
    & \includegraphics[width=\imgwidth]
    {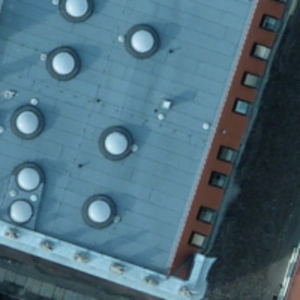} 
    & \includegraphics[width=\imgwidth]
    {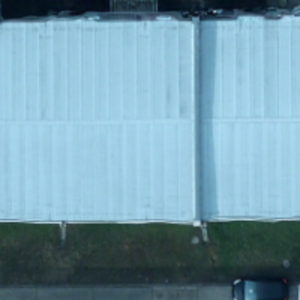}  \\
    & \includegraphics[width=\imgwidth]
    {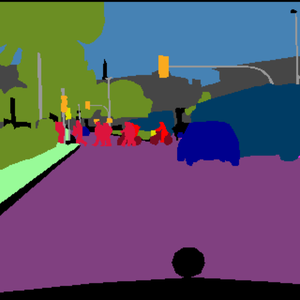} 
    & \includegraphics[width=\imgwidth]
    {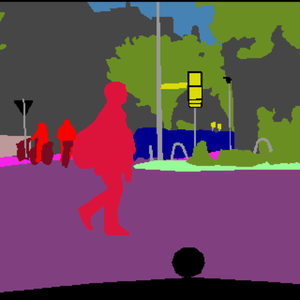} 
    & \includegraphics[width=\imgwidth]
    {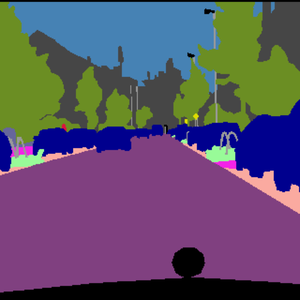} 
    & \includegraphics[width=\imgwidth]
    {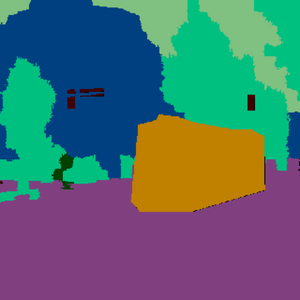} 
    & \includegraphics[width=\imgwidth]
    {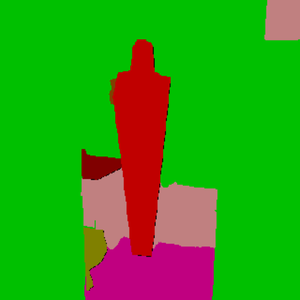} 
    & \includegraphics[width=\imgwidth]
    {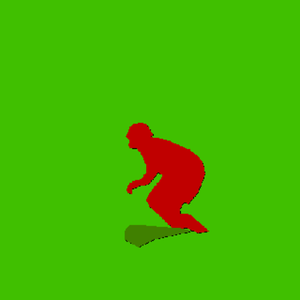}
    & \includegraphics[width=\imgwidth]
    {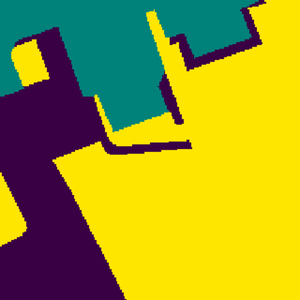}  
    & \includegraphics[width=\imgwidth]
    {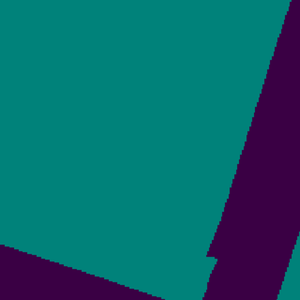} 
    & \includegraphics[width=\imgwidth]
    {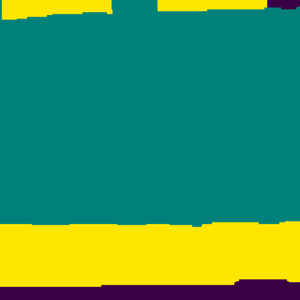}  \\
    & \includegraphics[width=\imgwidth]
    {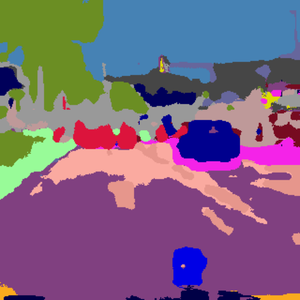} 
    & \includegraphics[width=\imgwidth]
    {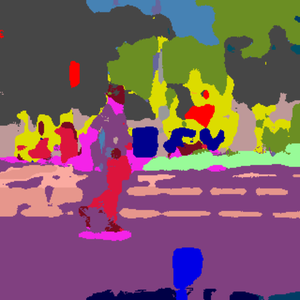} 
    & \includegraphics[width=\imgwidth]
    {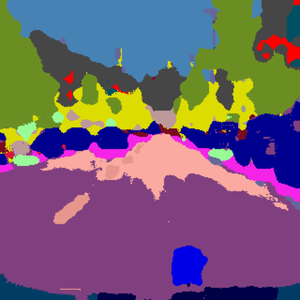} 
    & \includegraphics[width=\imgwidth]
    {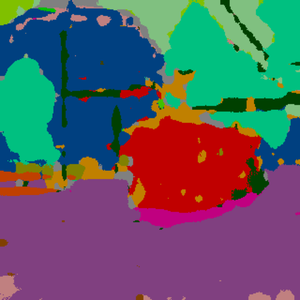} 
    & \includegraphics[width=\imgwidth]
    {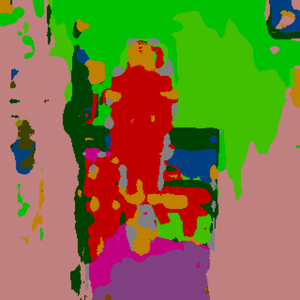} 
    & \includegraphics[width=\imgwidth]
    {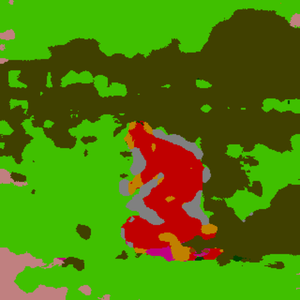}
    & \includegraphics[width=\imgwidth]
    {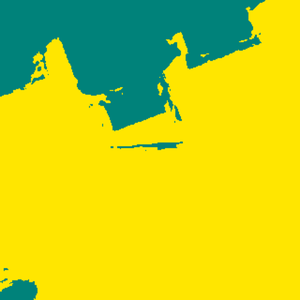}  
    & \includegraphics[width=\imgwidth]
    {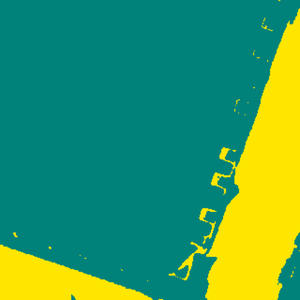} 
    & \includegraphics[width=\imgwidth]
    {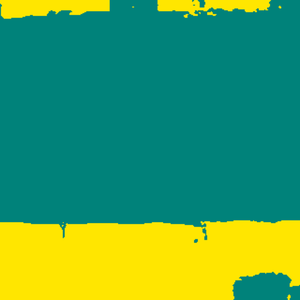}  \\
    & \includegraphics[width=\imgwidth]
    {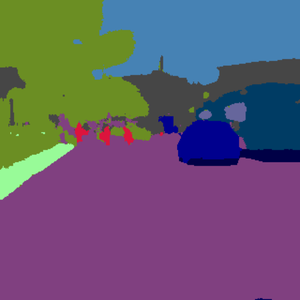} 
    & \includegraphics[width=\imgwidth]
    {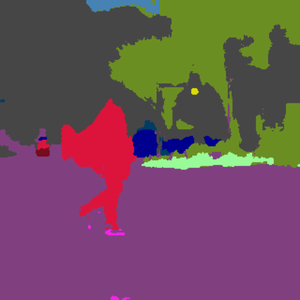} 
    & \includegraphics[width=\imgwidth]
    {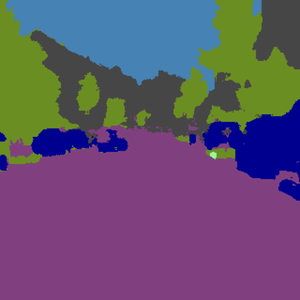} 
    & \includegraphics[width=\imgwidth]
    {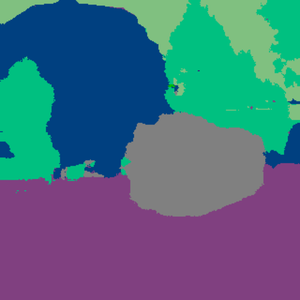} 
    & \includegraphics[width=\imgwidth]
    {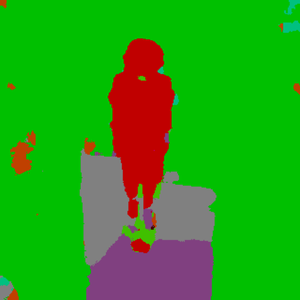} 
    & \includegraphics[width=\imgwidth]
    {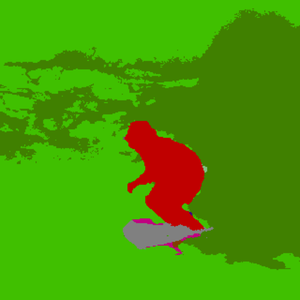}
    & \includegraphics[width=\imgwidth]
    {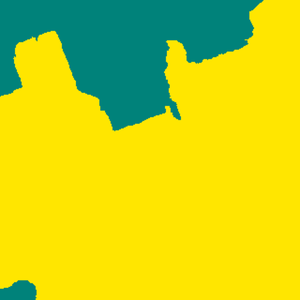}  
    & \includegraphics[width=\imgwidth]
    {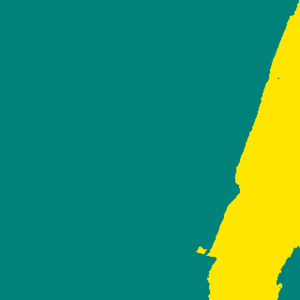} 
    & \includegraphics[width=\imgwidth]
    {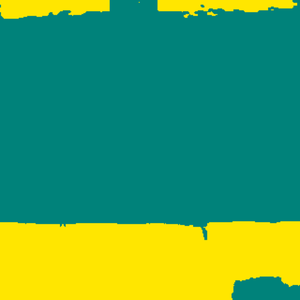}  \\
    & \includegraphics[width=\imgwidth]
    {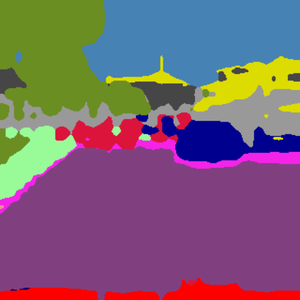} 
    & \includegraphics[width=\imgwidth]
    {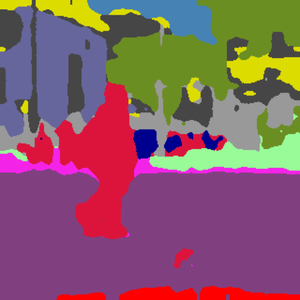} 
    & \includegraphics[width=\imgwidth]
    {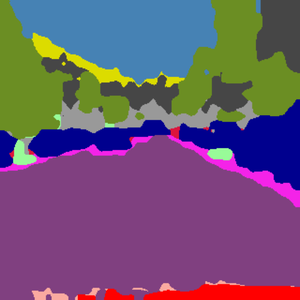} 
    & \includegraphics[width=\imgwidth]
    {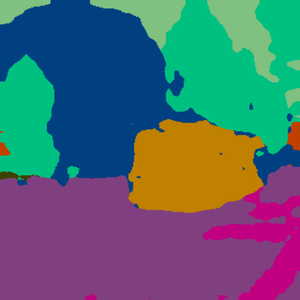} 
    & \includegraphics[width=\imgwidth]
    {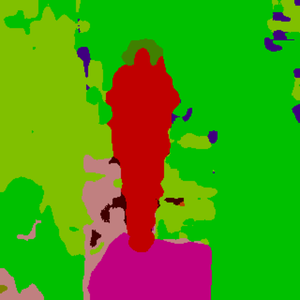} 
    & \includegraphics[width=\imgwidth]
    {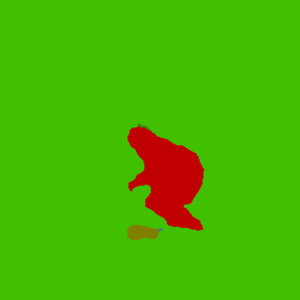}
    & \includegraphics[width=\imgwidth]
    {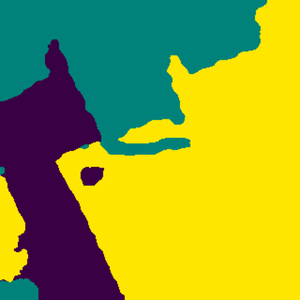}  
    & \includegraphics[width=\imgwidth]
    {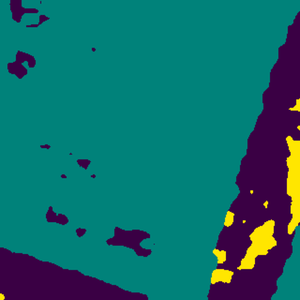} 
    & \includegraphics[width=\imgwidth]
    {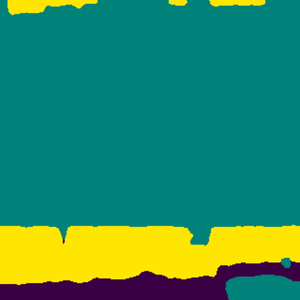}  \\
    
    & \includegraphics[width=\imgwidth]
    {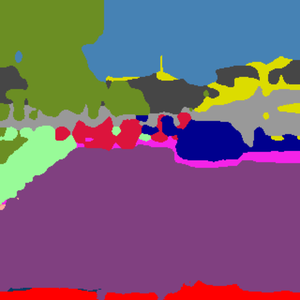} 
    & \includegraphics[width=\imgwidth]
    {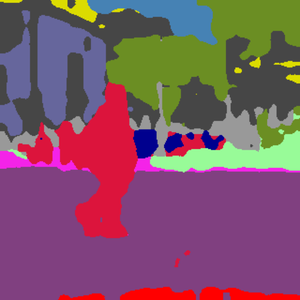} 
    & \includegraphics[width=\imgwidth]
    {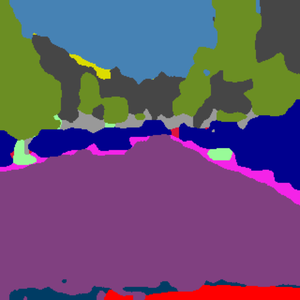} 
    & \includegraphics[width=\imgwidth]
    {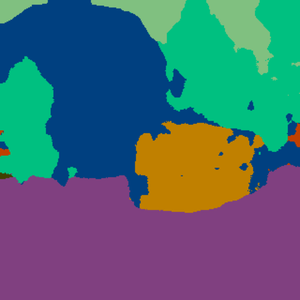} 
    & \includegraphics[width=\imgwidth]
    {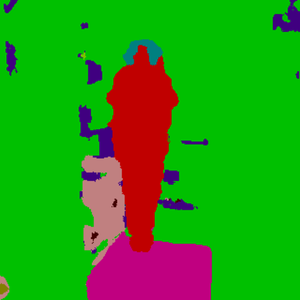} 
    & \includegraphics[width=\imgwidth]
    {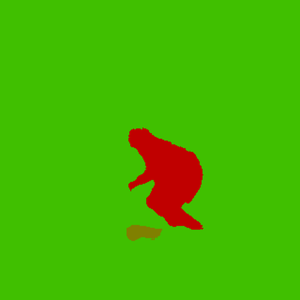}
    & \includegraphics[width=\imgwidth]
    {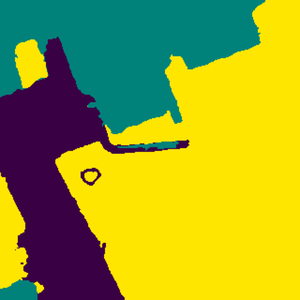}  
    & \includegraphics[width=\imgwidth]
    {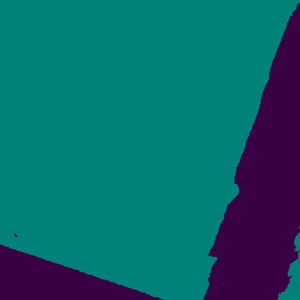} 
    & \includegraphics[width=\imgwidth]
    {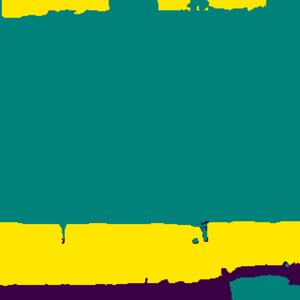}  \\
    \end{tabularx}
    \vspace{-0.5em}
    \caption{\textbf{Qualitative results} for the DINO ViT-S/8 Baseline, \ourmethodframework \textit{(Ours)}, HP \citep{Seong:2023:LHP}, and HP \citep{Seong:2023:LHP}+\ourmethodframework \textit{(Ours)} for all three datasets. Our method produces locally more consistent segmentation results, reducing misclassification. \label{fig:qualitative_hp}}
\end{figure*}

\begin{table}[t]
    \small
    \caption{\cng{\textbf{Comparing \ourmethodframework \textit{(Ours)} to HP using DINOv2} for the Cityscapes (ViT-B/14), COCO-Stuff (ViT-B/14), and Potsdam-3 (ViT-S/14) datasets. We report Accuracy and mean IoU (in \%) for unsupervised probing.}}
    \label{tab:hp_dinov2}
    \vspace{-0.5em}
    \smallskip
    \centering
    \setlength\tabcolsep{2pt}
    \begin{tabularx}{\linewidth}{@{}Xcccccc@{}}
        \toprule
        \multirow{2}{*}{\vspace{-0.55em}\textbf{Method}} & \multicolumn{2}{c}{\textit{Cityscapes}} & \multicolumn{2}{c}{\textit{COCO-Stuff}} & \multicolumn{2}{c}{\textit{Potsdam-3}} \\
        \cmidrule(l{0.5em}r{0.5em}){2-3} \cmidrule(l{0.5em}r{0.5em}){4-5} \cmidrule(l{0.5em}r{0.5em}){6-7}
        & \textbf{Acc} & \textbf{mIoU} & \textbf{Acc} & \textbf{mIoU}  & \textbf{Acc} & \textbf{mIoU}  \\
        \midrule 
        DINOv2 Baseline \citep{Oquab:2023:DLR}                       & 49.5 & 15.3 & 44.5 & 22.9 & 82.4 & 69.9 \\
        +\,HP \citep{Seong:2023:LHP}                                 & 67.9 & 15.9 & 48.9 & 19.8 & 79.4 & 65.7 \\
        \rowcolor{mylightgray}
        +\,\ourmethodframework                                       & 71.6 & \bn{19.0} & 46.4 & \bn{23.8} & \bn{83.1} & \bn{71.0} \\
        \rowcolor{mylightgray}
        +\,HP \citep{Seong:2023:LHP}\,+\,\ourmethodframework         & \bn{74.3} & 16.6 & \bn{49.3} & 20.2 & 79.6 & 66.0 \\
        \bottomrule
    \end{tabularx}
\end{table}

\def\imgwidth{0.1035\linewidth}
\begin{figure}[!t]
    \small
    \smallskip
    \setlength\tabcolsep{1.0pt}
    \renewcommand{\arraystretch}{0.6666}
    \centering

    \begin{tabularx}{\linewidth}{@{}Xccccccccc@{}}
    & \multicolumn{3}{c}{Cityscapes} & \multicolumn{3}{c}{COCO-Stuff} & \multicolumn{3}{c}{Potsdam-3} \\
    \cmidrule(l{0.5em}r{0.5em}){2-4} \cmidrule(l{0.5em}r{0.5em}){5-7} \cmidrule(l{0.5em}r{0.5em}){8-10}
    
    \rotatebox[origin=lB]{90}{\scriptsize{\hspace{-31.8em}\shortstack{DINOv2 +\\ \ourmethodframework} \hspace{1.0em} \shortstack{DINOv2\\ Baseline} \hspace{1.2em} \shortstack{DINO +\\ \ourmethodframework} \hspace{0.9em} \shortstack{DINO\\ Baseline} \hspace{1.2em}Ground Truth \hspace{1.2em} Image}}
    & \includegraphics[width=\imgwidth]
    {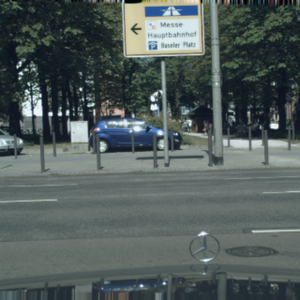} 
    & \includegraphics[width=\imgwidth]
    {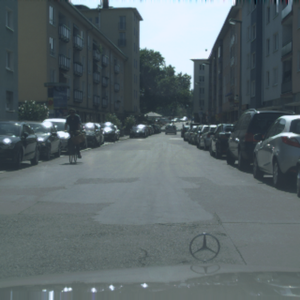} 
    & \includegraphics[width=\imgwidth]
    {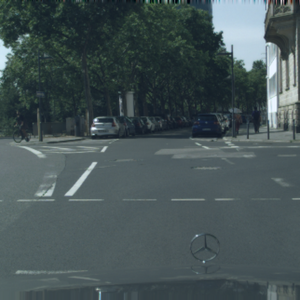} 
    & \includegraphics[width=\imgwidth]
    {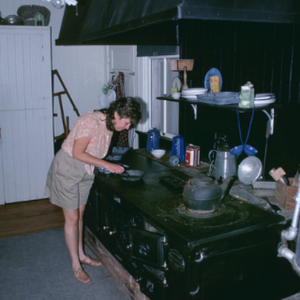} 
    & \includegraphics[width=\imgwidth]
    {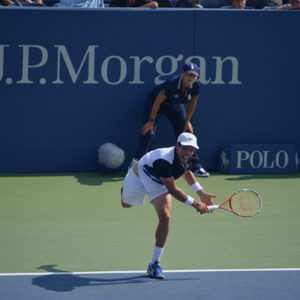} 
    & \includegraphics[width=\imgwidth]
    {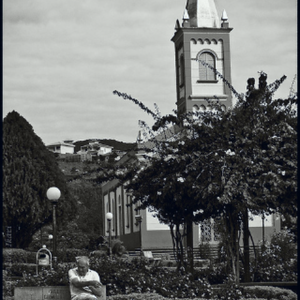} 
    & \includegraphics[width=\imgwidth]
    {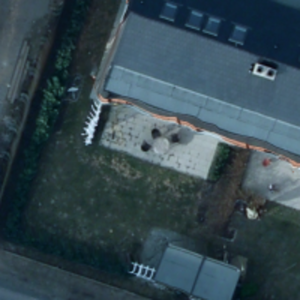} 
    & \includegraphics[width=\imgwidth]
    {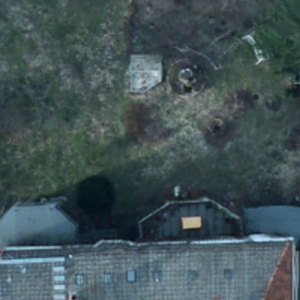} 
    & \includegraphics[width=\imgwidth]
    {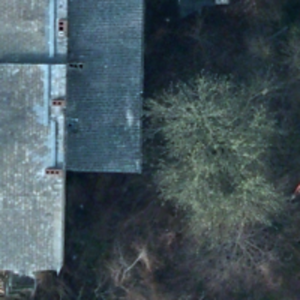}  \\
    & \includegraphics[width=\imgwidth]
    {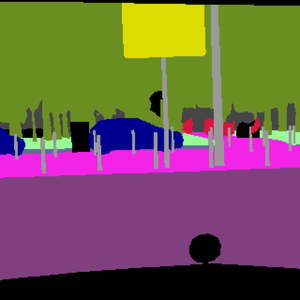} 
    & \includegraphics[width=\imgwidth]
    {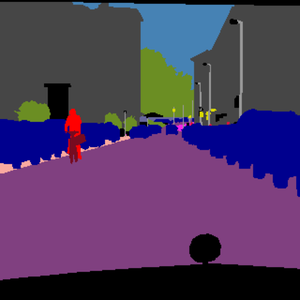} 
    & \includegraphics[width=\imgwidth]
    {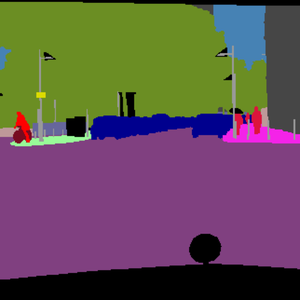} 
    & \includegraphics[width=\imgwidth]
    {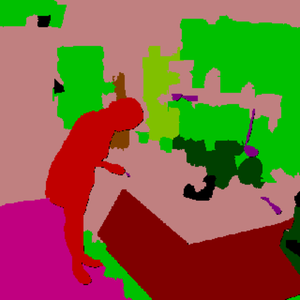} 
    & \includegraphics[width=\imgwidth]
    {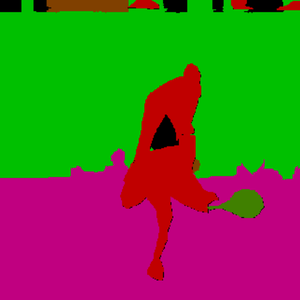} 
    & \includegraphics[width=\imgwidth]
    {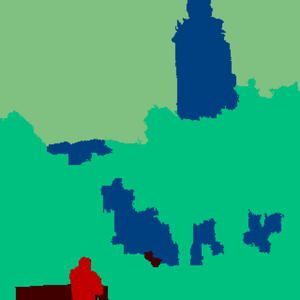} 
    & \includegraphics[width=\imgwidth]
    {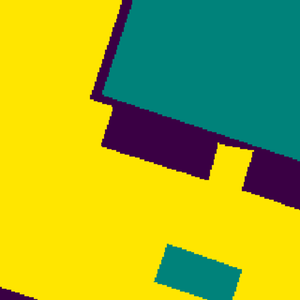} 
    & \includegraphics[width=\imgwidth]
    {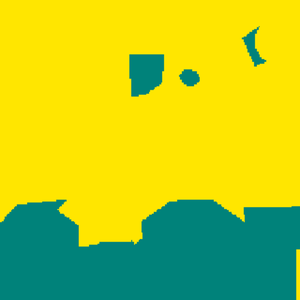} 
    & \includegraphics[width=\imgwidth]
    {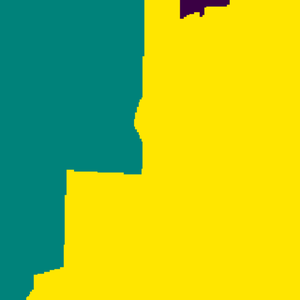}  \\
    & \includegraphics[width=\imgwidth]
    {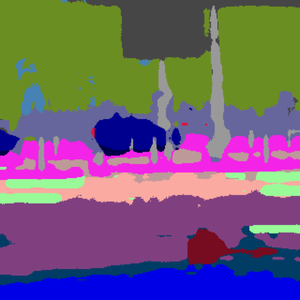} 
    & \includegraphics[width=\imgwidth]
    {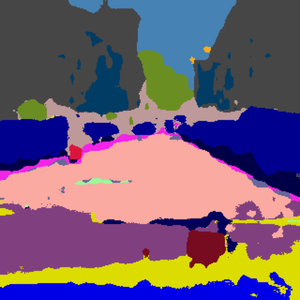} 
    & \includegraphics[width=\imgwidth]
    {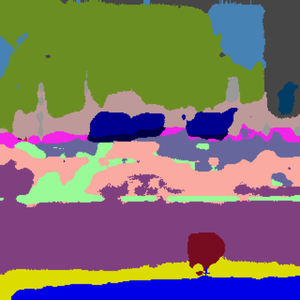} 
    & \includegraphics[width=\imgwidth]
    {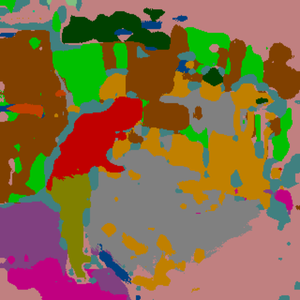} 
    & \includegraphics[width=\imgwidth]
    {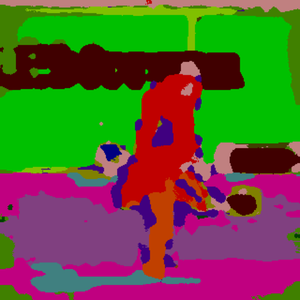} 
    & \includegraphics[width=\imgwidth]
    {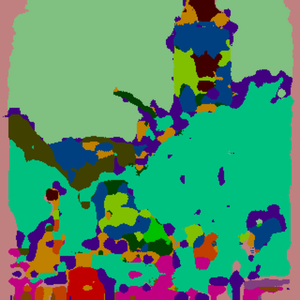} 
    & \includegraphics[width=\imgwidth]
    {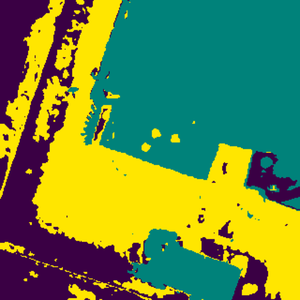} 
    & \includegraphics[width=\imgwidth]
    {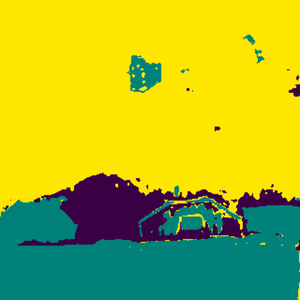} 
    & \includegraphics[width=\imgwidth]
    {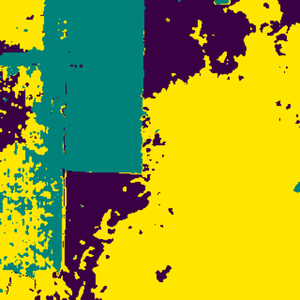}  \\
    & \includegraphics[width=\imgwidth]
    {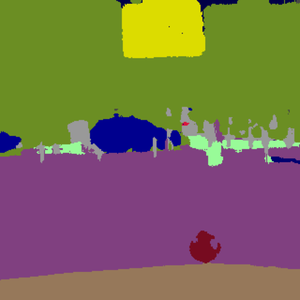} 
    & \includegraphics[width=\imgwidth]
    {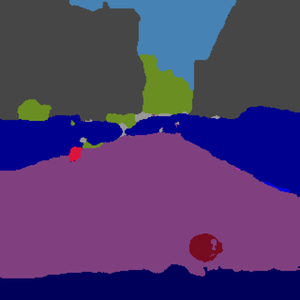} 
    & \includegraphics[width=\imgwidth]
    {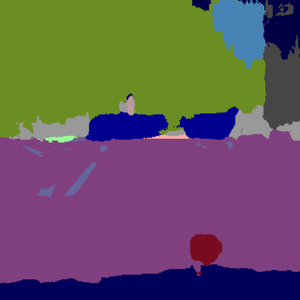} 
    & \includegraphics[width=\imgwidth]
    {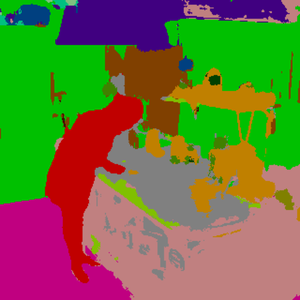} 
    & \includegraphics[width=\imgwidth]
    {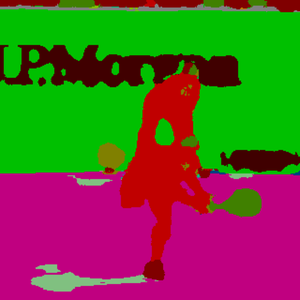} 
    & \includegraphics[width=\imgwidth]
    {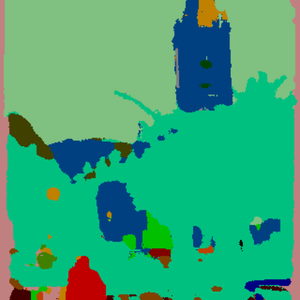} 
    & \includegraphics[width=\imgwidth]
    {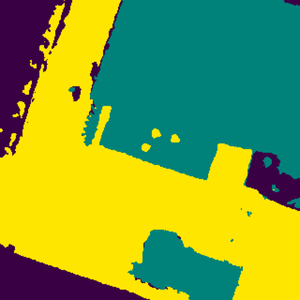} 
    & \includegraphics[width=\imgwidth]
    {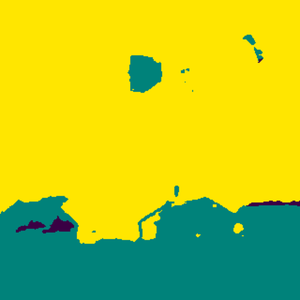} 
    & \includegraphics[width=\imgwidth]
    {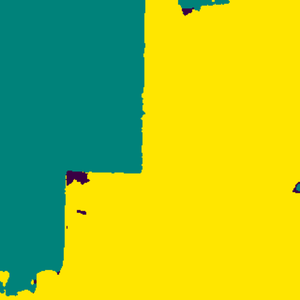}  \\
    & \includegraphics[width=\imgwidth]
    {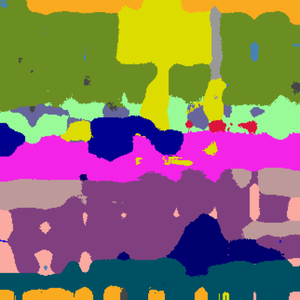} 
    & \includegraphics[width=\imgwidth]
    {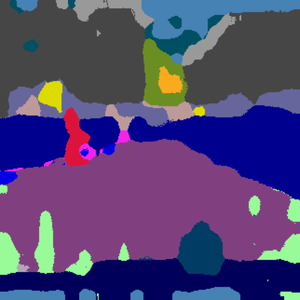} 
    & \includegraphics[width=\imgwidth]
    {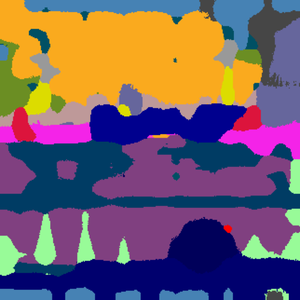} 
    & \includegraphics[width=\imgwidth]
    {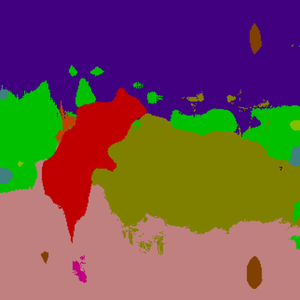} 
    & \includegraphics[width=\imgwidth]
    {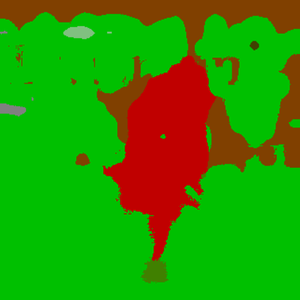} 
    & \includegraphics[width=\imgwidth]
    {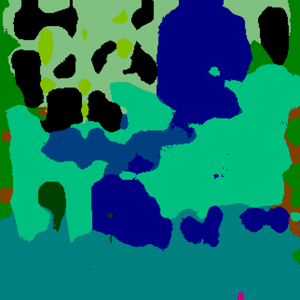} 
    & \includegraphics[width=\imgwidth]
    {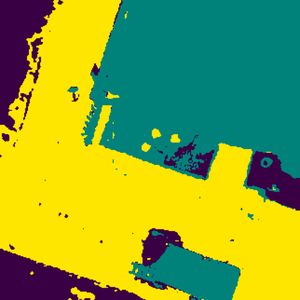} 
    & \includegraphics[width=\imgwidth]
    {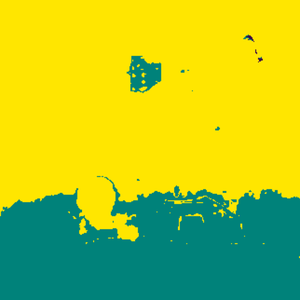} 
    & \includegraphics[width=\imgwidth]
    {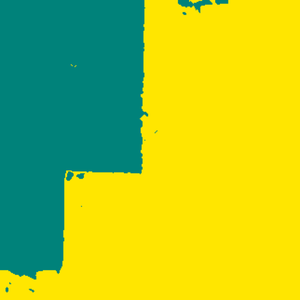}  \\
    
    & \includegraphics[width=\imgwidth]
    {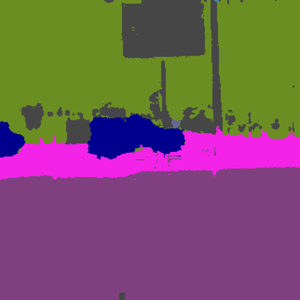} 
    & \includegraphics[width=\imgwidth]
    {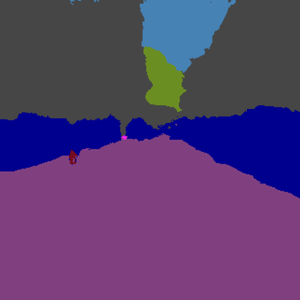} 
    & \includegraphics[width=\imgwidth]
    {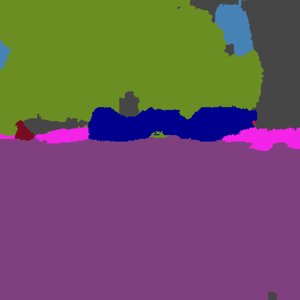} 
    & \includegraphics[width=\imgwidth]
    {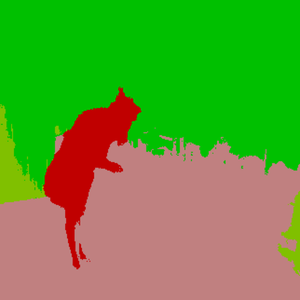} 
    & \includegraphics[width=\imgwidth]
    {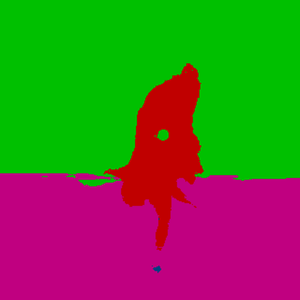} 
    & \includegraphics[width=\imgwidth]
    {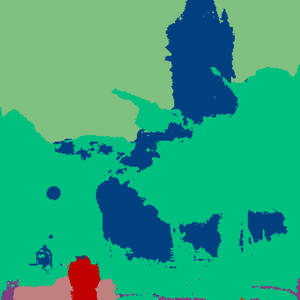} 
    & \includegraphics[width=\imgwidth]
    {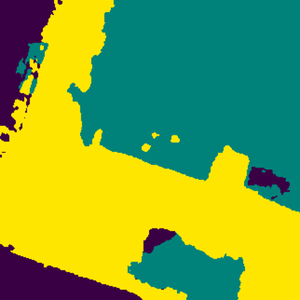} 
    & \includegraphics[width=\imgwidth]
    {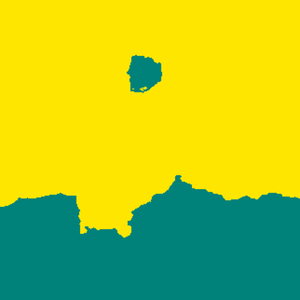} 
    & \includegraphics[width=\imgwidth]
    {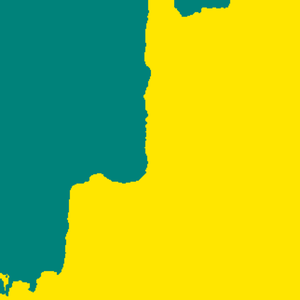}  \\
    \end{tabularx}

    \caption{\rebut{\textbf{Qualitative comparison} of DINO and DINOv2 using ViT-B/8 and ViT-B/14 respectively. We show the DINO Baseline, DINO+\ourmethodframework \textit{(Ours)}, DINOv2 Baseline and DINOv2+\ourmethodframework \textit{(Ours)} for Cityscapes, COCO-Stuff, and Potsdam-3. \label{fig:dinovsdinov2}}}
\end{figure}

\subsection{Comparison to HP using DINOv2}
Current state-of-the-art methods \cite{Hamilton:2022:USS, Seong:2023:LHP} for unsupervised segmentation do not provide experiments using DINOv2 features. To be able to compare with previous methods, we train HP using DINOv2 for the comparison in \cref{tab:hp_dinov2}. We strictly follow the training schedule and hyperparameters used in the original implementation for all respective datasets. HP does not generalize well to the DINOv2 features and hardly keeps up with the strong baseline. \ourmethodframework moderately but consistently improves upon both DINOv2 and HP across all datasets and metrics using the identical \ourmethodframework hyperparameters we use across all other experiments in this work.

\subsection{\rebut{Qualitative Comparison of DINO and DINOv2}}
\rebut{Our experiments could not identify any clear differences between DINOv1 and DINOv2 and the different transformer architectures in terms of the downstream task performance of unsupervised semantic segmentation. Both methods provide excellent feature embeddings and similar quantitative results as baselines. When applying \ourmethodframework, DINOv2 often provides slightly better quantitative results but coarser segmentation masks due to the larger patches (\cf \cref{tab:main_cs,tab:main_coco,tab:main_pd}). We visualize both the DINO and DINOv2 baselines and with \ourmethodframework in \cref{fig:dinovsdinov2}.}


\subsection{Qualitative \ourmethod Pseudo-Label Examples \label{sec:qual_pseudolabels}}
Following the quantitative assessment of the pseudo labels in \cref{sec:quanti_pseudolabels}, \cref{fig:oraclemask_examples} visualizes qualitative examples of the \ourmethod mask proposals and pseudo labels. We visualize individual masks, each in a different color (\emph{\ourmethod Colored}). We also display oracle pseudo labels, assigning each mask a color based on the ground-truth label (\emph{\ourmethod Oracle class IDs}). We observe that the mask proposals align well with the ground-truth labels across all three datasets, generalizing across three distinct domains. \ourmethod effectively partitions images into semantically meaningful masks.

\def\imgwidth{0.1035\linewidth}
\begin{figure}[!t]
    \vspace{-0.5em}
    \small
    \smallskip
    \setlength\tabcolsep{1.0pt}
    \renewcommand{\arraystretch}{0.6666}
    \centering
    
    \begin{tabularx}{\linewidth}{@{}Xccccccccc@{}}
    & \multicolumn{3}{c}{Cityscapes} & \multicolumn{3}{c}{COCO-Stuff} & \multicolumn{3}{c}{Potsdam-3} \\
    \cmidrule(l{0.5em}r{0.5em}){2-4} \cmidrule(l{0.5em}r{0.5em}){5-7} \cmidrule(l{0.5em}r{0.5em}){8-10}
    \rotatebox[origin=lB]{90}{\scriptsize{\hspace{-19.2em}\shortstack{\vspace{0.4em}Ground truth}\hspace{1.1em}\shortstack{\ourmethod\\Oracle IDs}\hspace{2.3em}\shortstack{\ourmethod\\Colored}\hspace{3.0em}\shortstack{\vspace{0.4em}Image}}}
    & \includegraphics[width=\imgwidth]
    {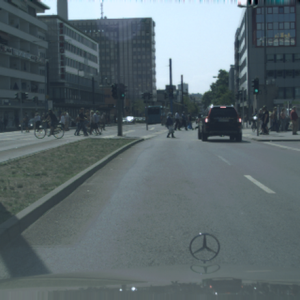} 
    & \includegraphics[width=\imgwidth]
    {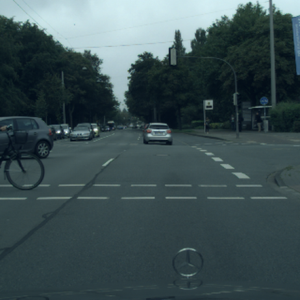} 
    & \includegraphics[width=\imgwidth]
    {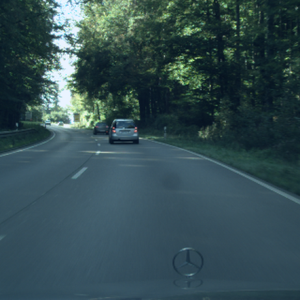} 
    & \includegraphics[width=\imgwidth]
    {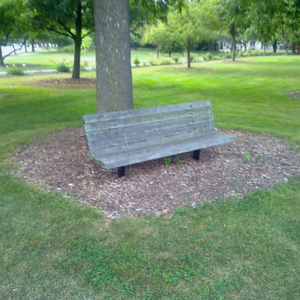} 
    & \includegraphics[width=\imgwidth]
    {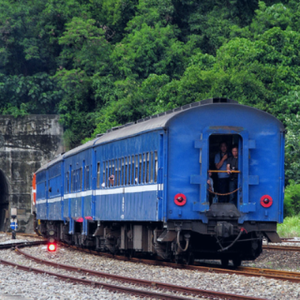} 
    & \includegraphics[width=\imgwidth]
    {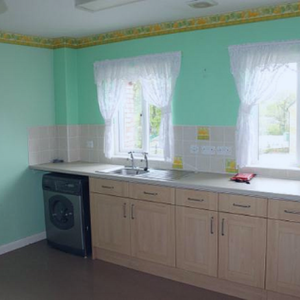} 
    & \includegraphics[width=\imgwidth]
    {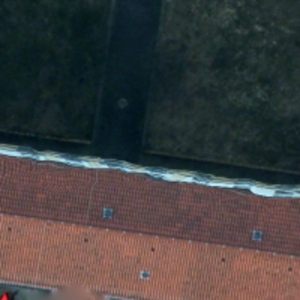}  
    & \includegraphics[width=\imgwidth]
    {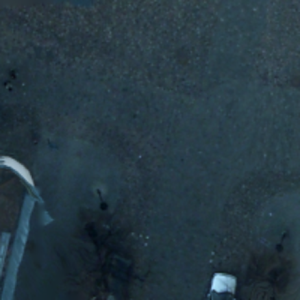} 
    & \includegraphics[width=\imgwidth]
    {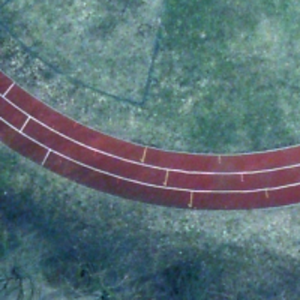}  \\
    & \includegraphics[width=\imgwidth]
    {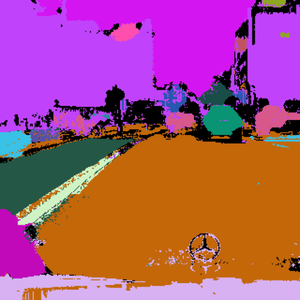} 
    & \includegraphics[width=\imgwidth]
    {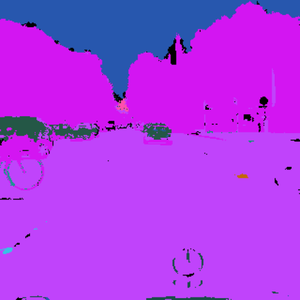} 
    & \includegraphics[width=\imgwidth]
    {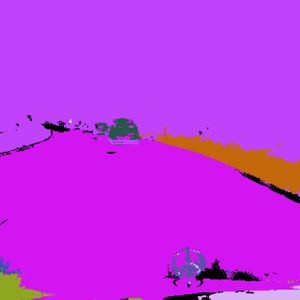} 
    & \includegraphics[width=\imgwidth]
    {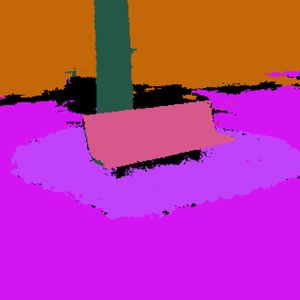} 
    & \includegraphics[width=\imgwidth]
    {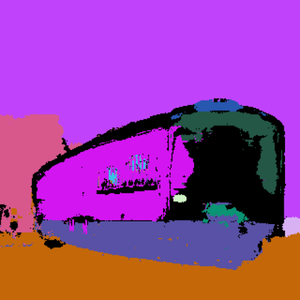} 
    & \includegraphics[width=\imgwidth]
    {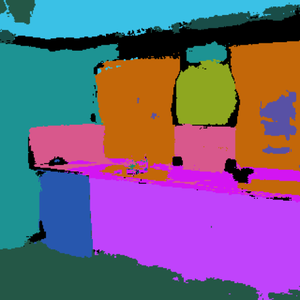} 
    & \includegraphics[width=\imgwidth]
    {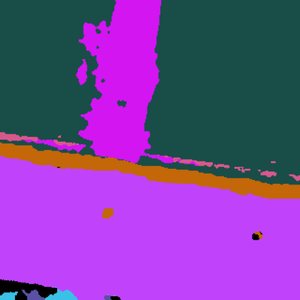}  
    & \includegraphics[width=\imgwidth]
    {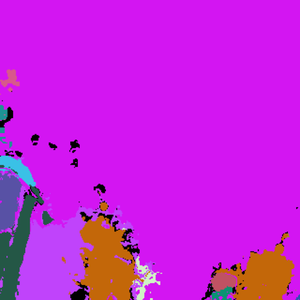} 
    & \includegraphics[width=\imgwidth]
    {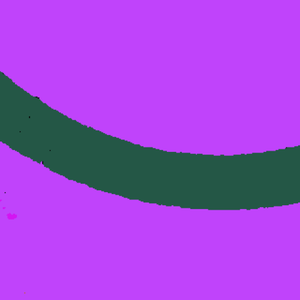}  \\
    & \includegraphics[width=\imgwidth]
    {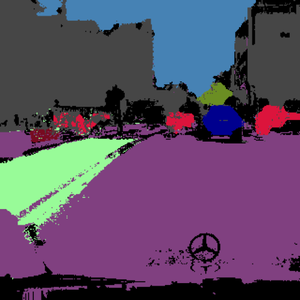} 
    & \includegraphics[width=\imgwidth]
    {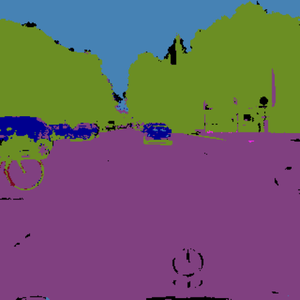} 
    & \includegraphics[width=\imgwidth]
    {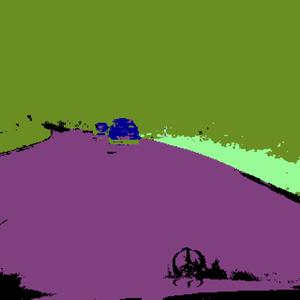} 
    & \includegraphics[width=\imgwidth]
    {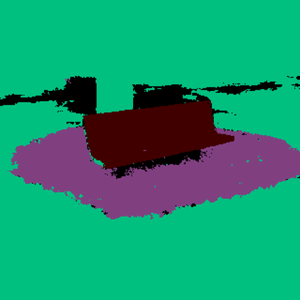} 
    & \includegraphics[width=\imgwidth]
    {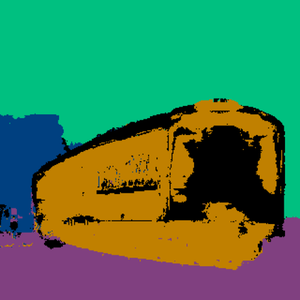} 
    & \includegraphics[width=\imgwidth]
    {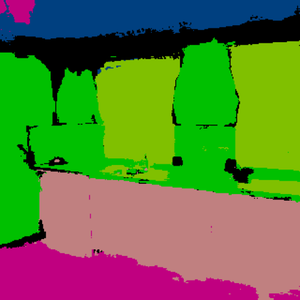} 
    & \includegraphics[width=\imgwidth]
    {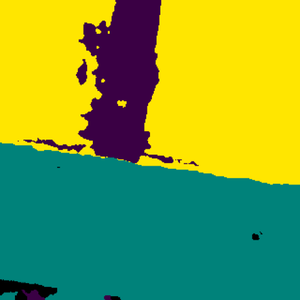}  
    & \includegraphics[width=\imgwidth]
    {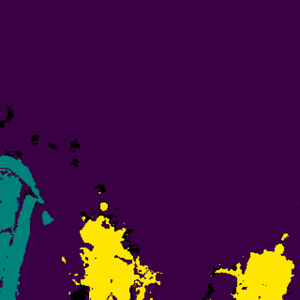} 
    & \includegraphics[width=\imgwidth]
    {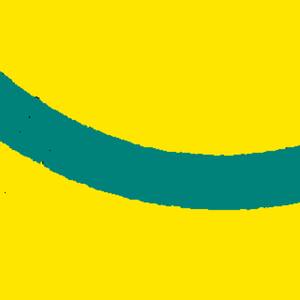}  \\
    & \includegraphics[width=\imgwidth]
    {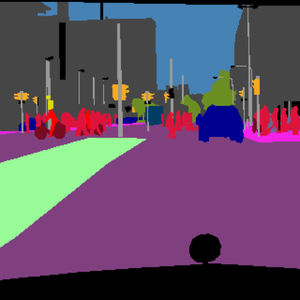} 
    & \includegraphics[width=\imgwidth]
    {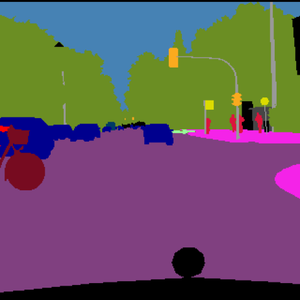} 
    & \includegraphics[width=\imgwidth]
    {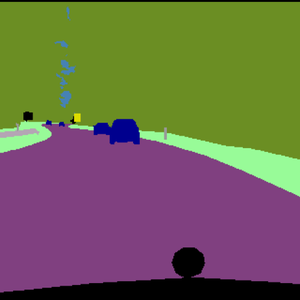} 
    & \includegraphics[width=\imgwidth]
    {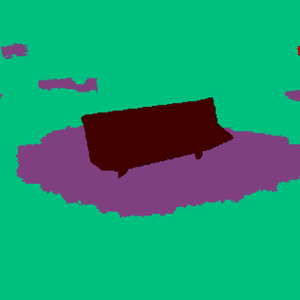} 
    & \includegraphics[width=\imgwidth]
    {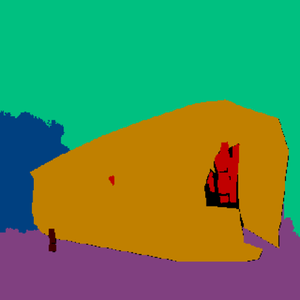} 
    & \includegraphics[width=\imgwidth]
    {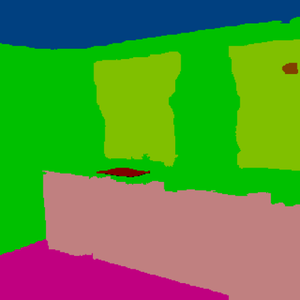} 
    & \includegraphics[width=\imgwidth]
    {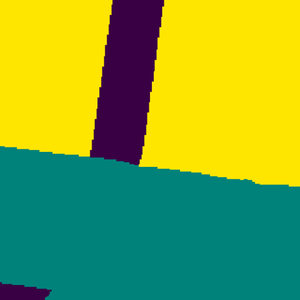}  
    & \includegraphics[width=\imgwidth]
    {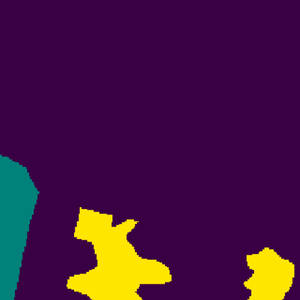} 
    & \includegraphics[width=\imgwidth]
    {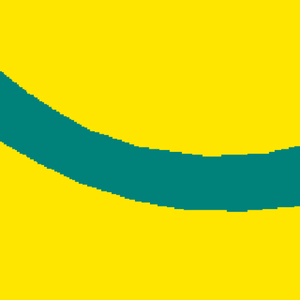}  \\

    \end{tabularx}
    \vspace{-0.5em}
    \caption{\cng{\textbf{Qualitative \ourmethod examples} using DINO ViT-B/8 for Cityscapes, COCO-Stuff, and Potsdam-3. \emph{\ourmethod Colored}  -- each mask proposal is visualized in a different color. \emph{\ourmethod Oracle class IDs} -- each mask is colored in the corresponding ground-truth class color.}} \label{fig:oraclemask_examples}
\end{figure}

\subsection{Failure Cases}
Finally, we would like to discuss observed failure cases of \ourmethodframework. \cref{fig:fail_cases} shows examples of failure cases occurring in the segmentation predictions as well as failure examples for \ourmethod pseudo labels. \cngtwo{For \ourmethodframework, we} observe misclassifications, such as in Cityscapes where buses are often partially segmented as the class ``car'' or that cobblestone is misclassified as ``sidewalk''. For COCO-Stuff, shadows and structures influence the segmentation predictions and confusions also occur (see ``ground'' and ``floor'' in example two). For Potsdam-3, we observe that vehicles and road markings are sometimes erroneously attributed to the class ``building'' instead of ``road''. 
\cngtwo{For \ourmethod pseudo labels, we} identify two main sources of error. First, small objects in cluttered images are sometimes assigned to larger neighboring masks, a phenomenon that can be attributed to the limited backbone feature resolution. This is particularly noticeable for Cityscapes, where the center horizontal area is often detailed (\cf \cref{fig:oraclemask_examples} ``Pole'', ``Traffic Light'', ``Traffic Sign''). Second, \ourmethod sometimes oversegment images containing a large foreground object as seen in the COCO-Stuff example. \cngtwo{Similarly, different visual appearances of the same semantic class can lead to multiple masks (Potsdam-3).} Despite these observations, the simple \ourmethod provide promising mask proposals that correspond well with the ground-truth label.

\def\imgwidth{0.1015\linewidth}
\begin{figure}
    \small
    \smallskip
    \setlength\tabcolsep{1.0pt}
    \renewcommand{\arraystretch}{0.666}
    \centering
    \begin{tabularx}{\textwidth}{@{}rccccccYccc@{}}
    & \multicolumn{2}{c}{Cityscapes} & \multicolumn{2}{c}{COCO-Stuff} & \multicolumn{2}{c}{Potsdam-3} & & \multicolumn{3}{c}{Cityscapes COCO-Stuff Potsdam-3} \\
    \cmidrule(l{0.5em}r{0.5em}){2-3} \cmidrule(l{0.5em}r{0.5em}){4-5} \cmidrule(l{0.5em}r{0.5em}){6-7}
    \rotatebox[origin=lB]{90}{\scriptsize\hspace{-18.7em}\ourmethodframework\hspace{1.0em} Baseline \hspace{0.95em}Ground truth \hspace{1.2em} Image}
    & \includegraphics[width=\imgwidth]
    {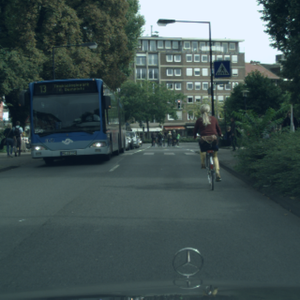} 
    & \includegraphics[width=\imgwidth]
    {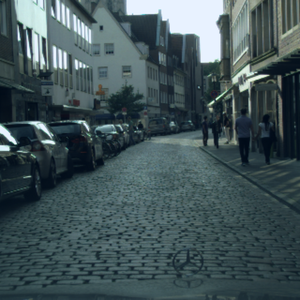} 
    & \includegraphics[width=\imgwidth]
    {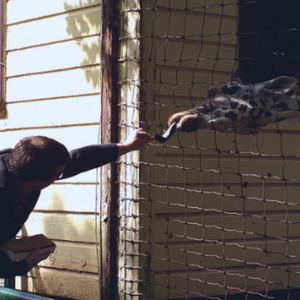} 
    & \includegraphics[width=\imgwidth]
    {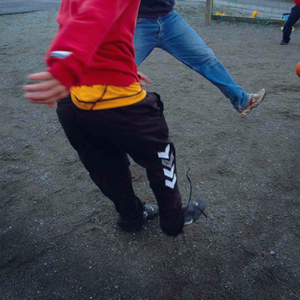} 
    & \includegraphics[width=\imgwidth]
    {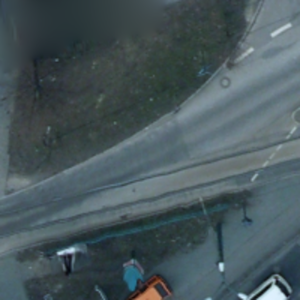} 
    & \includegraphics[width=\imgwidth]
    {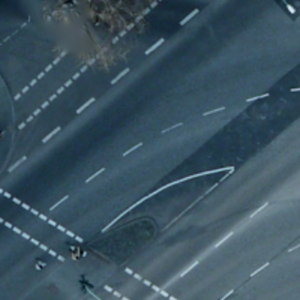} 

    & \rotatebox[origin=lB]{90}{\scriptsize{\hspace{-18.8em}\shortstack{\vspace{0.4em}Ground truth}\hspace{0.9em}\shortstack{\ourmethod\\Oracle IDs}\hspace{2.1em}\shortstack{\ourmethod\\Colored}\hspace{3.0em}\shortstack{\vspace{0.4em}Image}}}
    & \includegraphics[width=\imgwidth]
    {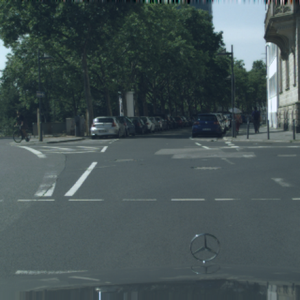} 
    & \includegraphics[width=\imgwidth]
    {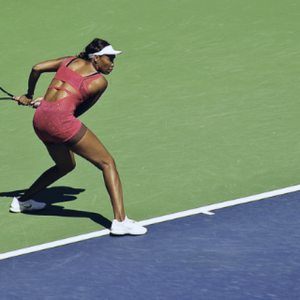} 
    & \includegraphics[width=\imgwidth]
    {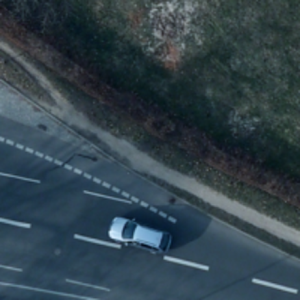}  \\
    & \includegraphics[width=\imgwidth]
    {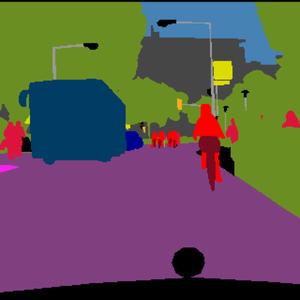} 
    & \includegraphics[width=\imgwidth]
    {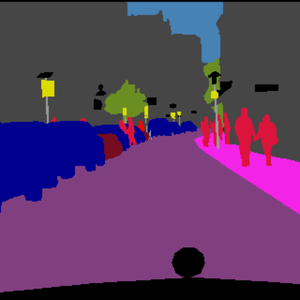} 
    & \includegraphics[width=\imgwidth]
    {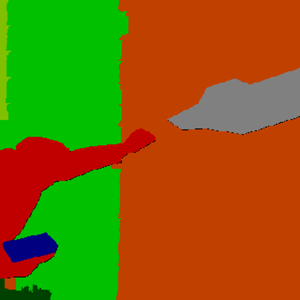} 
    & \includegraphics[width=\imgwidth]
    {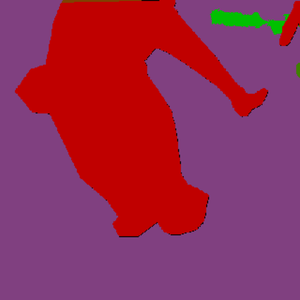} 
    & \includegraphics[width=\imgwidth]
    {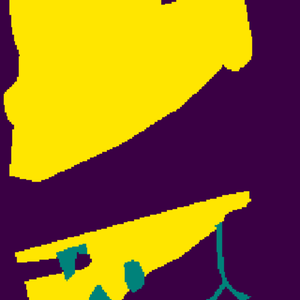} 
    & \includegraphics[width=\imgwidth]
    {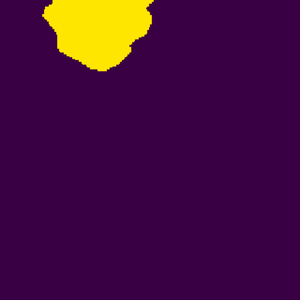}  
    &
    & \includegraphics[width=\imgwidth]
    {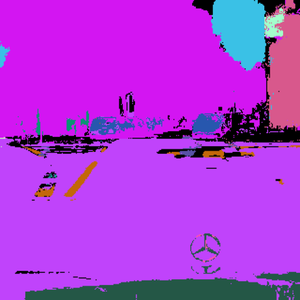} 
    & \includegraphics[width=\imgwidth]
    {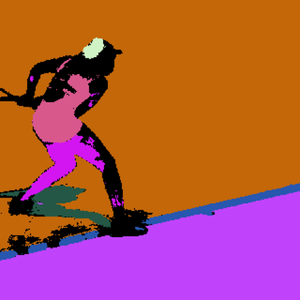} 
    & \includegraphics[width=\imgwidth]
    {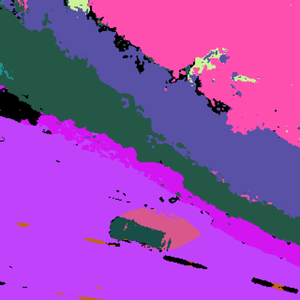}  \\

    & \includegraphics[width=\imgwidth]
    {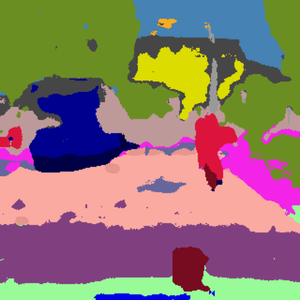} 
    & \includegraphics[width=\imgwidth]
    {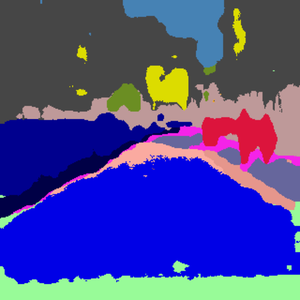} 
    & \includegraphics[width=\imgwidth]
    {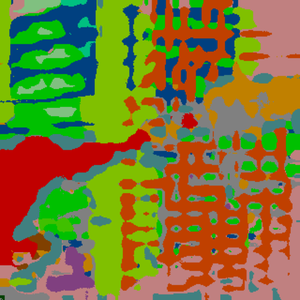} 
    & \includegraphics[width=\imgwidth]
    {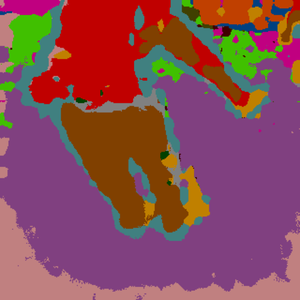} 
    & \includegraphics[width=\imgwidth]
    {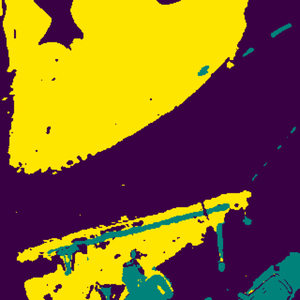} 
    & \includegraphics[width=\imgwidth]
    {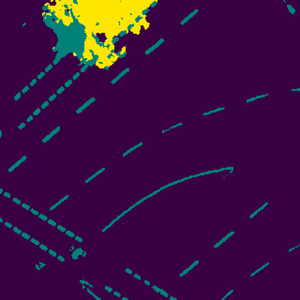}  
    &
    & \includegraphics[width=\imgwidth]
    {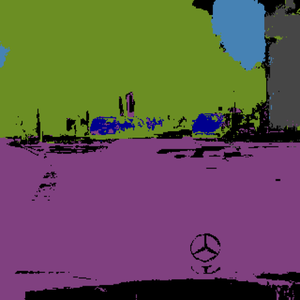} 
    & \includegraphics[width=\imgwidth]
    {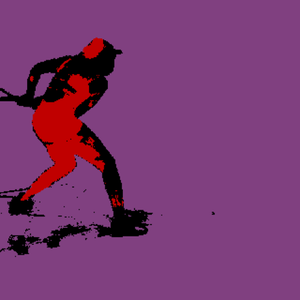} 
    & \includegraphics[width=\imgwidth]
    {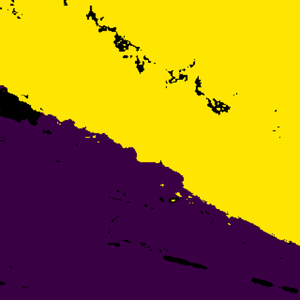}  \\
    & \includegraphics[width=\imgwidth]
    {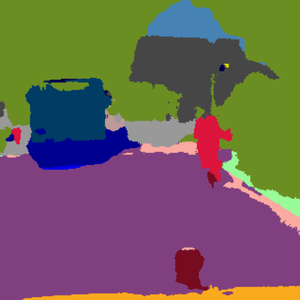} 
    & \includegraphics[width=\imgwidth]
    {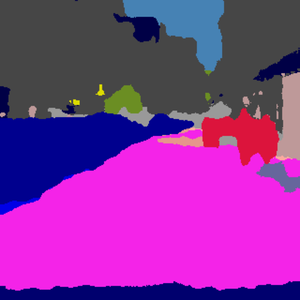} 
    & \includegraphics[width=\imgwidth]
    {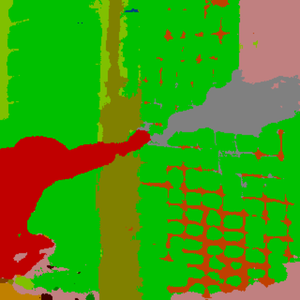} 
    & \includegraphics[width=\imgwidth]
    {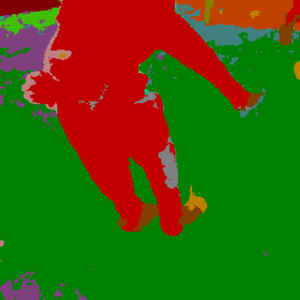} 
    & \includegraphics[width=\imgwidth]
    {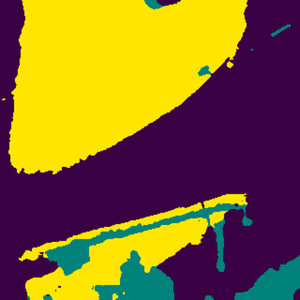} 
    & \includegraphics[width=\imgwidth]
    {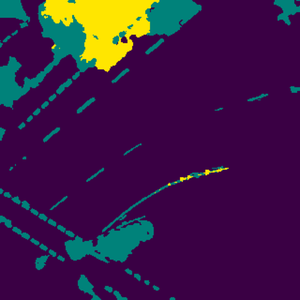}  
    &
    & \includegraphics[width=\imgwidth]
    {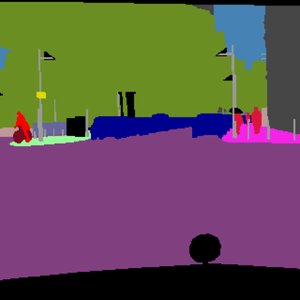} 
    & \includegraphics[width=\imgwidth]
    {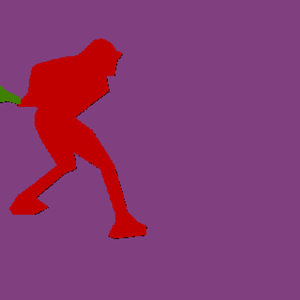} 
    & \includegraphics[width=\imgwidth]
    {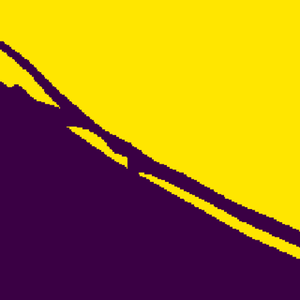}  \\
    \end{tabularx}
\vspace{-0.5em}
\caption{\textbf{Failure cases} for the \ourmethodframework segmentation \emph{(left)} as well as \ourmethod pseudo labels \emph{(right)} using DINO ViT-B/8 for Cityscapes, COCO-Stuff, and Potsdam-3. \label{fig:fail_cases}}
\end{figure}

\section{Implementation}
Since all significant high-level implementation details and hyperparameters have been addressed in the main paper, this section addresses only few remaining details. \cngsr{Both the code and models are publicly available at \url{https://github.com/visinf/primaps}}. Our work is implemented in PyTorch \citep{pytorch_citation}. We build up on the code of \citet{Ji:20219:IIC}, \citet{Gansbeke:2021:USS} and \citet{Hamilton:2022:USS}.

\subsection{Backbone Models}
For each backbone model, we use the corresponding original implementation. Specifically, for DINO \citep{Caron:2021:EPS} and DINOv2 \citep{Oquab:2023:DLR}, we utilize the PyTorch Hub implementation. In the case of STEGO \citep{Hamilton:2022:USS} and HP \citep{Seong:2023:LHP}, we integrate the respective original implementations into our framework.

\rebut{\subsection{Computational Requirements}
We perform all experiments on a single NVIDIA A6000 GPU. The \ourmethodframework optimization runtime varies based on dataset size and the ViT architecture used as the backbone, with ViT-B/8 exhibiting the longest runtime and highest memory consumption. Specifically, using DINO ViT-B/8, PriMaPs-EM requires about 2.5 hours for Cityscapes and Potsdam, and approximately 4 hours for COCO-Stuff, utilizing about 30GB of memory. For the lightest backbone, DINOv2 ViT-S/14, optimization times decrease to about 1.5 hours for Cityscapes and Potsdam-3, and around 2.5 hours for COCO-Stuff, with a memory usage of about 20GB. DINO ViT-S/8 and DINOv2 ViT-B/14 lie in between.}

\subsection{Datasets}
We close with some further details regarding the datasets used.

\inparagraph{Cityscapes} \citep{Cordts:2016:CDS} is an ego-centric street-scene dataset containing $5000$
high-resolution images with $2048\times1024$ pixels. It is split into $2975$ train, $500$ val, and $1525$ test images. Following previous work \citep{Ji:20219:IIC, Cho:2021:PUS, Yin:2022:TTA, Hamilton:2022:USS, Seong:2023:LHP}, evaluation is conducted on the $27$ classes setup using the val split. 

\inparagraph{COCO-Stuff} \citep{Caesar:2018:CST} is a dataset of everyday life scenes containing $80$ things and $91$ stuff classes. Following previous work \citep{Ji:20219:IIC, Cho:2021:PUS, Hamilton:2022:USS, Yin:2022:TTA,Li:2023DCN, Seong:2023:LHP}, we use a reduced variant by \citet{Ji:20219:IIC} containing $49629$ train and $2175$ test images. Hereby, all images consist of at least $75\:\%$ stuff pixels. The dataset is evaluated on the $27$ classes setup. 

\inparagraph{Potsdam-3} \citep{PotsdamDataset} is a remote sensing dataset consisting of $8550$ RGBIR satellite images with $200\times200$ pixels, which is split into $4545$ train and $855$ test images, as well as $3150$ additional unlabeled images. In our experiments, the 3-label variant of Potsdam is evaluated and the additional unlabeled images are not used.

\end{document}